\RequirePackage{fix-cm}
\documentclass[smallextended]{svjour3}       % onecolumn (second format)
\smartqed  % flush right qed marks, e.g. at end of proof
\usepackage{graphicx}
\usepackage[colorlinks=true, allcolors=blue]{hyperref}
\usepackage{booktabs}
\usepackage{algorithmicx}
\usepackage{algorithm}
\usepackage{algpseudocode}
\usepackage{placeins}
\usepackage{afterpage}
\usepackage{subfig}
\usepackage{color, colortbl}
\usepackage[dvipsnames]{xcolor}
\definecolor{highlight}{gray}{0.6}
\newcommand{\code}[1]{\texttt{#1}}

\begin{document}

\title{Evolving Continuous Optimisers from Scratch%\thanks{Grants or other notes
}


\author{Michael A. Lones}


\institute{M Lones \at
              School of Mathematical and Computer Sciences \\
              Heriot-Watt University\\
              Edinburgh, Scotland, UK\\
              \email{m.lones@hw.ac.uk}           %  \\
}

\date{Received: date / Accepted: date}

\maketitle

\begin{abstract}
This work uses genetic programming to explore the space of continuous optimisers, with the goal of discovering novel ways of doing optimisation. In order to keep the search space broad, the optimisers are evolved from scratch using Push, a Turing-complete, general-purpose, language. The resulting optimisers are found to be diverse, and explore their optimisation landscapes using a variety of interesting, and sometimes unusual, strategies. Significantly, when applied to problems that were not seen during training, many of the evolved optimisers generalise well, and often outperform existing optimisers. This supports the idea that novel and effective forms of optimisation can be discovered in an automated manner. This paper also shows that pools of evolved optimisers can be hybridised to further increase their generality, leading to optimisers that perform robustly over a broad variety of problem types and sizes.
\end{abstract}

\section{Introduction}

This work aims to explore the optimisation algorithms that lie outside of conventional human design space. It is based on the premise that innate biases in human thought cause us to only explore certain parts of any design space, and that evolutionary algorithms (EAs) can be a powerful tool for exploring these spaces more widely. For the most part, this work is curiosity-led. However, it also addresses recent issues surrounding the \textit{ad hoc} design of new optimisers through mimicry of natural phenomena. Despite early success with EAs and particle swarm optimisation (PSO), this trend has increasingly resulted in optimisers that are technically novel, but which differ in minor and often arbitrary ways from existing optimisers \cite{sorensen2015metaheuristics,lones2019mitigating,de2021similarity}. If we are to create new optimisation algorithms, arguably it is better to do this in a more systematic, objective and automated manner.

Specifically, this approach uses PushGP \cite{spector2001autoconstructive} to explore optimiser design space. This builds on a significant history of using genetic programming (GP) to explore optimisation behaviours, particularly within the hyperheuristics \cite{burke2013hyper} community, where it is used as a way of fitting existing optimisers to new problems. Since the focus of hyperheuristics is on adapting existing optimisation behaviours, these systems often work with very restricted languages. By comparison, if the aim is to explore a broad design space, it is desirable to use a language that allows the expression of diverse behaviours. This motivates the choice of PushGP \cite{spector2004push}, a genetic programming (GP) system based around a typed, Turing-complete language that contains both low-level primitives and control flow instructions. This provides a basis with which to design optimisers from scratch, largely avoiding the biases that result from re-using existing optimisation building blocks.

This paper presents results from applying this approach to problems in the domain of continuous function optimisation. It extends work reported in \cite{lones2020eurogp}, which built upon methods introduced in \cite{lones2019instruction} for evolving local optimisers. The main contributions are as follows:
\begin{itemize}
	\item It is shown that PushGP can be used to design both local and population-based optimisers, largely from scratch.
	\item It is shown that these optimisers display significant novelty, and are competitive against existing optimisers.
	\item It is shown that significant generality can be achieved by training optimisers on only a single problem, so long as the problem instances are diverse.
	\item It is shown that even more effective optimisers can be built by hybridising pools of diverse evolved optimisers.
\end{itemize}

The paper is organised as follows: Section \ref{related} discusses related work, Section \ref{methods} describes the methods used in this work, Section \ref{results} presents results and analysis, Section \ref{limitations} discusses limitations and future work, and Section \ref{conclusions} concludes.

\section{Related Work} \label{related}

Optimisation involves finding optimal solutions to problems. It is a broad field, but can roughly be broken into two sub-fields: gradient-based methods, which are typically mathematical approaches that use information provided by a function's derivatives, and gradient-free methods, which do not require a mathematical model of the problem. The study of gradient-free methods covers several, overlapping, communities, and this has led to somewhat diffusive terminology \cite{stork2020new}. However, a distinction is often made between local search and population-based methods. Local search includes methods that follow a single trajectory through the search space, typically evaluating a single solution in each iteration of the algorithm. Well-known examples are random search, hill climbing, simulated annealing, and tabu search \cite{handbookofheuristics}. 

Population-based methods follow multiple trajectories, and can be further divided into population-centric methods, which manage the population centrally, and process-centric methods, in which search processes communicate in a distributed fashion. The archetypes of these two methods are the genetic algorithm (GA), and PSO, both modelled upon biological processes. However, many more have followed in their footsteps, and this has led to some debate over what constitutes a novel optimisation algorithm. In part, this debate rests of the meaning of the term \textit{metaheuristic}, which is widely used to refer both to individual optimisation algorithms, and to broader optimisation concepts. If the latter meaning is adopted, then it has been argued that most so-called novel metaheuristics contain very little novelty \cite{sorensen2015metaheuristics,lones2019mitigating,de2021similarity}.

Rather than resting on human design, an alternative approach to developing new optimisers is to use machine learning. Most existing approaches to doing this use GP, and fall within an area that has come to be known as \textit{generative hyperheuristics}. Work in this area focuses on automatically fitting existing metaheuristics to particular problem instances. The targeted metaheuristics include GP itself \cite{teller1996aigp2,edmonds1998metagp,kantschik1999meta}, as well as various EA frameworks \cite{ross2002searching,lourencco2013learning,martin2013evolving,oltean2005evolving,woodward2012automatic}, swarm algorithms \cite{poli2005exploring,van2016evolving,bogdanova2019franken,junior2020franken} and memetic algorithms \cite{ryser2016iterative,kamrath2020automated}. There are two general approaches: assembling new optimisers from the components of existing optimisers \cite{ross2002searching,oltean2005evolving,martin2013evolving,ryser2016iterative,bogdanova2019franken}, or designing new components for existing optimisers, particularly variation operators \cite{teller1996aigp2,edmonds1998metagp,kantschik1999meta,poli2005exploring,diocsan2006evolving,woodward2012automatic}. Various forms of GP have been used for this, including tree-based GP \cite{edmonds1998metagp,martin2013evolving,richter2018automated}, linear GP \cite{oltean2005evolving,goldman2011self,woodward2012automatic}, graph-based GP \cite{teller1996aigp2,kantschik1999meta,shirakawa2009evolution,ryser2016iterative}, grammatical evolution \cite{lourencco2013learning,bogdanova2019franken,junior2020franken}, and recently (building upon our earlier work \cite{lones2019instruction}) PushGP \cite{kamrath2020automated}. Two ways in which most of these hyperheuristic approaches differ from this work are their focus on adapting existing metaheuristic frameworks, and the use of high-level primitives mined from existing algorithmic components. Possibly the closest existing approach is that of Shirakawa et al.\ \cite{shirakawa2009evolution}, who used a graph-based GP with control flow instructions to explore a wider space of optimisers. However, their approach uses high-level primitives that are commonly found in human-designed optimisers, and their aim was to find new optimisers that broadly resemble existing human-designed algorithms.

Whilst the development of hyperheuristics has mostly taken place within the GP community, recently the deep learning and AutoML communities have also been exploring the use of machine learning to design optimisation behaviours. Their main focus has been on using deep learning to design new optimisers for training neural networks \cite{andrychowicz2016learning,wichrowska2017learned,metz2019learned}, but, more recently, there has been interest from the same community in using GP \cite{real2020automlzero}. Similar to this work, the aim is to evolve optimisation (and, more generally, machine learning) behaviours from scratch, and they demonstrate how this approach can re-discover gradient descent. Although their goal is to reduce human bias, the approach uses a language largely comprised of mathematical primitives of the kind used by existing machine learning techniques, and no control flow operations. In this respect, the design space is significantly more constrained than the one used in this paper.

\section{Methods} \label{methods}

\begin{table}[tb!]
	\caption{Vector stack instructions}
	%\vspace{1mm}
	\label{tab:vector}
	\resizebox{\textwidth}{!}{
		\begin{tabular}{@{}llll@{}}
			\toprule
			Instruction&Pop from&Push to&Description\\
			\midrule
			\code{vector.+}&vector, vector&vector&Add two vectors\\
			\code{vector.-}&vector, vector&vector&Subtract two vectors\\
			\code{vector.*}&vector, vector&vector&Multiply two vectors\\
			\code{vector./}&vector, vector&vector&Divide two vectors\\
			\code{vector.scale}&vector, float&vector&Scale vector by scalar\\
			\code{vector.dprod}&vector, vector&float&Dot product of two vectors\\
			\code{vector.mag}&vector&float&Magnitude of vector\\
			\code{vector.dim+}&vector, float, int&vector&Add float to specified component\\
			\code{vector.dim*}&vector, float, int&vector&Multiply specified component by float\\
			\code{vector.apply}&vector, code&vector&Apply code to each component \\
			\code{vector.zip}&vector, vector, code&vector&Apply code to each pair of components\\
			\code{vector.between}&vector, vector, float&vector&Generate point between two vectors\\
			\midrule
			\code{vector.rand}&&vector&Generate random vector of floats\\
			\code{vector.urand}&&vector&Generate random unit vector\\
			\code{vector.wrand}&float&vector&Generate random vector within bounds\\
			\midrule
			\code{vector.current}&integer&vector&Get current point of given swarm member\\
			\code{vector.best}&integer&vector&Get best point of given swarm member\\
			\bottomrule
		\end{tabular}
	}
\end{table}

\begin{table}[!tb]
	\centering
	\caption{Psh parameter settings}
	%\vspace{1mm}
	\resizebox{\textwidth}{!}{
		\begin{tabular}{@{}p{\columnwidth}@{}}
			\toprule
			Population size  = 200\\
			Maximum generations = 50\\
			Tournament size = 5\\
			Program size limit = maximum of 100 instructions\\
			Execution limit = maximum of 100 instruction executions per move\\
			Instructions =
			\code{boolean/float/integer/vector.\{dup flush pop rand rot shove stackdepth swap yank yankdup\}; boolean.\{= and fromfloat frominteger not or xor\}; exec.\{= do*count do*range do*times if iflt noop\}; float.\{\% * + - / < = > abs cos erc exp fromboolean frominteger ln log max min neg pow sin tan\}; input.\{inall inallrev index\}; integer.\{\% * + - / < = > abs erc fromboolean fromfloat ln log max min neg pow\}; vector.\{* / + - apply between dim+ dim* dprod mag pop scale urand wrand zip\}; false; true}\\
			\bottomrule
		\end{tabular}
	}
	\label{table:settings}
\end{table}

\subsection{The Push Language} \label{push}

In this work, optimisation behaviours are expressed using the Push language. Push was designed for use within a GP context, and has a number of features that promote evolvability \cite{spector2001autoconstructive,spector2002genetic,spector2004push}. These include the use of stacks, a mechanism that enables evolving programs to maintain state with less fragility than using conventional indexed memory instructions \cite{langdon2012genetic}. However, it is also Turing-complete, meaning that it is more expressive that many languages used within GP systems. Another notable strength is its type system, which is designed so that all randomly generated programs are syntactically valid, meaning that (unlike type systems introduced to more conventional forms of GP) there is no need to complicate the variation operators or penalise/repair invalid solutions. This is implemented by means of multiple stacks; each stack contains values of a particular type, and all instructions are typed, and will only execute when values are present on their corresponding type stacks.

There are stacks for primitive data types (Booleans, floats, integers) and each of these have both special-purpose instructions (e.g. arithmetic instructions for the integer and float stacks, logic operators for the Boolean stack) and general-purpose stack instructions (push, pop, swap, duplicate, rot etc.) associated with them. Another important stack is the execution stack. At the start of execution, the instructions in a Push program are placed onto this stack and can be manipulated by special instructions; this allows behaviours like looping and conditional execution to be carried out. Finally, there is an input stack, which remains fixed during execution. This provides a way of passing non-volatile constants to a Push program; when popped from the input stack, corresponding values get pushed to the appropriate type stack.

GP systems that use the Push language are known as PushGP. Since a Push program is essentially a list of instructions, it can be represented as a linear array and manipulated using GA-like mutation and crossover operators. This work uses a modified version of \code{Psh} \footnote{\url{http://spiderland.org/Psh/}} (a Java implementation of PushGP) called \code{OptoPsh} \footnote{Available at  \url{https://github.com/michaellones/OptoPsh}}.

\subsection{Domain-Specific Instructions} \label{evolving}

During the course of an optimisation run, an optimiser moves between search points within a particular optimisation problem's search space, resulting in an optimisation trajectory. To allow optimisers to store, express and manipulate search points within this trajectory, an extra vector type has been added to the Push language. This represents search points as fixed-length floating point vectors, one for each dimension of the search space. These can be manipulated using the special-purpose vector instructions shown in Table \ref{tab:vector}, which include: 

\begin{itemize}
	\item Standard mathematical operations, such as pair-wise and component-wise arithmetic, scaling, dot product, and magnitude.
	\item Operations that create a random vector and push it to the vector stack. There are variants for generating unit vectors and vectors with defined bounds; for the latter, the current value $f$ on the float stack is used to set the bounds for each dimension to the interval $[-f,f]$.
	\item \code{vector.apply} and \code{vector.zip} instructions, which allow code to be applied to each component (or each pair of components in the case of zip) using a functional programming style.
	\item The \code{vector.between} instruction, which returns a point on a line between two vectors. For this instruction, the distance along the line is determined by a value popped from the float stack; if this value is between 0 and 1, then the point is a corresponding distance between the two vectors; if less than 0 or greater than 1, then the line is extended beyond the first or second vector, respectively.
\end{itemize}

\begin{algorithm}[tbp!]
	\caption{Evaluating an evolved PushGP optimiser}
	\label{alg::evaluation}
	\begin{algorithmic}[1]
		\State $\mathit{fitness} \gets 0$
		\For{$\mathit{repeats}$} \Comment{Measure fitness over multiple optimisation runs}
		\State $\mathit{pbest} \gets \infty$
		\For{$p \gets 1,\mathit{popsize}$} \Comment{Initialise population state}
		\State $\mathit{prog}_p \gets$ copy of evolved Push program
		\State $\textsc{clearstacks}(\mathit{prog}_p)$
		\State $\mathit{point}_p \gets$ random initial point within search bounds
		\State $\mathit{value}_p \gets \textsc{evaluate}(\mathit{point}_p)$
		\State \textsc{push}($\mathit{point}_p$, $\mathit{prog}_p$.\code{vector}) \Comment{Pass initial search point to program}
		\State \textsc{push}($\mathit{value}_p$, $\mathit{prog}_p$.\code{float}) \Comment{Pass initial objective value to program}
		\State \textsc{push}(\code{true}, $\mathit{prog}_p$.\code{boolean})
		\State \textsc{push}(bounds, $\mathit{prog}_p$.\code{input}) \Comment{Put search space bounds on input stack}
		\State $\mathit{bestval}_p\gets \mathit{value}_p$
		\If{$\mathit{bestval}_p < \mathit{pbest}$}
		\State $\mathit{pbest} \gets \mathit{bestval}_p, \hspace{2mm} \mathit{pbestindex} \gets p$
		\EndIf
		\EndFor
		\For{$m \gets 1,\mathit{moves}$}
		\For{$p \gets 1,\mathit{popsize}$}
		\State \textsc{push}($m$, $\mathit{prog}_p$.\code{integer}) \Comment{Pass move number to program}
		\State \textsc{push}($p$, $\mathit{prog}_p$.\code{integer}) \Comment{Pass population index to program}
		\State \textsc{push}($pbestindex$, $\mathit{prog}_p$.\code{integer}) \Comment{Pass  index of pbest to program}
		\State $\mathit{previous} \gets \mathit{value}_p$
		\State $\textsc{execute}(\mathit{prog}_p)$
		\State $\mathit{point}_p \gets$ \textsc{peek}($\mathit{prog}_p$.\code{vector})
		\Comment{Get next search point from program}
		
		\If{$\mathit{point}_p$ is within search bounds}
		\State $\mathit{value}_p \gets \textsc{evaluate}(\mathit{point}_p)$
		\If{$\mathit{value}_p < \mathit{bestval}_p$}
		\State $\mathit{bestval}_p \gets \mathit{value}_p, \hspace{2mm} \mathit{best}_p \gets \mathit{point}_p$
		\EndIf
		\If{$\mathit{value}_p < \mathit{previous}$}
		\State \textsc{push}(\code{true}, $\mathit{prog}_p$.\code{boolean}) \Comment{Tell program it improved}
		\Else
		\State \textsc{push}(\code{false}, $\mathit{prog}_p$.\code{boolean}) \Comment{Tell program it didn't improve}
		\State \textsc{push}($\mathit{best}_p$, $\mathit{prog}_p$.\code{vector}) \Comment{and remind it of its best point}
		\EndIf
		\State \textsc{push}($\mathit{value}_p$, $\mathit{prog}_p$.\code{float}) \Comment{Pass new objective value}
		\Else
		\State \textsc{push}(\code{false}, $\mathit{prog}_p$.\code{boolean})
		\State \textsc{push}($\infty$, $\mathit{prog}_p$.\code{float}) \Comment{Or indicate move was out of bounds}
		\EndIf
		
		\State \algorithmicif \hspace{0.5mm} $\mathit{best}_p < \mathit{pbest}$ \hspace{0.2mm} \algorithmicthen \hspace{0.5mm} $\mathit{pbest} \gets \mathit{best}_p$
		\EndFor
		\EndFor
		\State $\mathit{fitness} \gets \mathit{fitness} + \mathit{pbest}$
		\EndFor
		\State $\mathit{fitness} \gets \mathit{fitness} / \mathit{repeats}$ \Comment{Mean of best objective values found in each repeat}
	\end{algorithmic}
\end{algorithm}

\subsection{Evolving Optimisers} \label{evolving}

Algorithm \ref{alg::evaluation} outlines the fitness function for evaluating evolved Push programs. A Push program is required to carry out a single move, or optimisation step, each time it is called. In order to generate an optimisation trajectory within a given search space, the Push program is then called multiple times by an outer loop until a specified evaluation budget has been reached. After each call, the value at the top of the Push program's vector stack is popped and the corresponding search point is evaluated. The objective value of this search point, as well as information about whether it was an improving move and whether it moved outside the problem's search bounds, are then passed back to the Push program via the relevant type stacks. During an optimisation run, the contents of a program's stacks are preserved between calls, meaning that Push programs have the capacity to build up their own internal state, and consequently the potential to carry out different types of moves as search progresses.

This framework is used to evolve and evaluate both local and population-based optimisers. For a local optimiser, there is a single Push program, and a single set of stacks that are seeded with a random search point within the bounds of the current optimisation problem. For a population-based optimiser, multiple copies of the same Push program are executed in parallel, and are able to communicate with each other during the course of an optimisation run. At this point, it is important to make a distinction between the population of PushGP (i.e. the population of all programs that are being evolved) and the population of the optimiser (i.e. a population that is formed from multiple copies of a single evolved program during an optimisation run). Since there is scope for confusion, the latter will be referred to as a \textit{swarm} from now on. So, for a swarm of size $s$, there are $s$ copies of a particular program. Each swarm member is started at a different random search point within the bounds of the current optimisation problem. Each copy of the program has its own stacks, meaning that swarm members are able to build up their own internal state independently. Once a optimisation run has started, the swarm members remain persistent, i.e.\ there is no selection mechanism that creates and destroys them.

To allow coordination between swarm members during an optimisation run, two extra instructions are provided, \code{vector.current} and \code{vector.best}. These both look up information about another swarm member's search state, pushing either its current or best seen point of search to the vector stack. The target swarm member is determined by the value at the top of the integer stack (modulus the swarm size to ensure a valid number); if this stack is empty, or contains a negative value, the current or best search point of the current swarm member is returned. This sharing mechanism, combined with the use of persistent search processes, means that the evolved population-based optimisers resemble swarm algorithms such as PSO in their general mechanics. However, there is no selective pressure to use these mechanisms in any particular way, so evolved optimisers are not constrained by the design space of existing swarm optimisers.

The evolutionary parameters are shown in Table \ref{table:settings}.

\subsection{Hybridising Optimisers} \label{hybridising}

An advantage of using EAs is that they tend to find diverse solutions, both due to variance between runs, and due to the use of a population of solutions within runs. This diversity can often be leveraged by combining solutions to form an ensemble, which in turn can be advantageous in terms of generality. Within an optimisation context, the combining of optimisers is usually referred to as hybridisation, and there is a long history of hybridising optimisers in order to build on their individual strengths \cite{blum2016hybrid}. For instance, memetic algorithms, which combine EAs with local search metaheuristics, often out-perform their constituent parts \cite{cotta2017memetic}.

There are many ways in which evolved Push optimisers could potentially be hybridised. In this work, a heterogeneous swarm is used, where each member of the swarm can run a different evolved Push program. The Push programs used in a particular swarm can be sourced from a single PushGP population, or they can be assembled from multiple PushGP runs. The Push program assigned to a particular swarm member can then be persistent throughout an optimisation run, or it can be changed at each iteration. In this paper, a relatively simple approach is investigated, in which we take the best program from each one of a group of PushGP runs, and then randomly assign each swarm member of the heterogeneous swarm a randomly-chosen program at each iteration.

\subsection{Evaluating Optimisers} \label{evaluation}

Optimisation problems were selected from the widely used CEC 2005 real-valued parameter optimisation benchmarks \cite{suganthan2005problem}. These are all minimisation problems, meaning that the aim is to find the input vector (i.e. the search point) that generates the lowest value when passed as an argument to the function. Five of these were used to train optimisers. These were chosen partly because they provide a diverse range of optimisation landscapes, but also because they are relatively fast to evaluate:

\begin{itemize}
	\item $F_1$, the sphere function, a separable unimodal bowl-shaped function. It is the simplest of the benchmarks, and can be solved by gradient descent.
	\item $F_9$, Rastrigin's function, is non-separable and has a large number of regularly spaced local optima whose magnitudes curve towards a bowl where the global minimum is found. The difficulty of this function lies in avoiding the many local optima on the path to the global optimum, though it is made easier by the regular spacing, since the distance between local optima basins can in principle be learnt.
	\item $F_{12}$, Schwefel's problem number 2.13, is non-separable and multimodal and has a small number of peaks that can be followed down to a shared valley region. Gradient descent can be used to find the valley, but the difficulty lies in finding the global mimimum, since the valley contains multiple irregularly-spaced local optima.
	\item $F_{13}$ is a composition of Griewank's and Rosenbrock's functions. This composition leads to a complex surface that is non-separable, highly multimodal and irregular, and hence challenging for optimisers to navigate.
	\item $F_{14}$, a version of Schaffer's $F_6$ Function, comprises concentric elliptical ridges. In the centre is a region of greater complexity where the global optimum lies. It is non-separable and is challenging due to the lack of useful gradient information in most places, and the large number of local optima.
\end{itemize}

\noindent The higher-numbered functions in the CEC 2005 benchmarks comprise variants of four more composition functions, each of which is significantly more complex than those listed above, and each of which takes approximately two order of magnitude longer to evaluate. Due to the number of function evaluations required to evolve optimisers, it was not feasible to use these functions for this purpose. However, one of each type (namely $F_{15}$, $F_{18}$, $F_{21}$ and $F_{24}$) were selected to measure the broader generality of the evolved optimisers in the follow-up analysis.

\begin{figure}[p!]
	\centering
	\includegraphics[width=0.98\columnwidth]{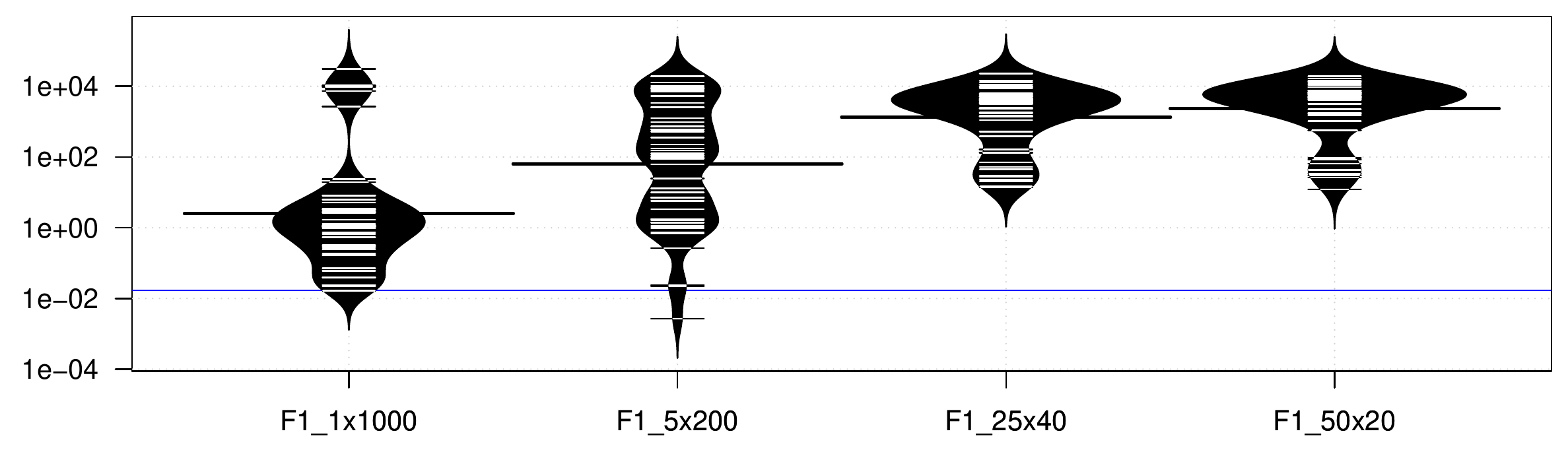}
	\includegraphics[width=0.98\columnwidth]{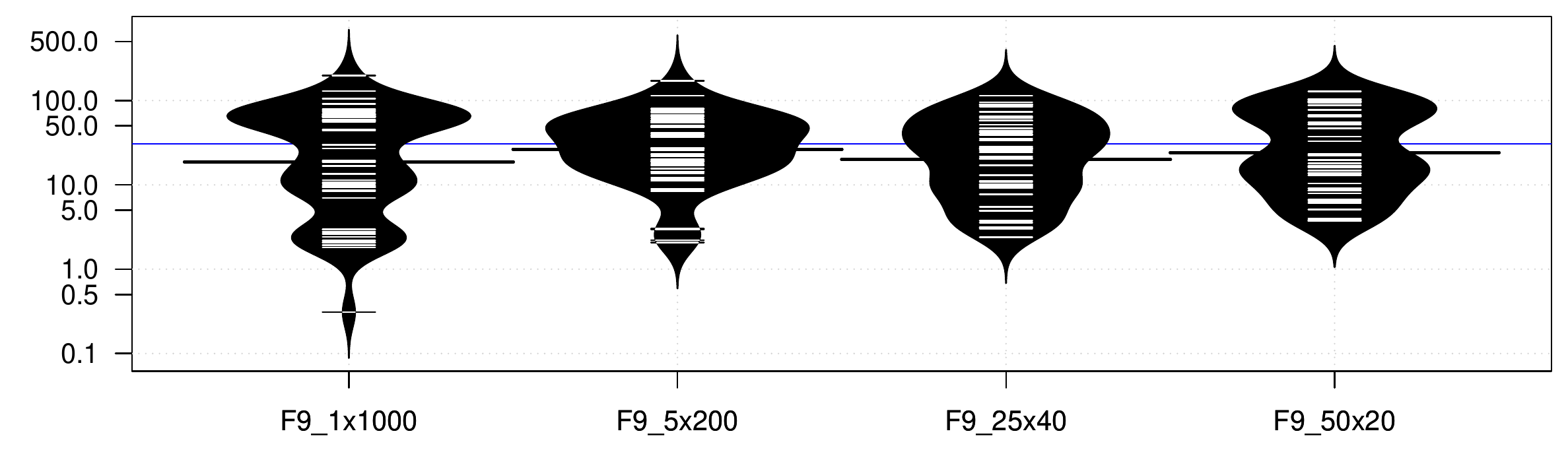}
	\includegraphics[width=0.98\columnwidth]{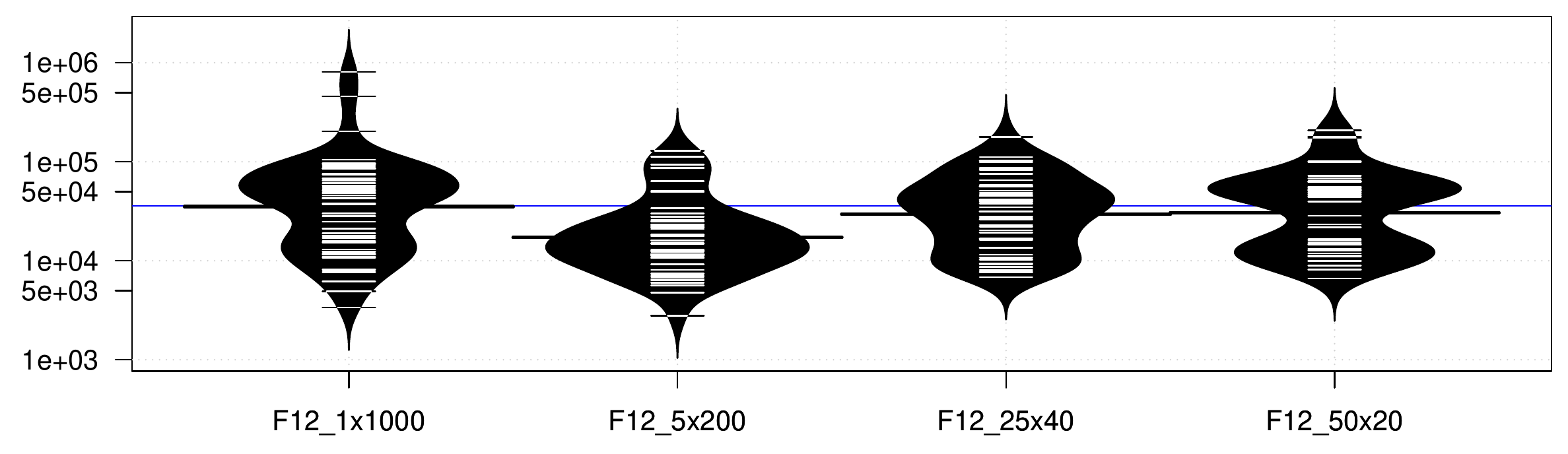}
	\includegraphics[width=0.98\columnwidth]{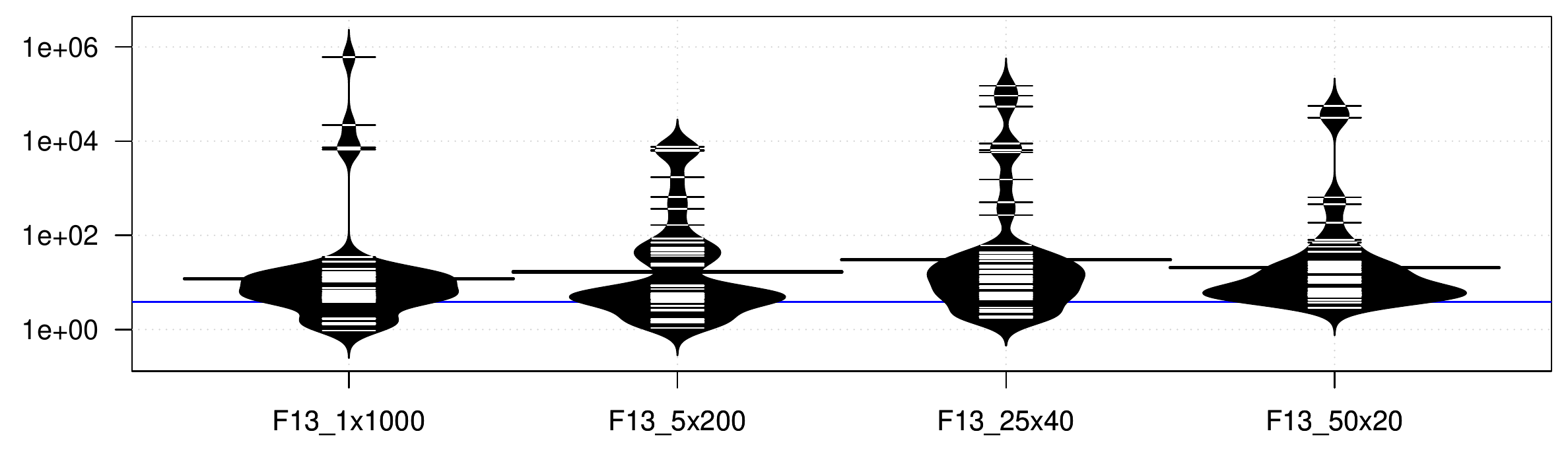}
	\includegraphics[width=0.98\columnwidth]{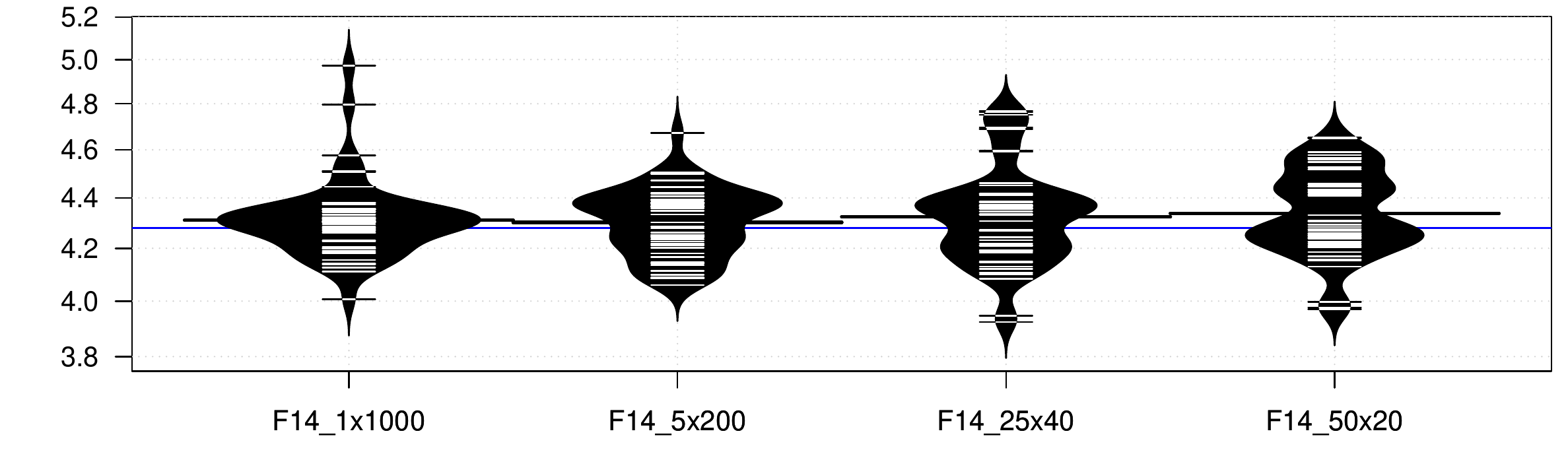}
	\caption{Fitness distributions of 50 runs for each training function and configuration (swarm size $\times$ iterations). The value shown for each run (short horizontal lines) is the mean error for the best solution over 25 reevaluations. Low values are better. Horizontal blue lines show comparative published results for the optimiser G-CMA-ES on the same problems.}
	\label{fig:distributions}
\end{figure}

To discourage overfitting to a particular problem instance, random transformations are applied to each dimension of these functions when they are used to measure fitness during the course of an evolutionary run. Random translations (of up to $\pm$50\% for each axis) prevent the evolving optimisers from learning the location of the optimum, random scalings (50-200\% for each axis) prevent them from learning the distance between features of the landscape, and random axis flips (with 50\% probability per axis) prevent directional biases, e.g. learning which corner of the landscape contains the global optimum.

Fitness is the mean of 10 optimisation runs, each with random initial locations and random transformations. During training, 10-dimensional versions of the functions are used, and the evaluation budget for a single optimisation run is $1E+3$ fitness evaluations (FEs). For the results tables and figures shown in the following section, the best-of-run optimisers are reevaluated over the CEC 2005 benchmark standard of 25 optimisation runs, and random transformations are not applied.

\section{Results} \label{results}

Push optimisers were trained separately on each of the five training functions. Both local search and population-based optimisers were evolved. The local search optimisers have a single point of search, and each optimisation run lasts for $1E+3$ iterations during training. For population-based optimisers, the $1E+3$ evaluation budget can be split between the swarm size and the number of iterations in different ways. In these experiments, splits of (swarm size $\times$ iterations) 50$\times$20, 25$\times$40 and 5$\times$200 are used. For each of these configurations, and for each of the five training functions, 50 independent runs of PushGP were carried out.

Fig. \ref{fig:distributions} shows the resulting fitness distributions, where fitness is the mean error when the best-of-run optimisers are reevaluated over 25 optimisation runs. To give an idea of how these error rates compare to an established general purpose optimiser, Fig. \ref{fig:distributions} also plots (as horizontal lines) the mean errors reported for G-CMA-ES \cite{auger2005restart} on each of these functions. G-CMA-ES is a variant of the Covariance Matrix Adaptation Evolution Strategy (CMA-ES), with the addition of restarts and an increasing population size at each restart. It is a relatively complex algorithm and is generally regarded as the overall winner of the CEC 2005 competition.

The first thing evident from these fitness distributions is that the trade-off between swarm size and number of iterations is more significant for some problems than others. For $F_1$, better optimisers are generally found for smaller swarm sizes, with the local search (i.e. 1$\times$1000) distribution having the lowest mean error. This makes sense, because the unimodal $F_1$ landscape favours intensification over diversification. For $F_{12}$, the sweet spot appears to be for 5$\times$200, possibly reflecting the number of peaks in the landscape (i.e. 5). For the other problems, the differences appear relatively minor, and effective optimisers could be evolved for all configurations. In most cases, the best optimiser for a particular problem is an outlier within the distributions, so may not reflect any intrinsic benefit of one configuration over another. That said, four of these best-in-problem optimisers used small populations (2 with 1$\times$1000 and 2 with 5$\times$200). This suggests that it might be easier to find optimisers that use smaller populations rather than larger ones, perhaps reflecting the additional effort required to evolve effective coordination mechanisms within population-based optimisers. 

Fig. \ref{fig:distributions} also shows that, for all the training problems, the PushGP runs found at least one optimiser that performed better, on average, than G-CMA-ES. For the simplest problem $F_1$, there was only one evolved optimiser that beat the general purpose optimiser. For the other problems, many optimisers were found that performed better. This is perhaps unsurprising, given that the capacity to overfit problems is a central motivation for existing work on hyperheuristics. However, an important difference in this work is the use of random problem transformations during training, since this causes the problems to exhibit greater generality, preventing optimisers from over-learning specific features of the landscape. The results suggest that this does not significantly impact the ability of evolved optimisers to out-perform general purpose optimisers on the problem on which they were trained.

% G-CMA-ES, DE, 
\setlength{\tabcolsep}{4pt}
\begin{table*}[tb!]
	\caption{Generality of evolved optimisers. For each optimiser trained with a fitness evaluation budget (FEs) of $1E+3$ function evaluations on 10D functions, mean errors are shown for 25 optimisation runs of the 10D, 30D and 50D versions of the functions for a budget of both $1E+3$ and $1E+4$. The mean rank is also shown, and the best result for each combination of problem dimensionality (D) and evaluation budget is underlined for each function number and ranking. Grey text shows where an optimiser is being evaluated on the function on which it was trained.}\label{table:errors}
	\resizebox{\textwidth}{!}{
		\centering
		\vspace{2mm}
		\begin{tabular}{@{}lllllllllllll@{}}
			\toprule
			D & FEs & Optimiser & $F_1$ & $F_9$ & $F_{12}$ & $F_{13}$ & $F_{14}$ & $F_{15}$ & $F_{18}$ & $F_{21}$ & $F_{24}$ & $\overline{\mathrm{Rank}}$\\
    \midrule
10    & 1E+3 & CMA-ES        & 1.70E$-$2 & 3.07E+1 & 3.59E+4 & 3.84E+0 & 4.28E+0 & 4.12E+2 & \underline{8.43E+2} & \underline{9.05E+2} & \underline{5.89E+2} & 2.56 \\
&       & $F_{1}$ best  & \textcolor{highlight}{\underline{2.48E$-$3}} & 7.28E+1 & 3.29E+4 & 5.26E+0 & 4.47E+0 & 6.93E+2 & 1.13E+3 & 1.22E+3 & 1.26E+3 & 3.67 \\
&       & $F_{9}$ best  & 1.32E+4 & \textcolor{highlight}{\underline{3.27E$-$1}} & 9.32E+3 & 1.18E+0 & 4.86E+0 & 2.04E+2 & 1.11E+3 & 1.32E+3 & 1.38E+3 & 3.44 \\
&       & $F_{12}$ best & 3.10E+3 & 7.28E+0 & \textcolor{highlight}{\underline{2.79E+3}} & 2.43E+0 & 4.52E+0 & \underline{1.53E+2} & 9.02E+2 & 9.66E+2 & 9.76E+2 & \underline{2.44} \\
&       & $F_{13}$ best & 3.56E+4 & 2.44E+0 & 4.63E+4 & \textcolor{highlight}{\underline{1.05E+0}} & 4.82E+0 & 3.71E+2 & 1.26E+3 & 1.38E+3 & 1.21E+3 & 4.11 \\
&       & $F_{14}$ best & 4.11E+2 & 7.76E+1 & 9.97E+4 & 2.69E+2 & \textcolor{highlight}{\underline{4.04E+0}} & 7.57E+2 & 1.19E+3 & 1.35E+3 & 1.28E+3 & 4.78 \\
\cmidrule{2-13}
& 1E+4 & CMA-ES        & \underline{5.20E$-$9} & 6.21E+0 & 2.98E+3 & 9.71E$-$1 & 3.91E+0 & 2.99E+2 & \underline{6.02E+2} & \underline{7.05E+2} & \underline{3.04E+2} & \underline{2.11} \\
&       & $F_{1}$ best  & \textcolor{highlight}{2.44E$-$6} & 8.05E+1 & 2.36E+4 & 3.60E+0 & 4.50E+0 & 7.35E+2 & 1.09E+3 & 1.14E+3 & 9.46E+2 & 4.22 \\
&       & $F_{9}$ best  & 1.45E$-$3 & \textcolor{highlight}{2.06E$-$1} & 7.72E+3 & 7.04E$-$1 & 4.85E+0 & 1.86E+2 & 1.12E+3 & 1.31E+3 & 1.32E+3 & 4.11 \\
&       & $F_{12}$ best & 5.96E$-$4 & 7.47E$-$2 & \textcolor{highlight}{\underline{3.93E+2}} & 4.98E$-$1 & 4.21E+0 & \underline{8.88E+1} & 8.82E+2 & 8.73E+2 & 3.21E+2 & \underline{2.11} \\
&       & $F_{13}$ best & 1.51E$-$4 & \underline{3.66E$-$6} & 3.07E+4 & \textcolor{highlight}{\underline{3.45E$-$1}} & 4.90E+0 & 3.74E+2 & 1.30E+3 & 1.36E+3 & 1.20E+3 & 4.11 \\
&       & $F_{14}$ best & 1.37E+1 & 5.16E+1 & 3.77E+4 & 1.62E+1 & \textcolor{highlight}{\underline{3.57E+0}} & 5.72E+2 & 1.06E+3 & 1.21E+3 & 8.85E+2 & 4.33 \\
\cmidrule{1-13}
30    & 1E+3 & CMA-ES        & \underline{8.16E+2} & 2.53E+2 & 1.67E+6 & 1.14E+2 & 1.42E+1 & 6.69E+2 & 9.45E+2 & \underline{9.44E+2} & \underline{3.05E+2} & 2.33 \\
&       & $F_{1}$ best  & \textcolor{highlight}{7.75E+4} & 4.36E+2 & 1.07E+6 & 3.47E+4 & 1.45E+1 & 1.10E+3 & 1.36E+3 & 1.40E+3 & 1.59E+3 & 4.78 \\
&       & $F_{9}$ best  & 7.63E+4 & \textcolor{highlight}{3.24E+2} & 1.07E+6 & 4.00E+3 & 1.45E+1 & 1.07E+3 & 1.37E+3 & 1.38E+3 & 1.47E+3 & 4.11 \\
&       & $F_{12}$ best & 5.74E+4 & 1.18E+2 & \textcolor{highlight}{3.46E+5} & 3.62E+1 & 1.44E+1 & 5.16E+2 & \underline{9.09E+2} & 1.30E+3 & 1.43E+3 & \underline{2.11} \\
&       & $F_{13}$ best & 1.63E+5 & \underline{1.00E+2} & \underline{1.73E+5} & \textcolor{highlight}{\underline{1.84E+1}} & 1.47E+1 & \underline{4.33E+2} & 1.40E+3 & 1.36E+3 & 1.71E+3 & 3.44 \\
&       & $F_{14}$ best & 2.14E+4 & 4.15E+2 & 2.19E+6 & 3.52E+4 & \textcolor{highlight}{\underline{1.38E+1}} & 1.16E+3 & 1.31E+3 & 1.37E+3 & 1.44E+3 & 4.00 \\
\cmidrule{2-13}
& 1E+4 & CMA-ES        & \underline{5.42E$-$9} & 4.78E+1 & 2.51E+5 & 3.80E+0 & 1.38E+1 & 3.87E+2 & 9.08E+2 & \underline{5.47E+2} & \underline{9.26E+2} & \underline{2.00} \\
&       & $F_{1}$ best  & \textcolor{highlight}{1.36E+2} & 3.68E+2 & 4.08E+5 & 4.18E+1 & 1.44E+1 & 9.10E+2 & 1.24E+3 & 1.02E+3 & 1.43E+3 & 4.00 \\
&       & $F_{9}$ best  & 6.40E+4 & \textcolor{highlight}{3.27E+2} & 1.09E+6 & 3.52E+3 & 1.46E+1 & 1.11E+3 & 1.36E+3 & 1.37E+3 & 1.46E+3 & 5.22 \\
&       & $F_{12}$ best & 5.97E$-$2 & 5.76E+0 & \textcolor{highlight}{\underline{3.43E+4}} & 5.00E+0 & 1.41E+1 & \underline{2.25E+2} & \underline{9.00E+2} & 1.25E+3 & 1.35E+3 & 2.22 \\
&       & $F_{13}$ best & 2.44E+4 & \underline{5.04E$-$2} & 1.26E+5 & \textcolor{highlight}{\underline{1.42E+0}} & 1.47E+1 & 3.80E+2 & 1.41E+3 & 1.29E+3 & 1.65E+3 & 3.78 \\
&       & $F_{14}$ best & 1.64E+2 & 3.33E+2 & 1.17E+6 & 3.97E+3 & \textcolor{highlight}{\underline{1.33E+1}} & 8.51E+2 & 1.13E+3 & 1.25E+3 & 1.30E+3 & 3.78 \\
\cmidrule{1-13}
50    & 1E+3 & CMA-ES        & \underline{1.12E+4} & 5.41E+2 & 8.04E+6 & 9.25E+4 & 2.41E+1 & 8.49E+2 & 1.07E+3 & \underline{1.05E+3} & \underline{1.09E+3} & 2.78 \\
&       & $F_{1}$ best  & \textcolor{highlight}{2.10E+5} & 9.18E+2 & 8.21E+6 & 8.58E+5 & 2.42E+1 & 1.38E+3 & 1.48E+3 & 1.56E+3 & 1.72E+3 & 5.56 \\
&       & $F_{9}$ best  & 1.27E+5 & \textcolor{highlight}{\underline{1.50E+2}} & \underline{7.49E+5} & \underline{1.80E+2} & \underline{2.40E+1} & \underline{4.16E+2} & 1.31E+3 & 1.40E+3 & 1.51E+3 & \underline{2.00} \\
&       & $F_{12}$ best & 1.28E+5 & 4.78E+2 & \textcolor{highlight}{3.20E+6} & 2.37E+2 & \underline{2.40E+1} & 8.34E+2 & \underline{9.09E+2} & 1.40E+3 & 1.51E+3 & 2.56 \\
&       & $F_{13}$ best & 2.75E+5 & 3.17E+2 & 1.10E+6 & \textcolor{highlight}{3.17E+4} & 2.45E+1 & 5.97E+2 & 1.48E+3 & 1.52E+3 & 1.85E+3 & 4.11 \\
&       & $F_{14}$ best & 6.39E+4 & 8.36E+2 & 9.09E+6 & 2.80E+5 & \textcolor{highlight}{2.36E+1} & 1.29E+3 & 1.40E+3 & 1.46E+3 & 1.54E+3 & 4.00 \\
\cmidrule{2-13}
& 1E+4 & CMA-ES        & \underline{5.87E$-$9} & 1.04E+2 & 2.52E+6 & 8.68E+0 & 2.37E+1 & 4.22E+2 & 9.25E+2 & \underline{1.01E+3} & \underline{9.72E+2} & \underline{2.44} \\
&       & $F_{1}$ best  & \textcolor{highlight}{2.92E+4} & 7.26E+2 & 2.87E+6 & 1.46E+2 & 2.43E+1 & 1.02E+3 & 1.31E+3 & 1.18E+3 & 1.54E+3 & 4.89 \\
&       & $F_{9}$ best  & 2.37E+4 & \textcolor{highlight}{1.05E+0} & \underline{1.42E+5} & 4.86E+0 & 2.39E+1 & \underline{1.65E+2} & 1.29E+3 & 1.40E+3 & 1.50E+3 & 3.00 \\
&       & $F_{12}$ best & 6.53E+3 & 4.87E+1 & \textcolor{highlight}{2.64E+5} & 1.22E+1 & 2.39E+1 & 3.89E+2 & \underline{9.00E+2} & 1.31E+3 & 1.43E+3 & 3.33 \\
&       & $F_{13}$ best & 1.06E+5 & \underline{7.00E$-$1} & 4.24E+5 & \textcolor{highlight}{\underline{3.56E+0}} & 2.45E+1 & 2.93E+2 & 1.42E+3 & 1.50E+3 & 1.78E+3 & 4.11 \\
&       & $F_{14}$ best & 4.66E+2 & 7.04E+2 & 6.51E+6 & 6.00E+4 & \textcolor{highlight}{\underline{2.29E+1}} & 9.58E+2 & 1.23E+3 & 1.35E+3 & 1.40E+3 & 3.78 \\
\bottomrule
		\end{tabular}
	}
\end{table*}

\subsection{Generality of Evolved Optimisers}

However, the ability to out-perform general purpose optimisers on the problem on which they were trained is arguably not that important, especially given the substantial overhead of evolving the optimiser in the first place. Of more interest is how the evolved optimisers generalise to larger and different problems that they were not trained on. Table \ref{table:errors} gives insight into this, showing how well the best evolved optimiser for each training function generalises to both larger instances of the same function and to the other eight functions (i.e. the other four training functions, and the four used only for evaluation). Mean error rates are shown both for the 10-dimensional problems used in training, and for the more difficult 30 and 50-dimensional versions, both with the $1E+3$ evaluation budget used in training and for a larger $1E+4$ evaluation budget. To give an indication of relative performance, the average rank of each optimiser across the nine functions is also shown for each combination of dimensionality and evaluation budget.

Table \ref{table:errors} shows that most of the evolved optimisers generalise well to the 30D and 50D versions of the 10D function on which they were trained. The optimisers trained on $F_{12}$, $F_{13}$ and $F_{14}$ functions do best in this regard, outperforming G-CMA-ES on the 30D (in addition to the 10D) versions of these functions. The $F_1$ optimiser is the only one which generalises relatively poorly to larger problems, being beaten by G-CMA-ES and several of the other evolved optimisers on the 30D and 50D versions.

However, the most interesting observation from Table \ref{table:errors} is that many of the optimisers also generalise well to other functions, often out-performing the optimiser that was trained on lower-dimensional versions of the same function. Notably, in terms of its mean ranking across these nine functions, the $F_{12}$ optimiser does better than G-CMA-ES at all dimensionalities for a budget of $1E+3$ FEs. This level of generality is quite surprising, given that it only saw one of the nine functions (and only at 10D) during training.

Whilst the $F_{12}$ optimiser has the best mean rank at 10D and 30D, it is surpassed by the $F_9$ optimiser at 50D, which ranks first for five out of the nine functions. This is also quite suprising, not only because the $F_9$ optimiser does relatively poorly at lower dimensionalities, but also because it was trained on a landscape where the optima are always regularly spaced --- which is not true of the other functions. However, it is known that the relative rankings of optimisers can change considerably as the dimensionality of a search space increases \cite{graham2019thesis}, and this seems to be a reflection of this more general observation.

It is notable that G-CMA-ES always performs best on two of the most complex landscapes, $F_{21}$ and $F_{24}$. This may suggest that training on simpler functions limits the generality of optimisers when applied to more complex landscapes. However, G-CMA-ES is beaten in most cases by the $F_{12}$ optimiser on the other two complex landscapes, $F_{15}$ and $F_{18}$, so this conclusion is certainly not clear.

%Also notable is that the $F_{13}$ optimiser comes first in three out of the five 30D problems, though this is balanced by coming last in four others.

Table \ref{table:errors} also shows what happens when the evaluation budget is increased to $1E+4$ FEs, an order of magnitude more than the budget the evolved optimisers were able to use per optimisation run during training. First of all, it is evident that all the evolved optimisers continue to make progress when run for longer, so the restricted training budget does not appear to have impaired their generality in this regard. At 10D, the $F_{12}$ optimiser ties with G-CMA-ES. However, at 30D and 50D, G-CMA-ES does out-perform the evolved optimisers when using a larger evaluation budget. This suggests that training on $1E+3$ FEs pushes evolution towards optimisers that make relatively good progress on small budgets, but which may be less competitive on larger budgets. Nevertheless, it is notable that the $F_{12}$ optimiser still ranks first for three of the 30D functions, the same number as G-CMA-ES.

Overall, these results demonstrate that it is possible to train an optimiser on a single function at a relatively low dimensionality, and, so long as the instances are diverse enough to prevent overfitting (i.e using random transformations in this case), the resulting optimiser can generalise to a diverse range of optimisation landscapes at significantly higher dimensionalities. This is useful to know, since it means that optimisers can be trained relatively efficiently, i.e. they do not need to be evaluated on multiple functions at high dimensionalities during the evolutionary process.

\begin{figure}[!p]
	\centering
	\includegraphics[width=0.32\columnwidth, trim={0.8cm 0.8cm 0.2cm 1.5cm},clip]{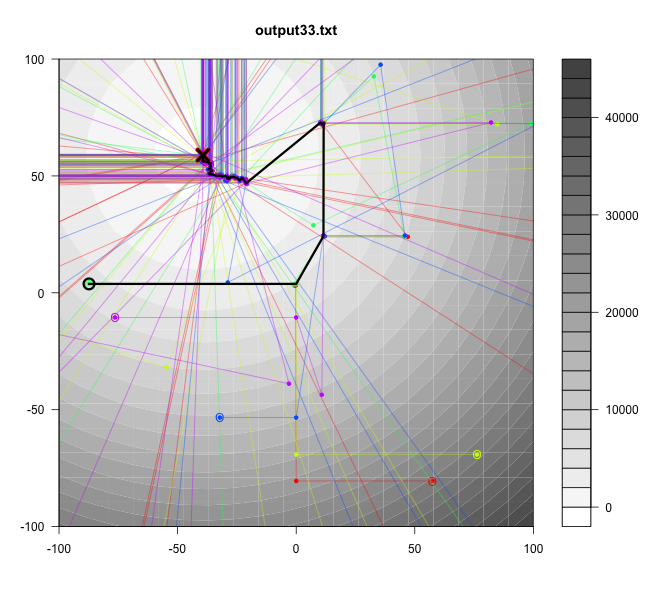}\hspace{-1mm}
	\includegraphics[width=0.32\columnwidth, trim={0.8cm 0.8cm 0.2cm 1.5cm},clip]{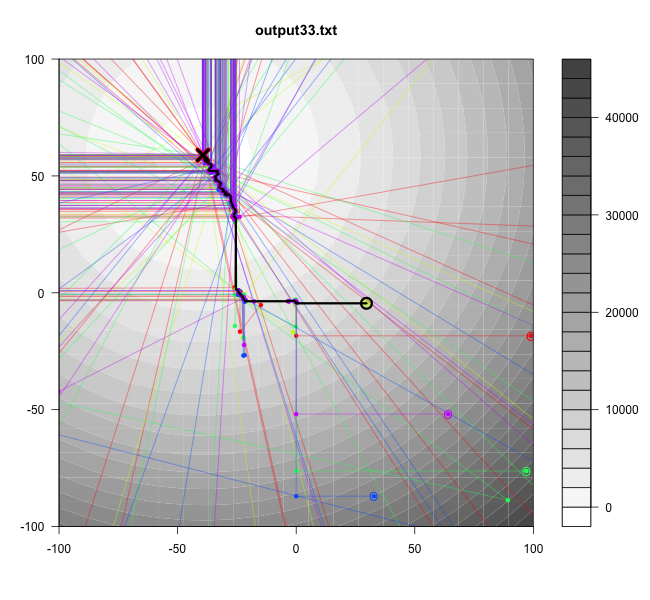}\hspace{-2mm}
	\includegraphics[width=0.32\columnwidth, trim={0.8cm 0.8cm 0.2cm 1.5cm},clip]{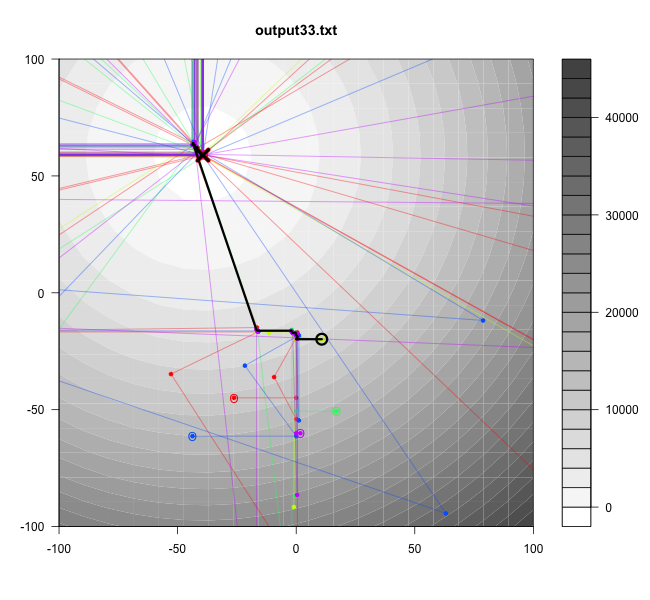}\hspace{-1mm}\\
	\includegraphics[width=0.32\columnwidth, trim={0.8cm 0.8cm 0.2cm 1.5cm},clip]{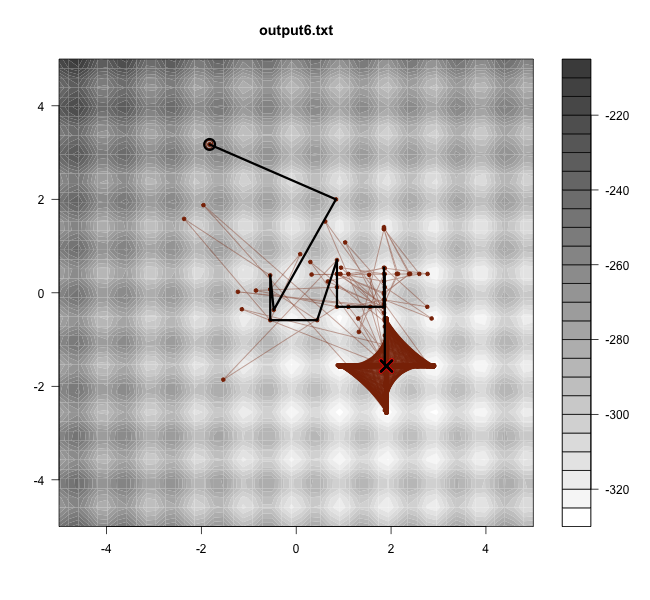}\hspace{-1mm}
	\includegraphics[width=0.32\columnwidth, trim={0.8cm 0.8cm 0.2cm 1.5cm},clip]{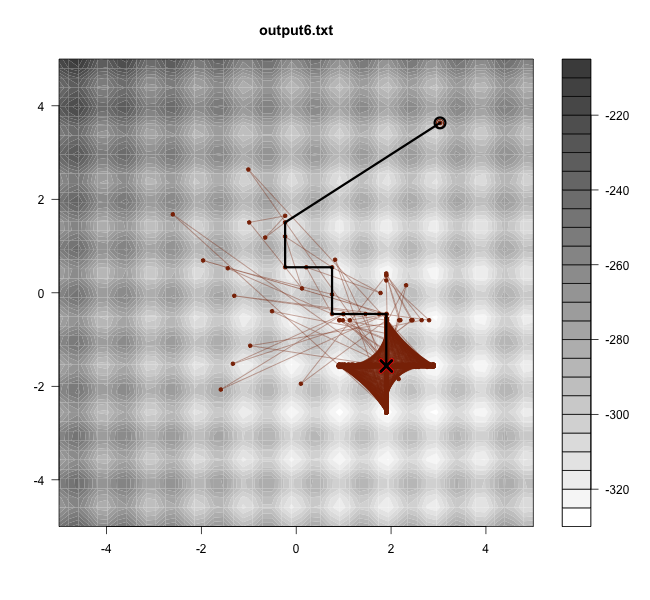}\hspace{-2mm}
	\includegraphics[width=0.32\columnwidth, trim={0.8cm 0.8cm 0.2cm 1.5cm},clip]{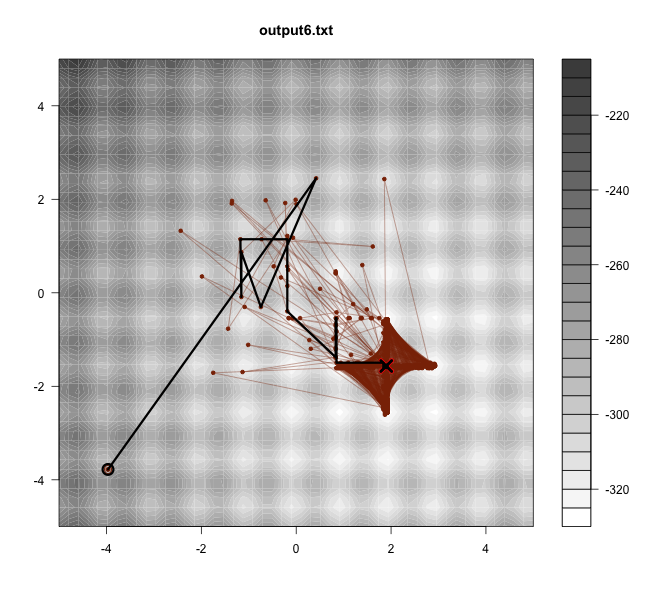}\hspace{-1mm}\\
	\includegraphics[width=0.32\columnwidth, trim={0.8cm 0.8cm 0.2cm 1.5cm},clip]{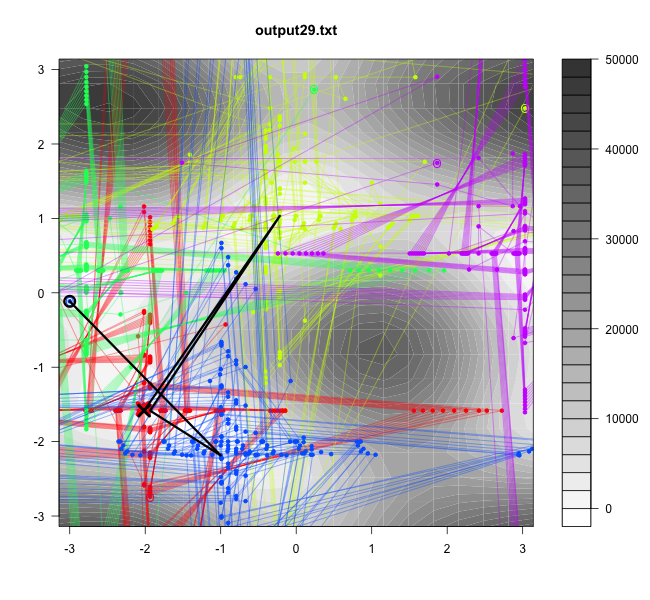}\hspace{-1mm}
	\includegraphics[width=0.32\columnwidth, trim={0.8cm 0.8cm 0.2cm 1.5cm},clip]{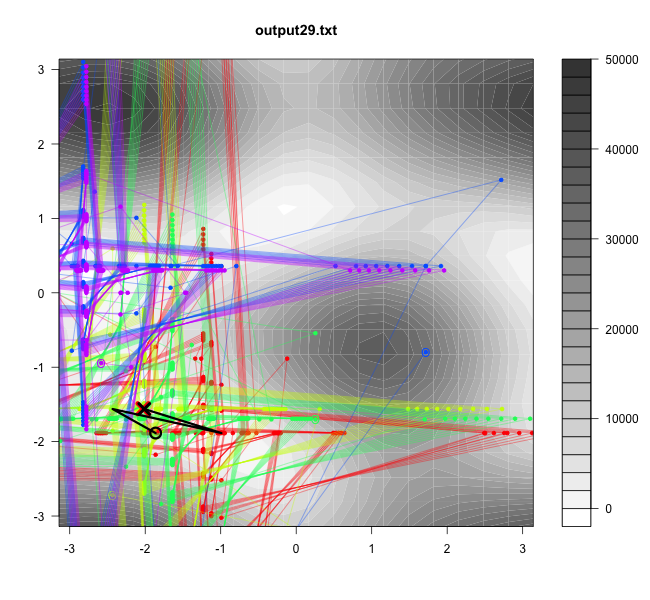}\hspace{-2mm}
	\includegraphics[width=0.32\columnwidth, trim={0.8cm 0.8cm 0.2cm 1.5cm},clip]{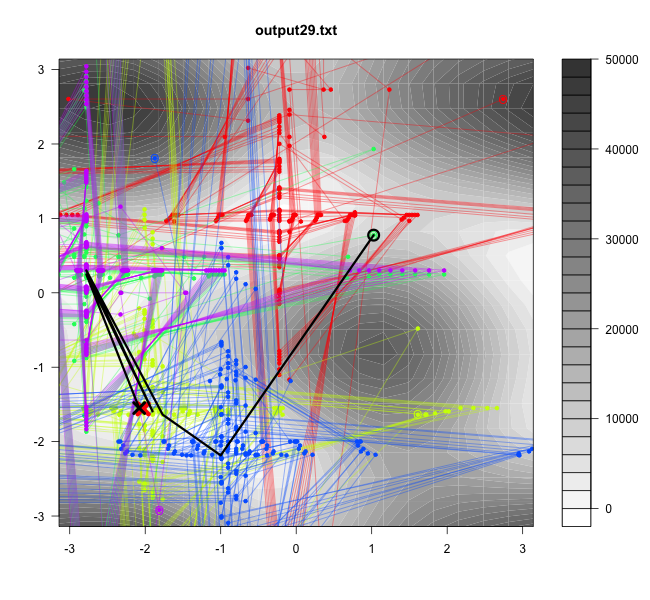}\hspace{-1mm}\\
	\includegraphics[width=0.32\columnwidth, trim={0.8cm 0.8cm 0.2cm 1.5cm},clip]{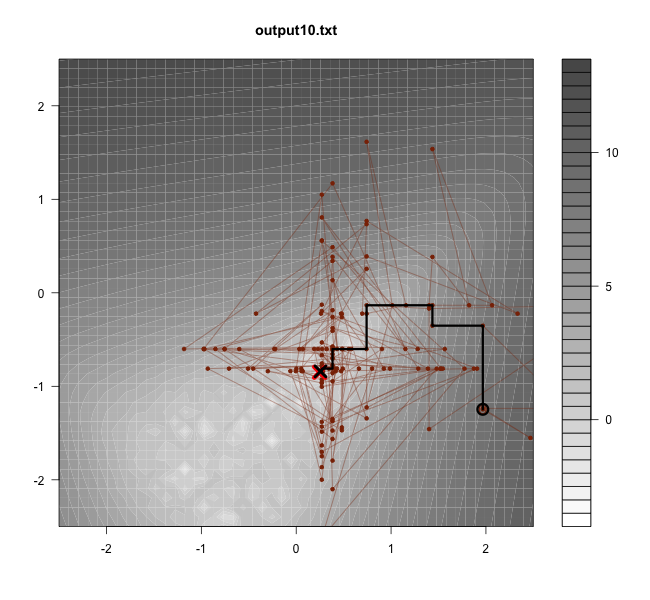}\hspace{-1mm}
	\includegraphics[width=0.32\columnwidth, trim={0.8cm 0.8cm 0.2cm 1.5cm},clip]{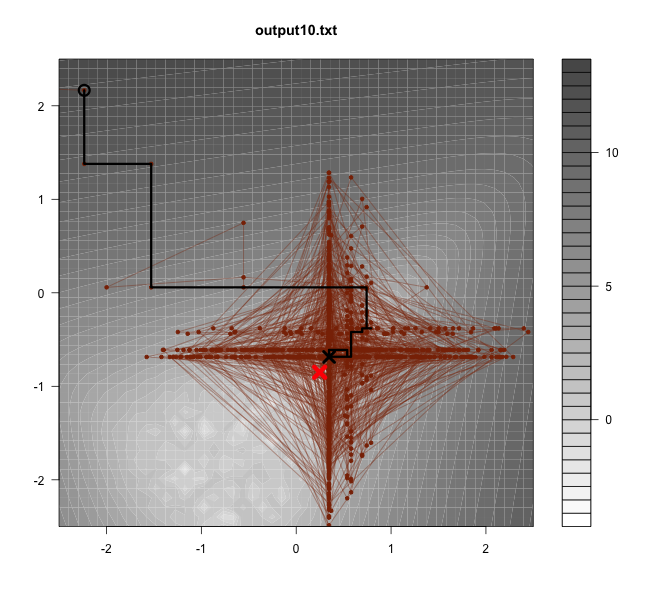}\hspace{-2mm}
	\includegraphics[width=0.32\columnwidth, trim={0.8cm 0.8cm 0.2cm 1.5cm},clip]{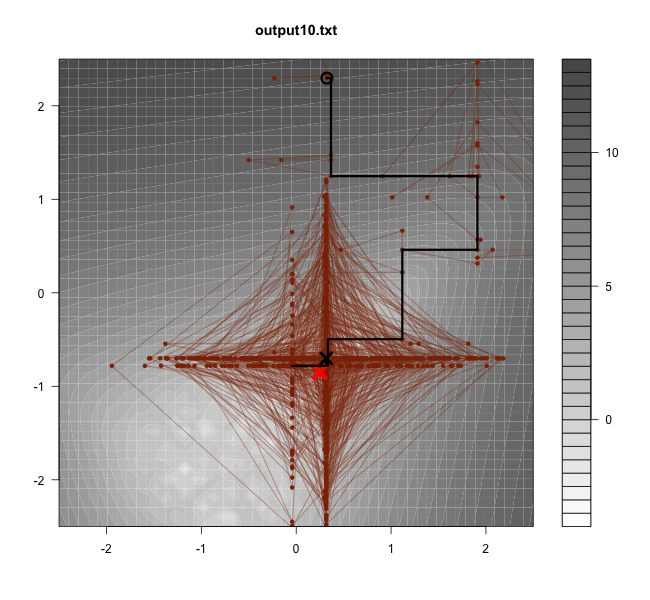}\hspace{-1mm}\\
	\includegraphics[width=0.32\columnwidth, trim={0.8cm 0.8cm 0.2cm 1.5cm},clip]{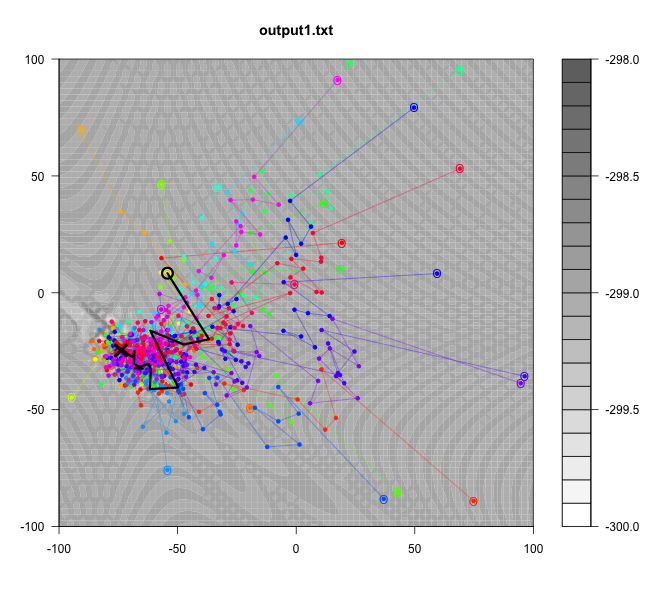}\hspace{-1mm}
	\includegraphics[width=0.32\columnwidth, trim={0.8cm 0.8cm 0.2cm 1.5cm},clip]{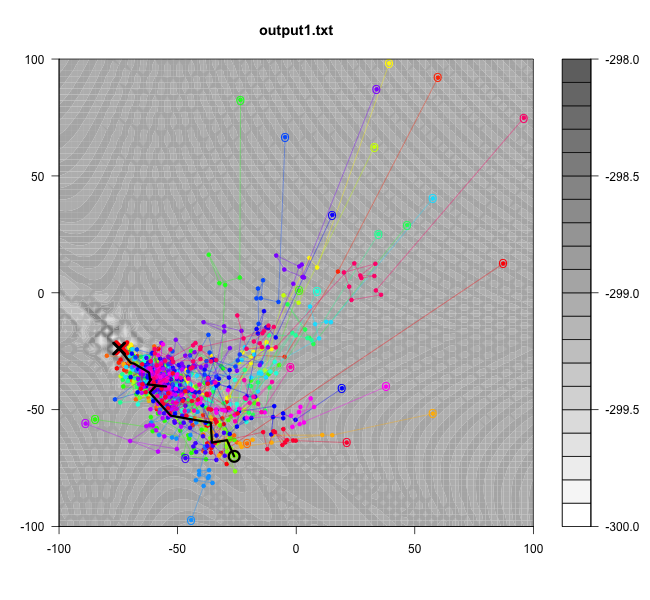}\hspace{-2mm}
	\includegraphics[width=0.32\columnwidth, trim={0.8cm 0.8cm 0.2cm 1.5cm},clip]{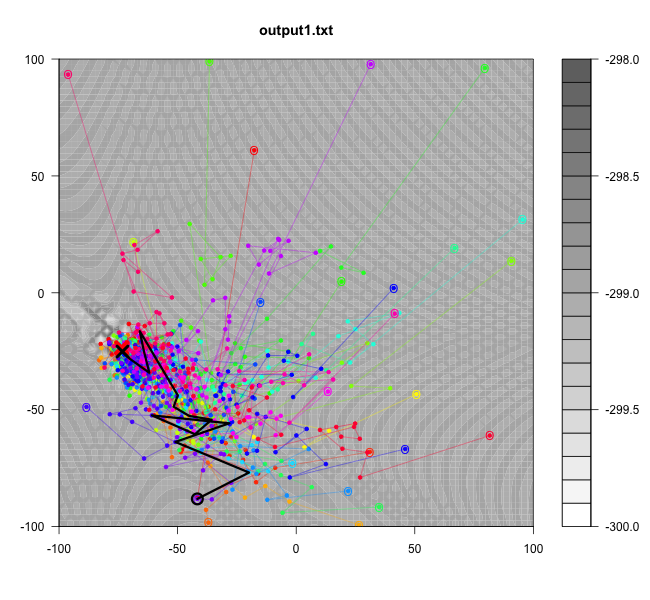}\hspace{-1mm}
	\caption{Example trajectories of best-in-problem optimisers (top-bottom: $F_1$, $F_9$, $F_{12}$, $F_{13}$, $F_{14}$) on 2D versions of their training functions. The global optimum is shown as a red cross. The best point reached by the optimiser is shown as a black cross. The optimiser's trajectory is shown as a black line, starting from a black circle. Each swarm member's trajectory is also shown using separate colours. The search landscape is shown in the background as a contour plot.}
	\label{fig:bests}
\end{figure}

% trim={<left> <lower> <right> <upper>}

\subsection{Behaviour of Evolved Optimisers}

\begin{table*}[tb!]
	\caption{Evolved Push expressions of best-in-problem optimisers}
	\label{tab:optimisers}
	%\vspace{1mm}
	\resizebox{\textwidth}{!}{
		\centering
		\begin{tabular}{@{}lp{11cm}@{}}
			\toprule
			$F_1$&\code{(exec.dup float.- vector.- float.pop vector.zip vector.zip integer.swap float.cos float.- float.cos float.- float.yank vector.best vector.wrand float.abs float.dup float.frominteger vector.- vector.dim*)}\\
			$F_9$&\code{(input.stackdepth float.frominteger vector.yank vector.wrand boolean.dup integer.fromboolean vector.swap integer.rot float.frominteger float.sin vector.yank vector.shove vector.dim+ vector.yank 0.0 float.> input.inall boolean.not 1 boolean.dup vector.pop boolean.stackdepth)}\\
			$F_{12}$&\code{(vector.stackdepth vector.swap float.fromboolean integer.fromboolean integer.rand vector.dim+ float.+ vector.swap integer.rand 0 vector.swap integer.max integer.= vector.stackdepth integer.dup vector.- integer.dup integer.rand vector.-  vector.dim+ vector.mag float.frominteger float.tan integer.rot vector.dim+)}\\
			$F_{13}$&\code{(integer.- float.sin vector.wrand integer.yankdup vector.dim* vector.- input.inall float.sin vector.-)}\\
			$F_{14}$&\code{(float.< float./ vector.best vector.yankdup float.ln float.max float.stackdepth 0.48999998 float.abs vector.between vector.wrand vector.scale integer.yank input.index vector.- float.rand float.neg 0.97999996 float.- 0.97999996 vector.wrand vector.scale vector.-)}\\
			\bottomrule
		\end{tabular}
	}
\end{table*}

\begin{table}[htbp]
  \centering
  \caption{Relative rate of instruction use within all the best-of-run optimisers, showing, for each swarm size $\times$ iterations configuration, the 20 most commonly used Push instructions.}
  \resizebox{\textwidth}{!}{
  \code{
    \begin{tabular}{@{}rllll@{}}
    	\toprule
    \multicolumn{1}{l}{\normalfont Rank} & \normalfont 1x1000 & \normalfont 5x200 & \normalfont 25x40 & \normalfont 50x20 \\
    \midrule
    \normalfont 1     & vector.wrand & vector.wrand & float.rand & float.rand \\
    \normalfont 2     & float.rand & integer.rand & vector.best & vector.dim* \\
    \normalfont 3     & vector.+ & float.rand & vector.dim* & vector.best \\
    \normalfont 4     & vector.- & float.tan & vector.dim+ & vector.between \\
    \normalfont 5     & float.tan & vector.best & float.tan & float.cos \\
    \normalfont 6     & input.inall & vector.dim+ & float.cos & vector.dim+ \\
    \normalfont 7     & input.inallrev & vector.dim* & integer.rand & float.sin \\
    \normalfont 8     & integer.rand & float.sin & vector.wrand & float.tan \\
    \normalfont 9     & vector.dim* & vector.- & vector.between & exec.flush \\
    \normalfont 10    & vector.best & float.cos & float.sin & integer.rand \\
    \normalfont 11    & float.sin & vector.+ & exec.flush & vector.wrand \\
    \normalfont 12    & vector.dim+ & float.frominteger & boolean.shove & vector.+ \\
    \normalfont 13    & float.cos & vector.between & integer.+ & float.- \\
    \normalfont 14    & input.index & vector.stackdepth & float.neg & integer.< \\
    \normalfont 15    & integer.ln & exec.flush & float.frominteger & float.abs \\
    \normalfont 16    & float.fromboolean & input.index & vector.+ & float./ \\
    \normalfont 17    & float.frominteger & input.inallrev & boolean.xor & float.fromboolean \\
    \normalfont 18    & float.\% & float.ln & integer.= & float.min \\
    \normalfont 19    & float.yank & float.log & boolean.and & integer.rot \\
    \normalfont 20    & float.log & float.* & boolean.pop & boolean.shove \\
    \bottomrule
    \end{tabular}}}%
  \label{tab:usage}%
\end{table}%

Table \ref{tab:optimisers} shows the evolved Push expression used by the best evolved optimiser for each training function, in each case slightly simplified by removing instructions that have no effect on their fitness. Whilst it is sometimes possible to understand their behaviour by looking at the evolved expressions alone, it is usually possible to gain more insight by observing the interpreter's stack states as they run, and by observing their trajectories on 2D versions of the function landscapes. Figs. \ref{fig:bests} and \ref{fig:generality} show examples of the latter, both for the functions they were trained on, and the other eight functions, respectively. In almost all cases, optimisers generalise well to the easier 2D version of the function they were trained on, and it can be seen in Fig. \ref{fig:bests} that in each case the optimiser's trajectory approaches the global optimum. Overall, these five optimisers display a diverse range of search behaviours, a number of which are quite novel:

\begin{figure}[!p]
	\centering
	\includegraphics[width=0.19\columnwidth, trim={0.7cm 0 3.5cm 1.5cm},clip]{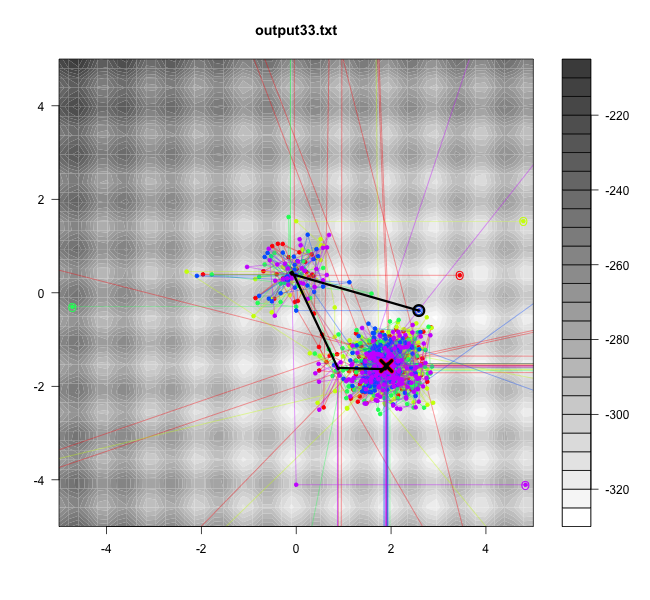}
	\includegraphics[width=0.19\columnwidth, trim={0.7cm 0 3.5cm 1.5cm},clip]{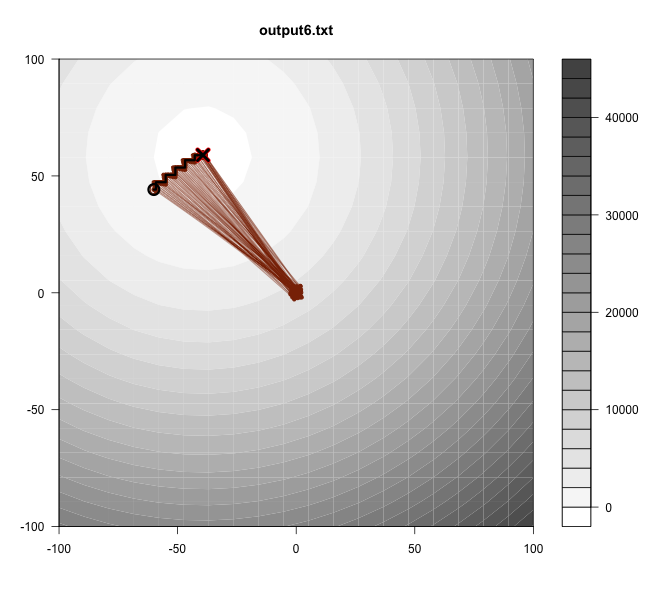}
	\includegraphics[width=0.19\columnwidth, trim={0.7cm 0 3.5cm 1.5cm},clip]{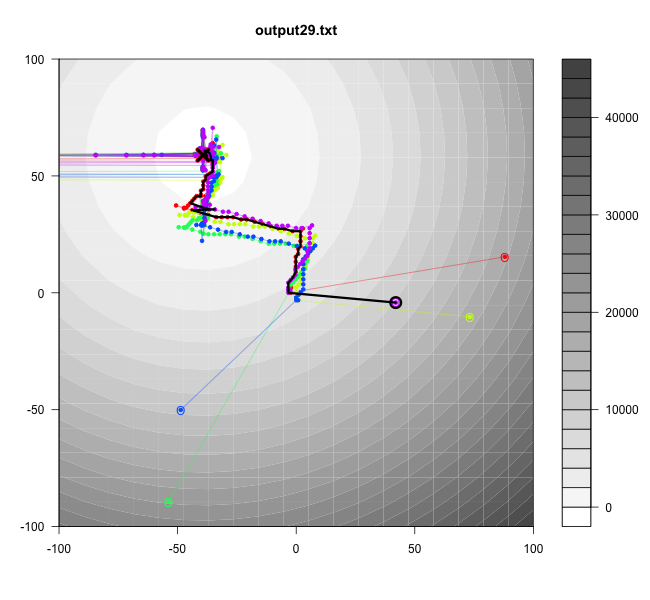}
	\includegraphics[width=0.19\columnwidth, trim={0.7cm 0 3.5cm 1.5cm},clip]{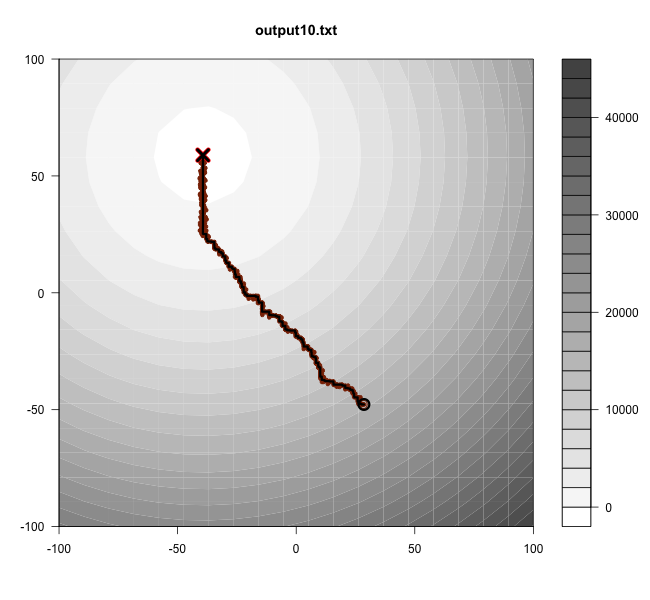}
	\includegraphics[width=0.19\columnwidth, trim={0.7cm 0 3.5cm 1.5cm},clip]{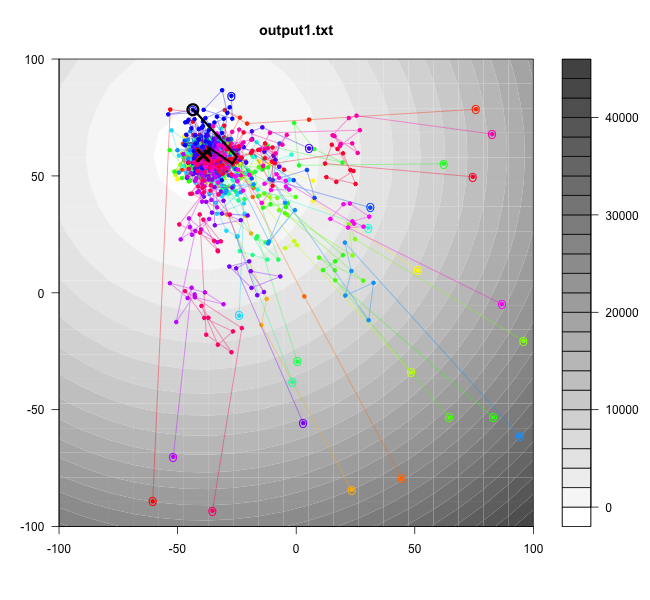}\\
	\vspace{-1mm}
	\includegraphics[width=0.19\columnwidth, trim={0.7cm 0 3.5cm 1.5cm},clip]{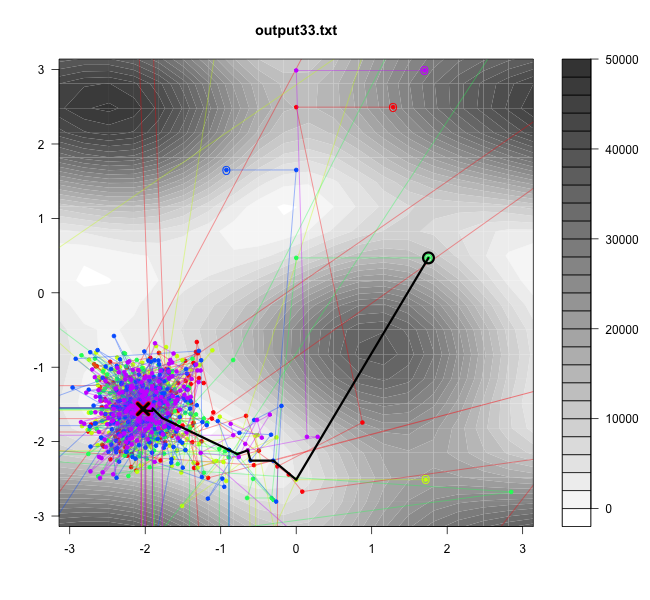}
	\includegraphics[width=0.19\columnwidth, trim={0.7cm 0 3.5cm 1.5cm},clip]{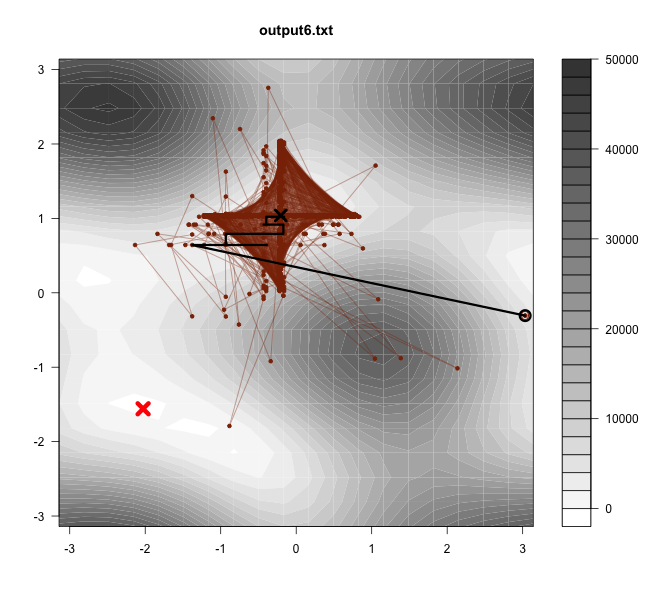}
	\includegraphics[width=0.19\columnwidth, trim={0.7cm 0 3.5cm 1.5cm},clip]{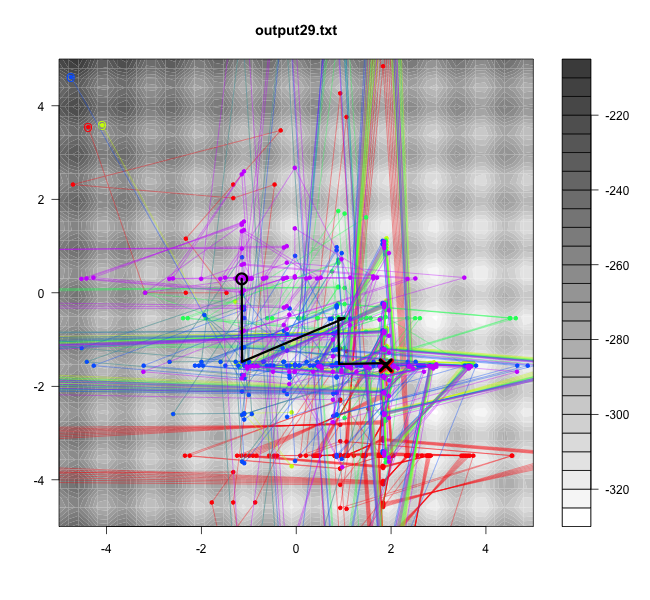}
	\includegraphics[width=0.19\columnwidth, trim={0.7cm 0 3.5cm 1.5cm},clip]{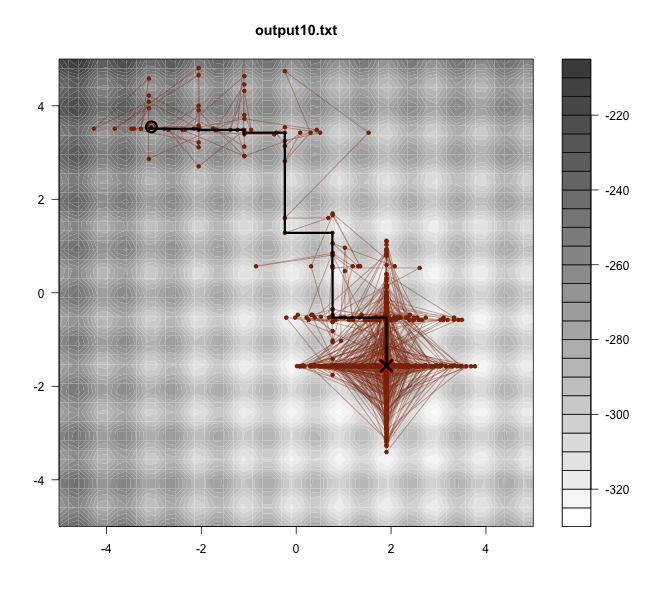}
	\includegraphics[width=0.19\columnwidth, trim={0.7cm 0 3.5cm 1.5cm},clip]{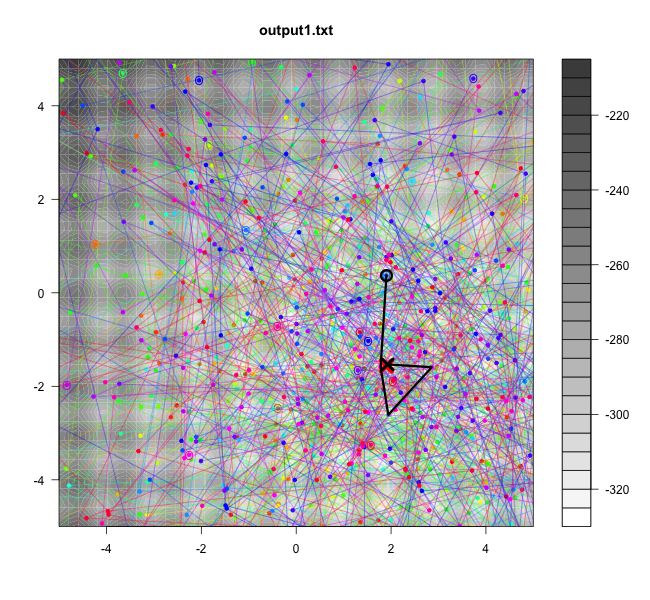}\\
	\vspace{-1mm}
	\includegraphics[width=0.19\columnwidth, trim={0.7cm 0 3.5cm 1.5cm},clip]{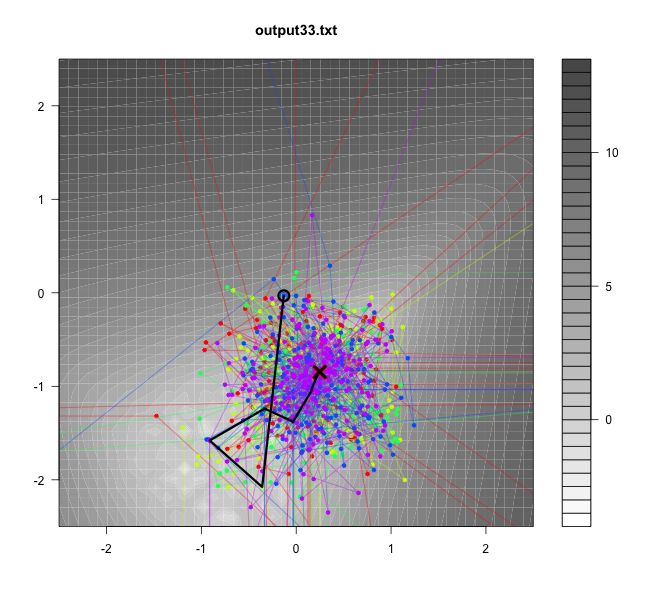}
	\includegraphics[width=0.19\columnwidth, trim={0.7cm 0 3.5cm 1.5cm},clip]{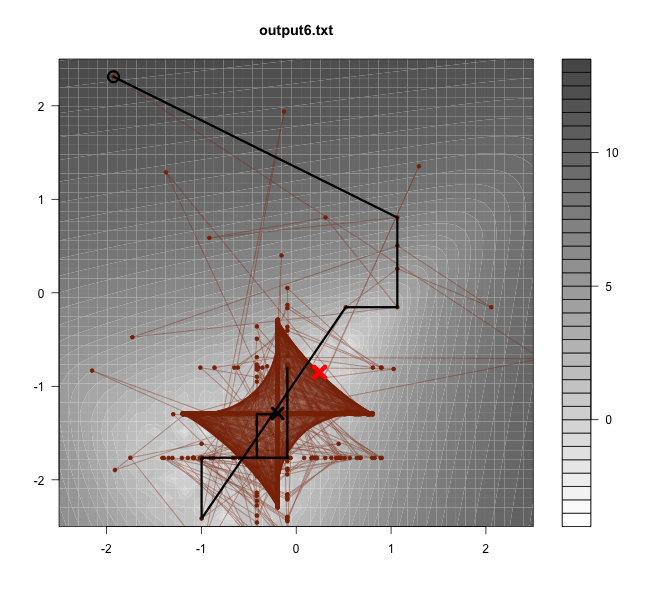}
	\includegraphics[width=0.19\columnwidth, trim={0.7cm 0 3.5cm 1.5cm},clip]{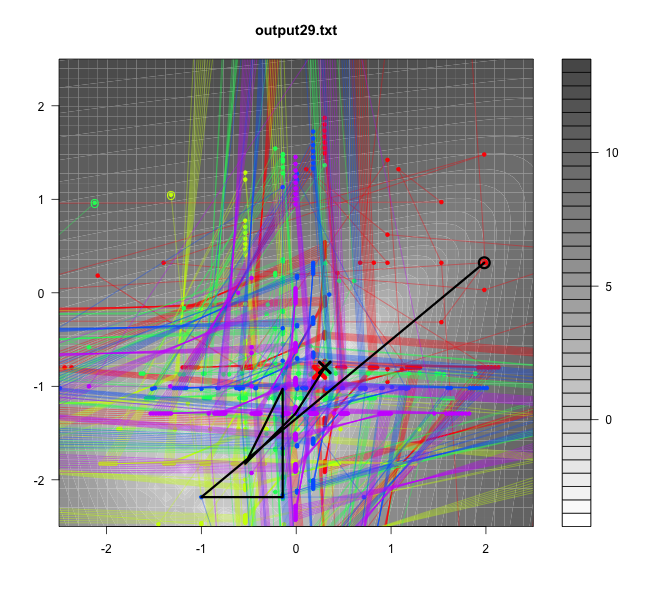}
	\includegraphics[width=0.19\columnwidth, trim={0.7cm 0 3.5cm 1.5cm},clip]{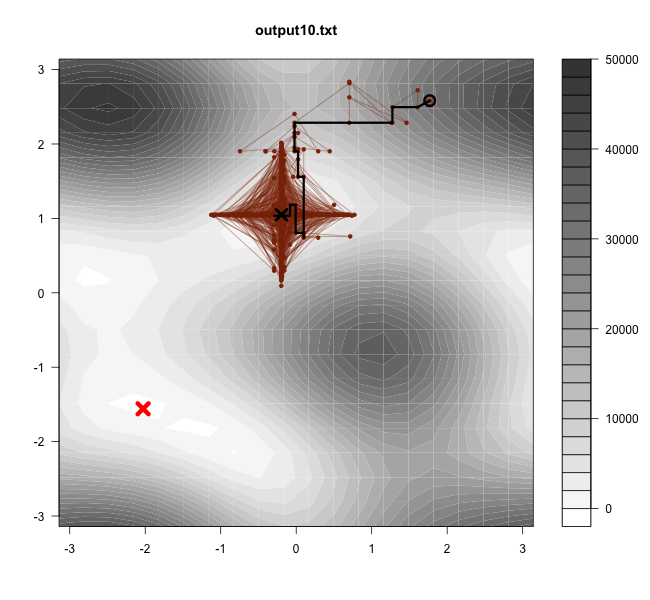}
	\includegraphics[width=0.19\columnwidth, trim={0.7cm 0 3.5cm 1.5cm},clip]{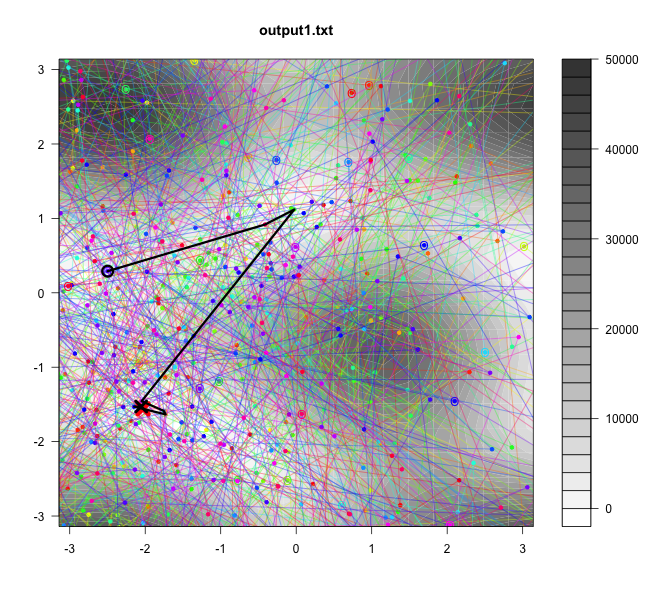}\\
	\vspace{-1mm}
	\includegraphics[width=0.19\columnwidth, trim={0.7cm 0 3.5cm 1.5cm},clip]{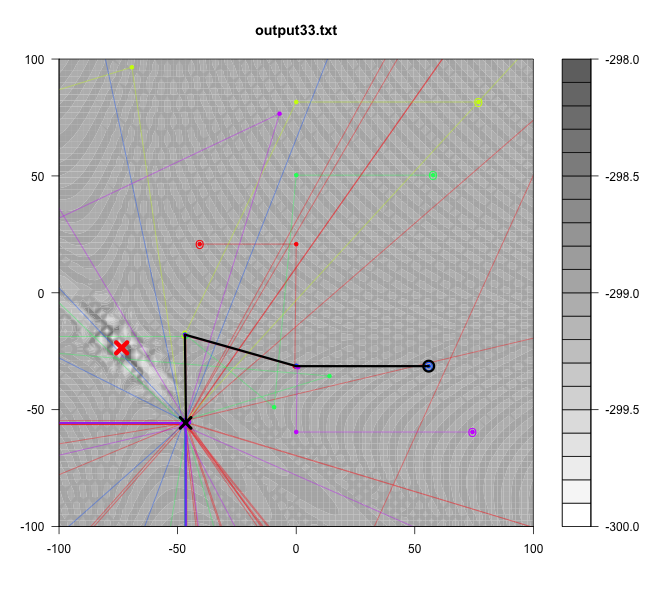}
	\includegraphics[width=0.19\columnwidth, trim={0.7cm 0 3.5cm 1.5cm},clip]{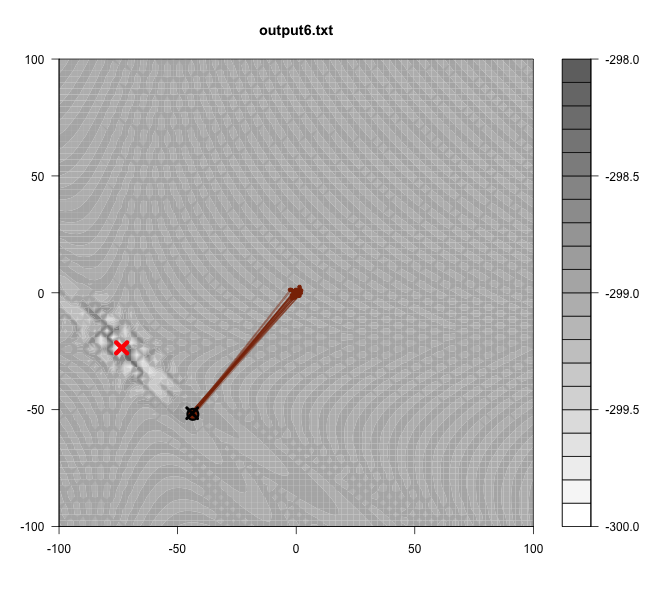}
	\includegraphics[width=0.19\columnwidth, trim={0.7cm 0 3.5cm 1.5cm},clip]{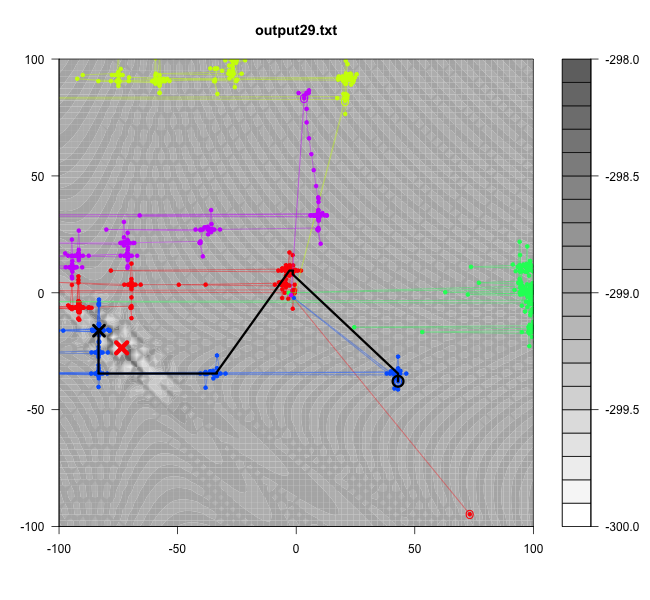}
	\includegraphics[width=0.19\columnwidth, trim={0.7cm 0 3.5cm 1.5cm},clip]{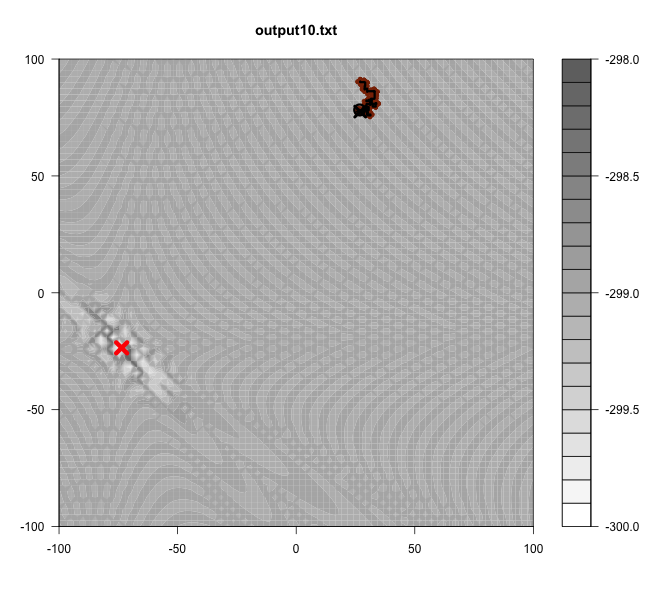}
	\includegraphics[width=0.19\columnwidth, trim={0.7cm 0 3.5cm 1.5cm},clip]{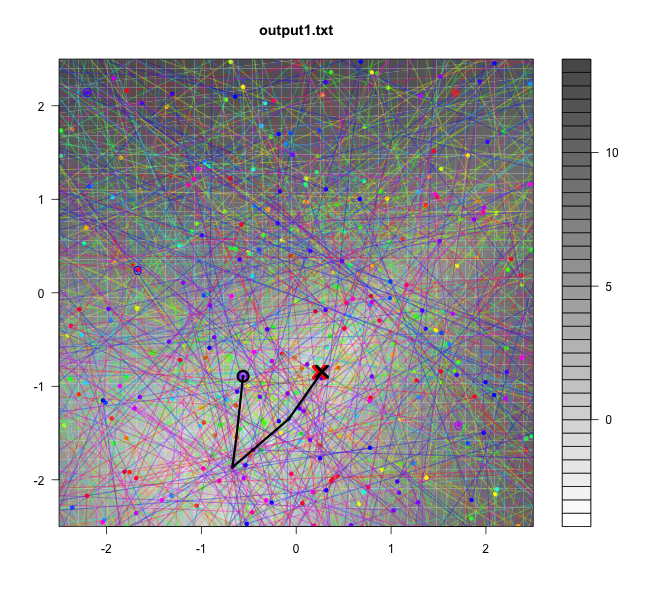}\\
	\vspace{-1mm}
	\includegraphics[width=0.19\columnwidth, trim={0.7cm 0 3.5cm 1.5cm},clip]{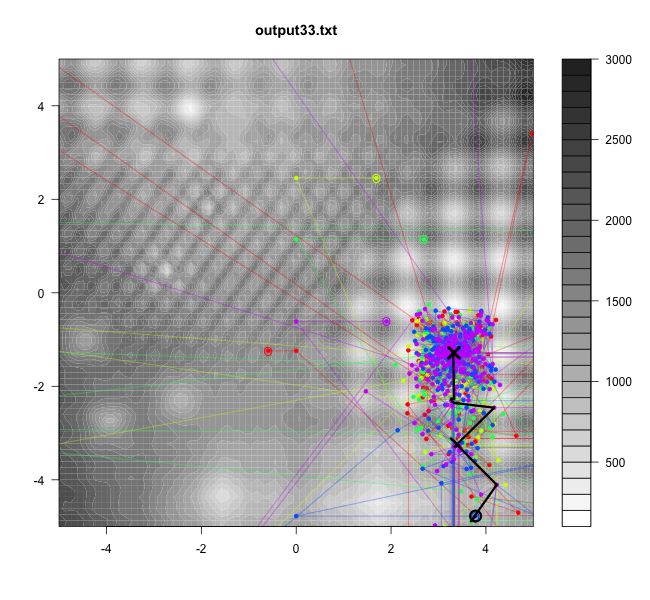}
	\includegraphics[width=0.19\columnwidth, trim={0.7cm 0 3.5cm 1.5cm},clip]{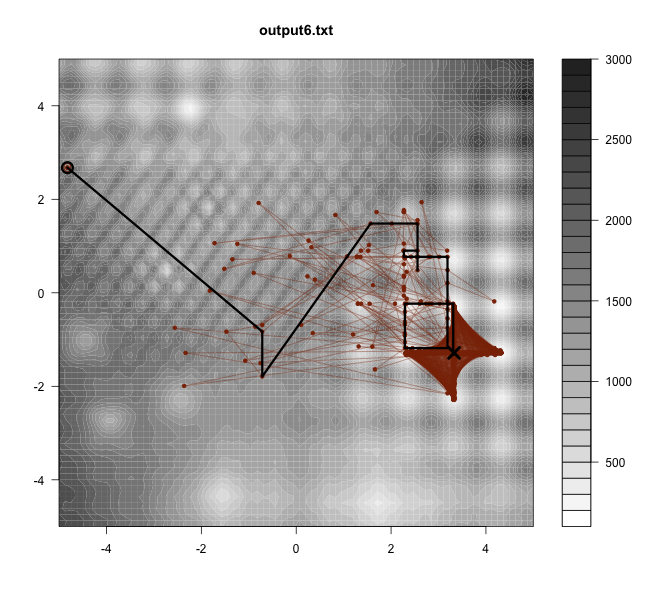}
	\includegraphics[width=0.19\columnwidth, trim={0.7cm 0 3.5cm 1.5cm},clip]{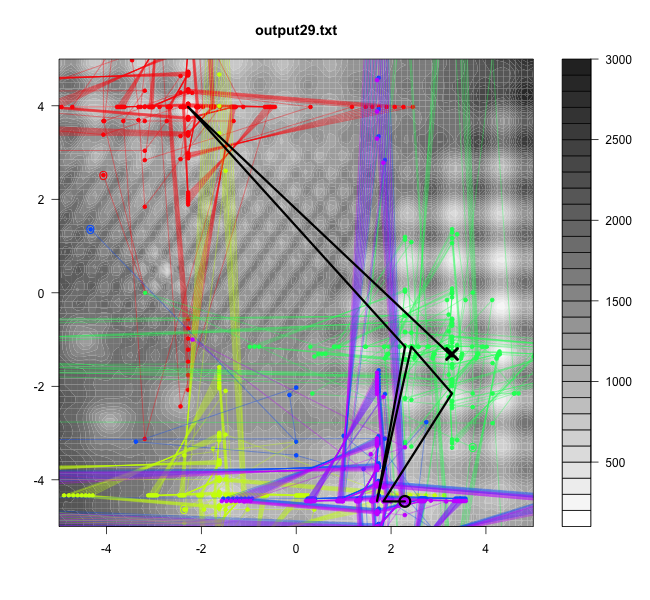}
	\includegraphics[width=0.19\columnwidth, trim={0.7cm 0 3.5cm 1.5cm},clip]{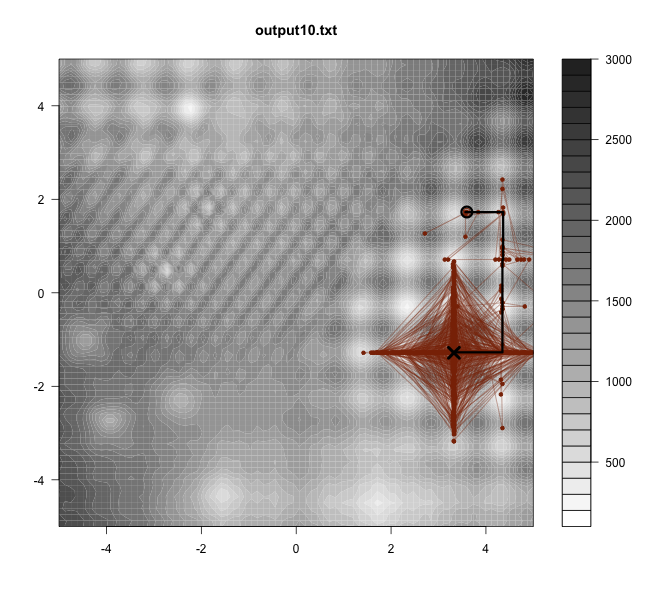}
	\includegraphics[width=0.19\columnwidth, trim={0.7cm 0 3.5cm 1.5cm},clip]{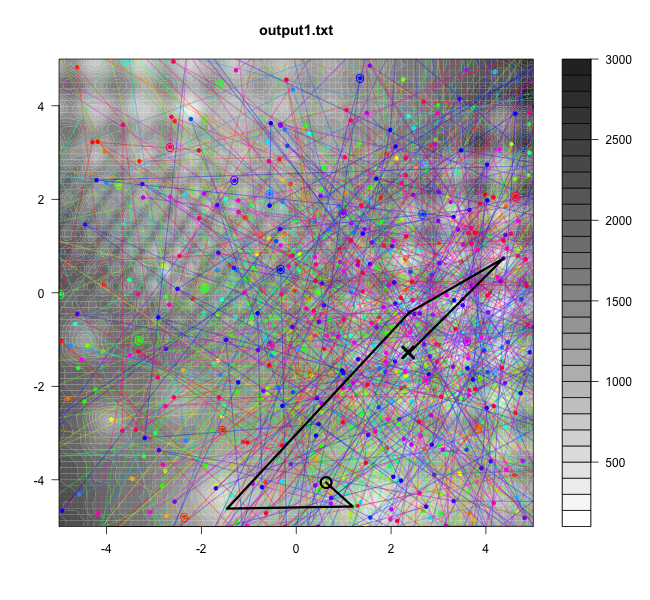}\\
	\vspace{-1mm}
	\includegraphics[width=0.19\columnwidth, trim={0.7cm 0 3.5cm 1.5cm},clip]{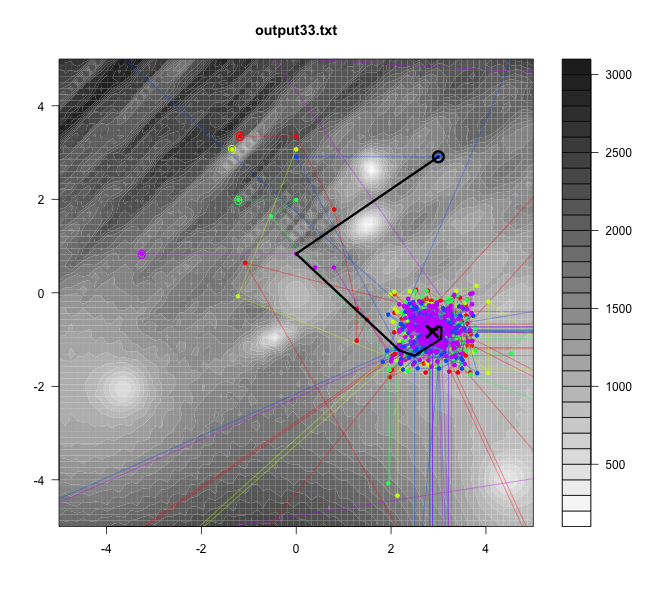}
	\includegraphics[width=0.19\columnwidth, trim={0.7cm 0 3.5cm 1.5cm},clip]{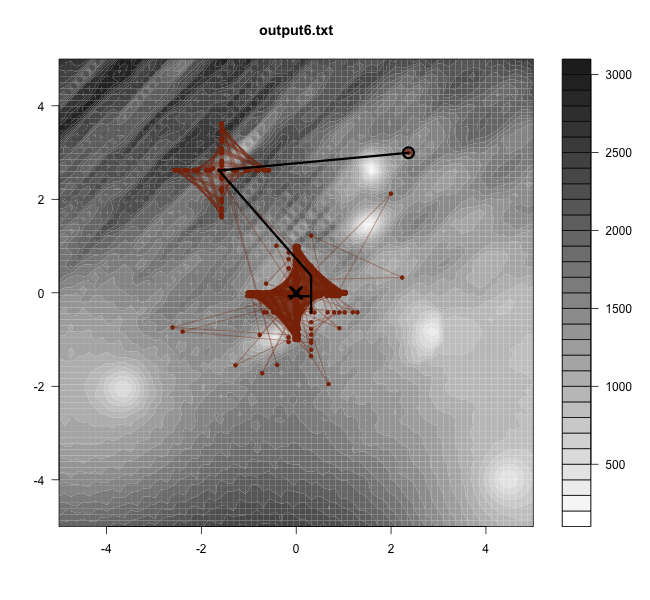}
	\includegraphics[width=0.19\columnwidth, trim={0.7cm 0 3.5cm 1.5cm},clip]{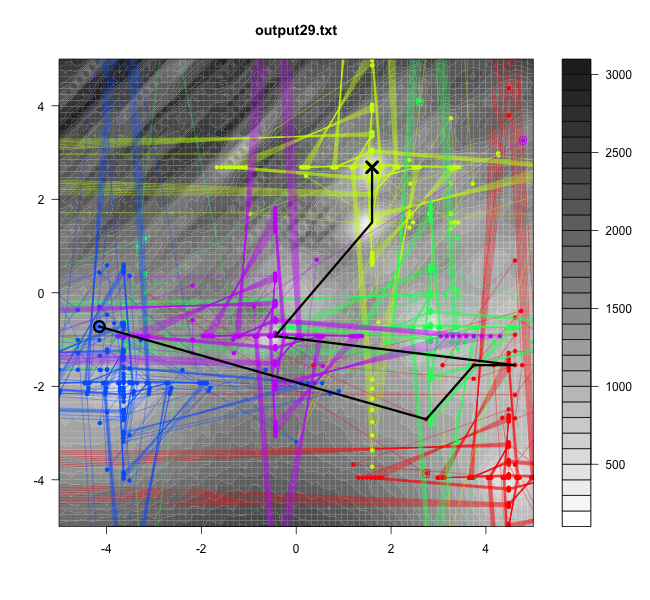}
	\includegraphics[width=0.19\columnwidth, trim={0.7cm 0 3.5cm 1.5cm},clip]{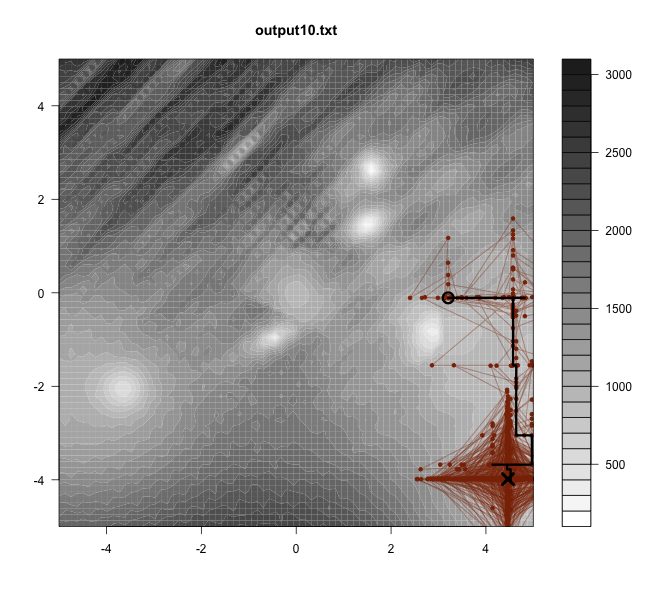}
	\includegraphics[width=0.19\columnwidth, trim={0.7cm 0 3.5cm 1.5cm},clip]{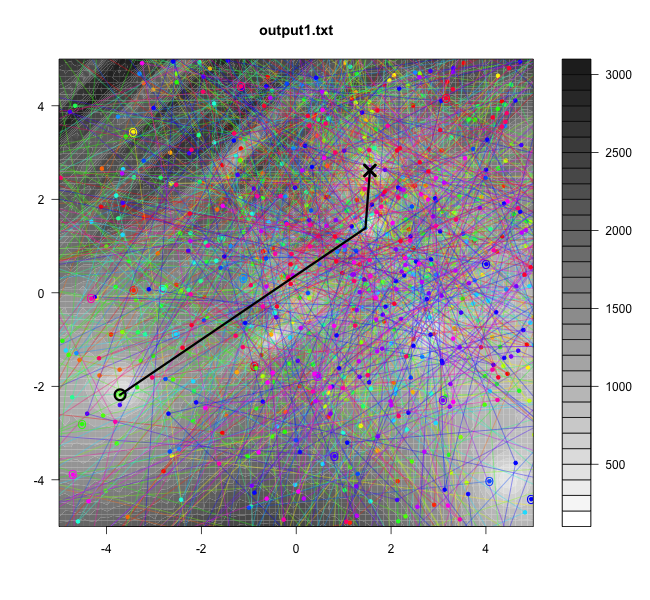}\\
	\vspace{-1mm}
	\includegraphics[width=0.19\columnwidth, trim={0.7cm 0 3.5cm 1.5cm},clip]{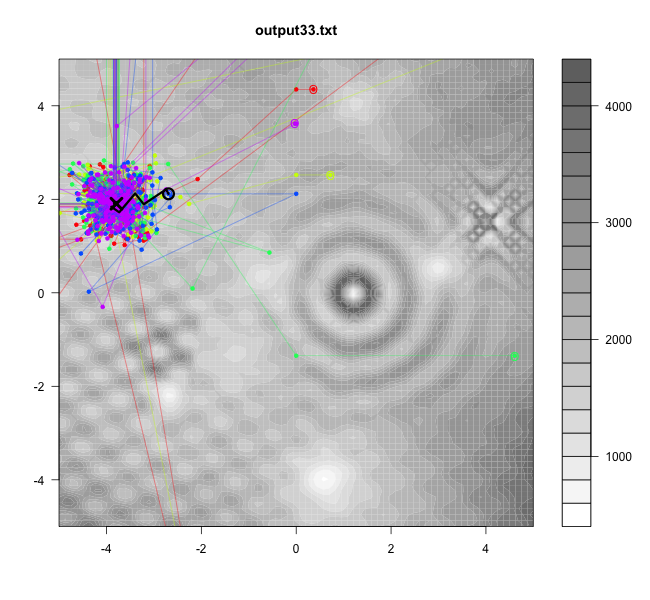}
	\includegraphics[width=0.19\columnwidth, trim={0.7cm 0 3.5cm 1.5cm},clip]{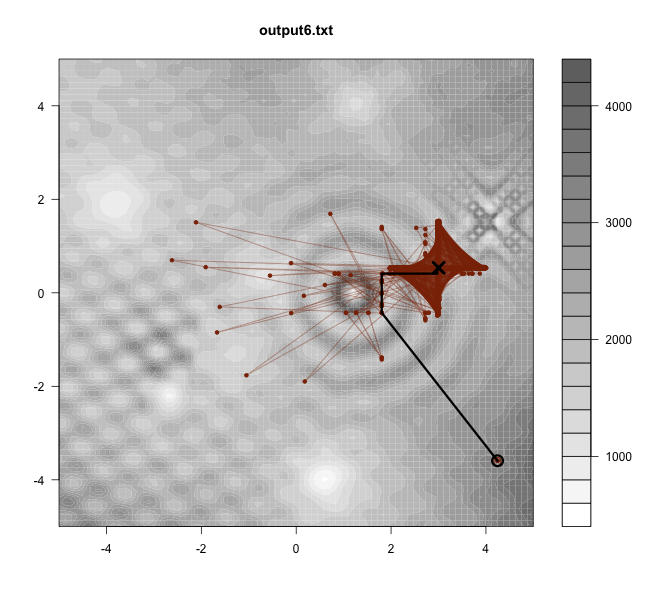}
	\includegraphics[width=0.19\columnwidth, trim={0.7cm 0 3.5cm 1.5cm},clip]{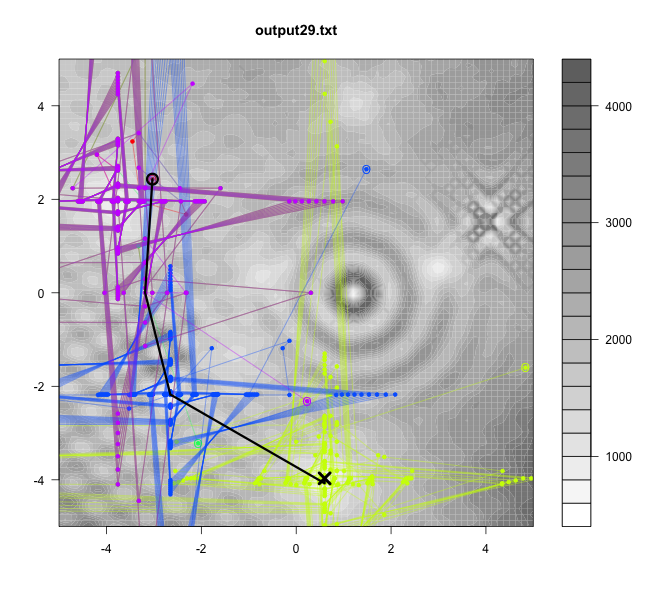}
	\includegraphics[width=0.19\columnwidth, trim={0.7cm 0 3.5cm 1.5cm},clip]{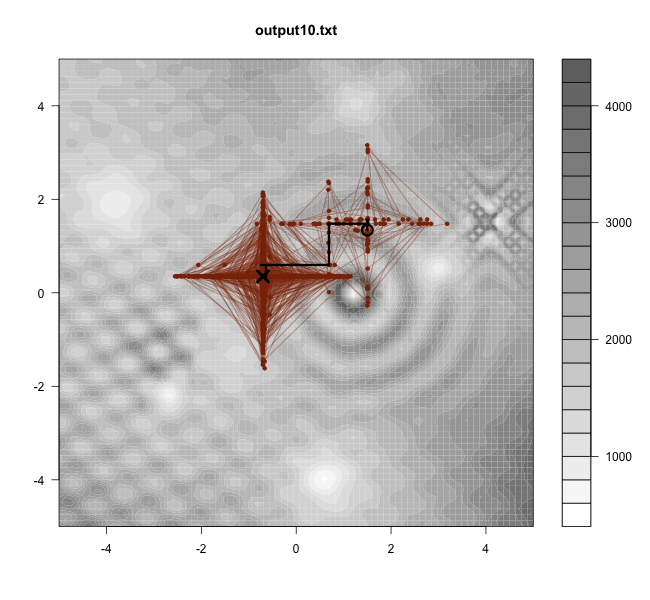}
	\includegraphics[width=0.19\columnwidth, trim={0.7cm 0 3.5cm 1.5cm},clip]{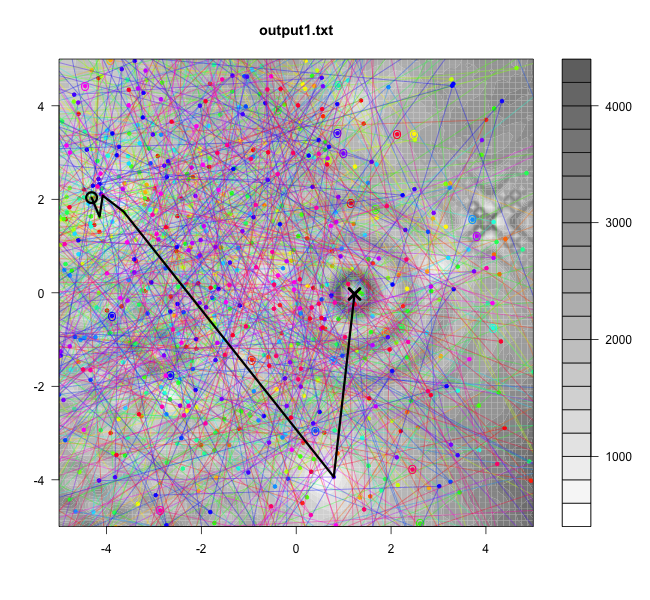}\\
	\vspace{-1mm}
	\includegraphics[width=0.19\columnwidth, trim={0.7cm 0 3.5cm 1.5cm},clip]{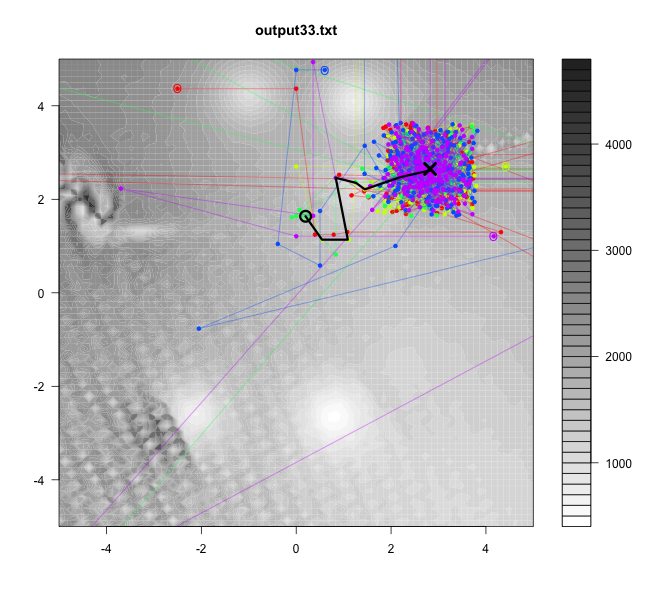}
	\includegraphics[width=0.19\columnwidth, trim={0.7cm 0 3.5cm 1.5cm},clip]{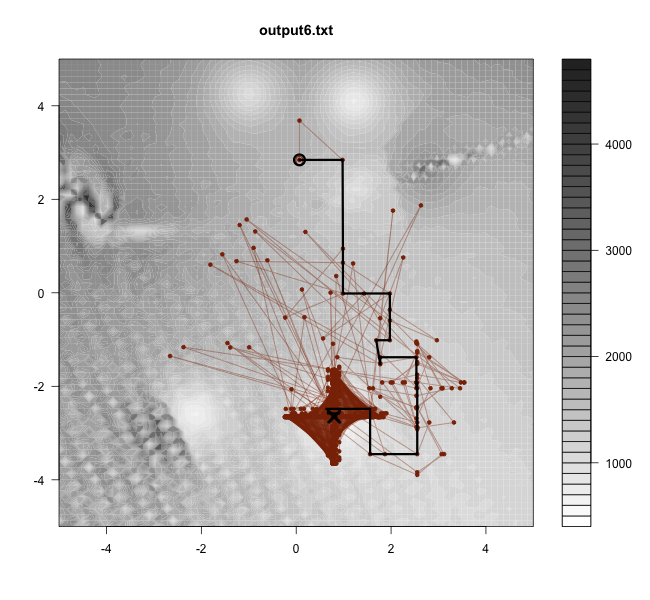}
	\includegraphics[width=0.19\columnwidth, trim={0.7cm 0 3.5cm 1.5cm},clip]{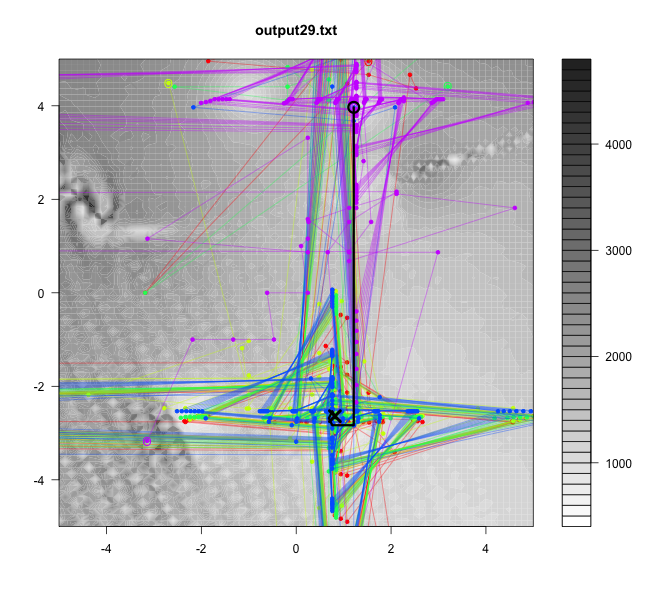}
	\includegraphics[width=0.19\columnwidth, trim={0.7cm 0 3.5cm 1.5cm},clip]{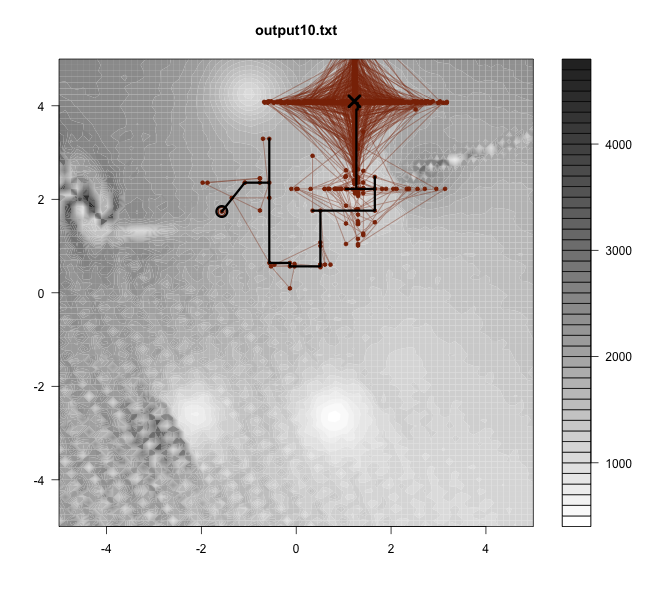}
	\includegraphics[width=0.19\columnwidth, trim={0.7cm 0 3.5cm 1.5cm},clip]{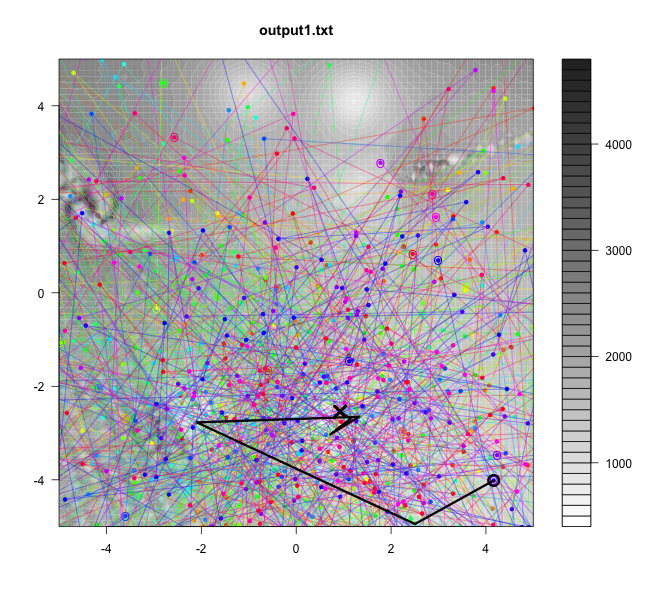}\vspace{-1mm}\\
	
	\caption{Examples of each best optimiser (left to right: $F_1$, $F_9$, $F_{12}$, $F_{13}$, $F_{14}$) applied to each of the other functions (top to bottom in numerical order).}
	\label{fig:generality}
\end{figure}

\paragraph{$F_1$ best optimiser} This optimiser uses a swarm of five parallel search processes. At each iteration, each of these moves to a position relative to the swarm best (i.e the best point seen so far by all swarm members). In each case, this is done by adding a random vector (\code{vector.wrand}) to the swarm best. The size of this random vector is determined using a trigonometric expression based on the value of certain dimensions of the swarm member's current and best search points. This means that the move size carried out by each swarm member at each iteration is different, which leads to the generation of move sizes over multiple scales. For the $F_1$ landscape, the small moves are visible as the trajectory approaches the optima, and the large moves can be seen by the radiating coloured lines. %The smaller move sizes also appear useful in the other function landscapes.
	
\paragraph{$F_9$ best optimiser} This is a local optimiser with only one point of search. It continually switches between searching around the best-seen search point and evaluating a random search point. When searching around the best-seen point, at each iteration it adds the sine of the iteration number to a single dimension of the vector, moving along two dimensions each time. In essence, the periodic nature of the sine function causes the trajectory to systematically explore the nearby search space, which can be seen by the space-filling patterns centred around the final point of search in Fig. \ref{fig:bests}. This use of trigonometric functions appears to be relatively common amongst evolved optimisers, as reflected in Table \ref{tab:usage}, which lists the 20 most frequent instructions found for each of the four swarm size $\times$ iterations configurations.

\begin{figure}[htb!]
	\centering
	\subfloat[][4: 50 steps] {
		\includegraphics[width=0.22\columnwidth, trim={0.7cm 0 3.5cm 1.5cm},clip]{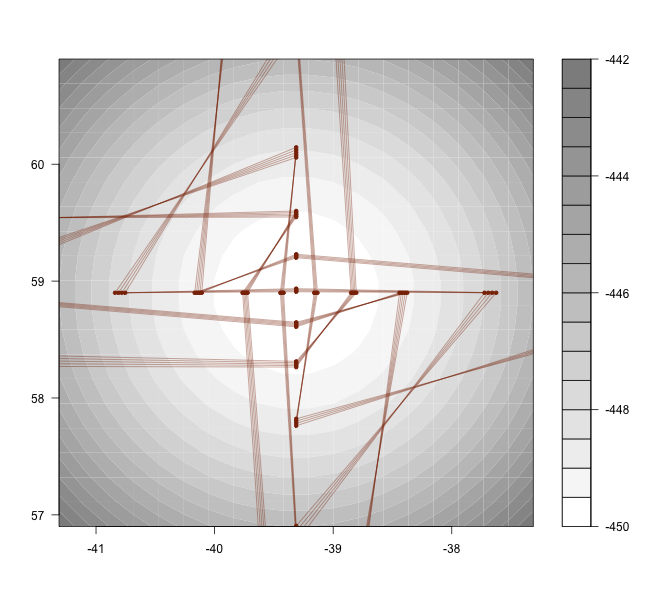}
		\label{fig:f12neighbourhood:2}
	}
	\subfloat[][14: 350 \& 700 steps] {
		\includegraphics[width=0.22\columnwidth, trim={0.7cm 0 3.5cm 1.5cm},clip]{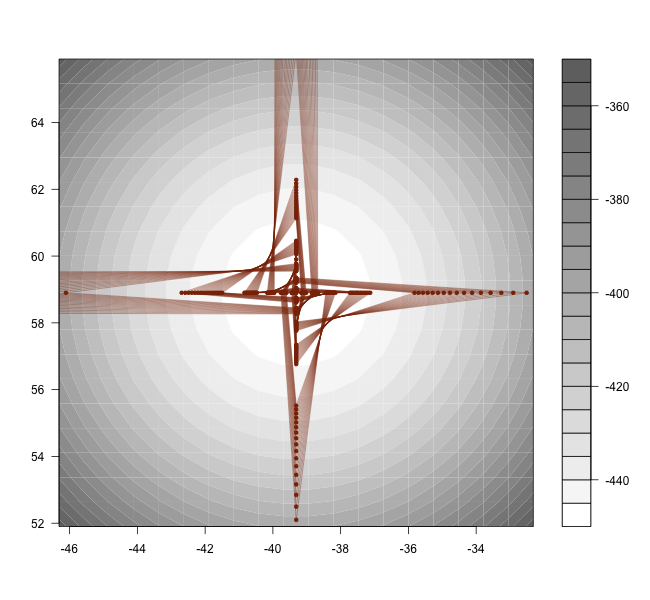}
		\includegraphics[width=0.22\columnwidth, trim={0.7cm 0 3.5cm 1.5cm},clip]{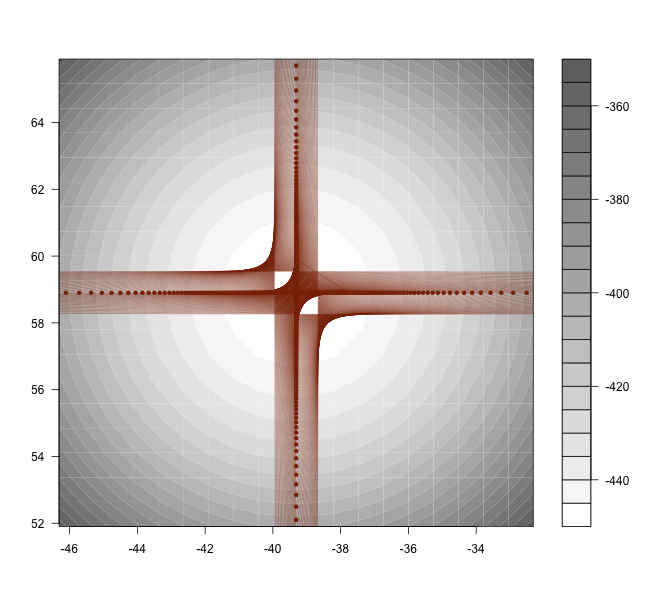}
		\label{fig:f12neighbourhood:7}
	}
	\subfloat[][200: 1000 steps] {
		\includegraphics[width=0.22\columnwidth, trim={0.7cm 0 3.5cm 1.5cm},clip]{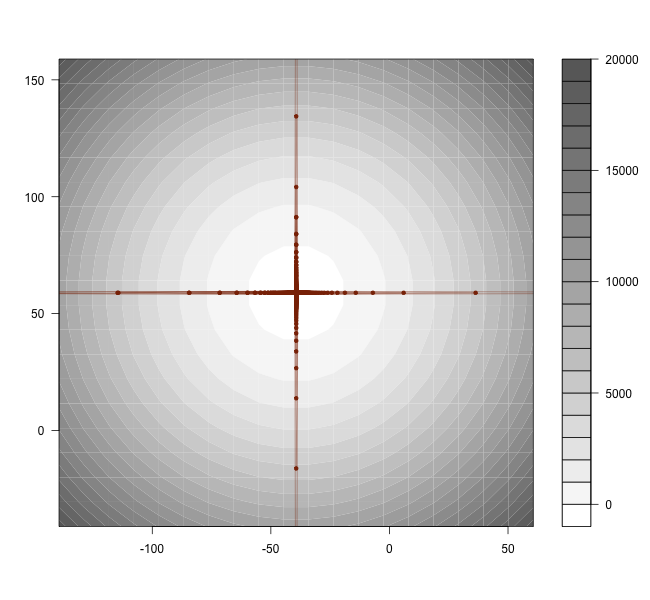}
		\label{fig:f12neighbourhood:100}
	}
	\caption{Search neighbourhoods of the $F_{12}$ best optimiser, showing how the scale-free pattern develops over search spaces with bounds of size 4, 14 \& 200.}
	\label{fig:f12neighbourhood}
\end{figure}

\paragraph{$F_{12}$ best optimiser} This optimiser is the most complex at the instruction level, and it uses each swarm member's index, the index (but not the vector) of the current best swarm member, and both the improvement and out-of-bounds Boolean signals to determine each move. Notably, each swarm member uses a deterministic geometric pattern to explore the space around the current search point. This is depicted in Fig. \ref{fig:f12neighbourhood}, showing how it causes the optimiser to explore neighbourhoods that lie at powers-of-two distance from its centre. Once a neighbourhood has been seeded, it progressively grows outwards, until it eventually meets the surrounding neighbourhoods. The sampling rate for a neighbourhood is proportional to its distance from the centre, meaning that closer regions are explored with greater granularity, and the neighbourhoods are seeded progressively, meaning that it only searches further out if the nearer neighbourhoods are not productive. This pattern causes it to explore local neighbourhoods over a number of scales, enabling it to search within landscapes that are considerably larger than the one it was trained on, such as $F_1$ and $F_{14}$.

\begin{figure}[tb!]
	\centering
	
	\includegraphics[width=0.32\columnwidth, trim={0.7cm 0 3.5cm 1.5cm},clip]{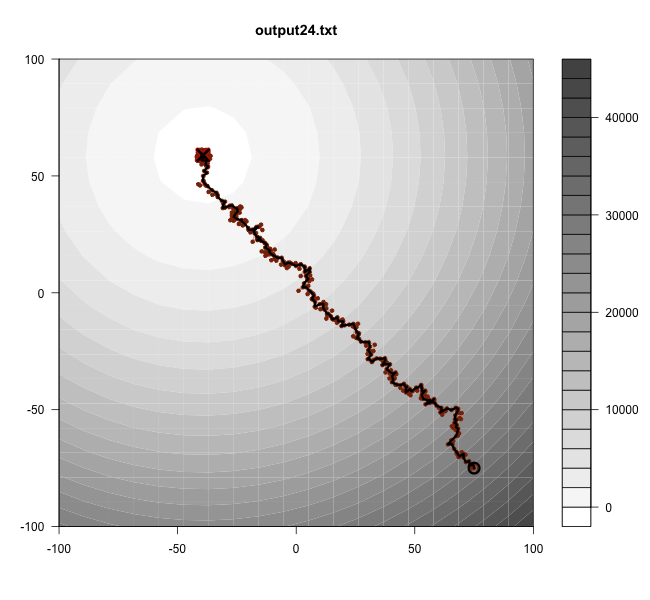}
	\includegraphics[width=0.32\columnwidth, trim={0.7cm 0 3.5cm 1.5cm},clip]{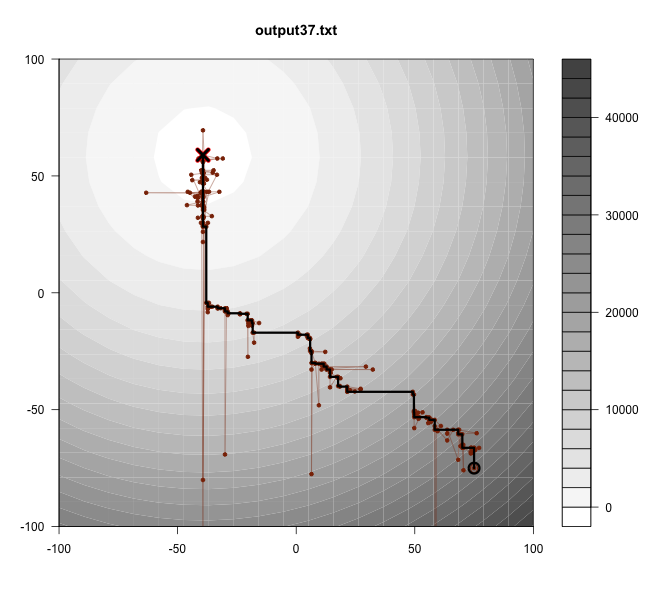}
	\includegraphics[width=0.32\columnwidth, trim={0.7cm 0 3.5cm 1.5cm},clip]{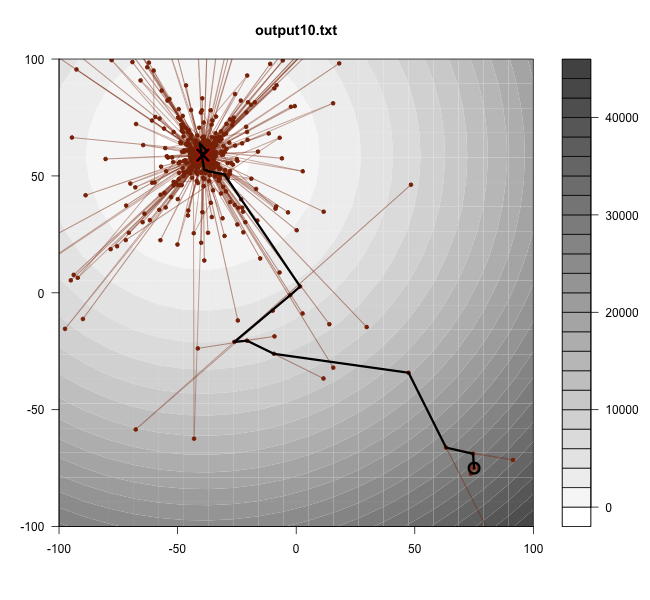}\\
	\includegraphics[width=0.32\columnwidth, trim={0.7cm 0 3.5cm 1.5cm},clip]{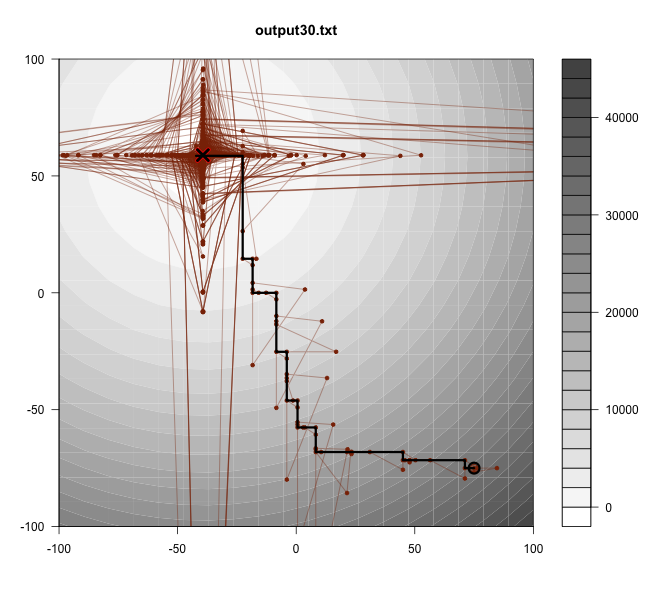}
	\includegraphics[width=0.32\columnwidth, trim={0.7cm 0 3.5cm 1.5cm},clip]{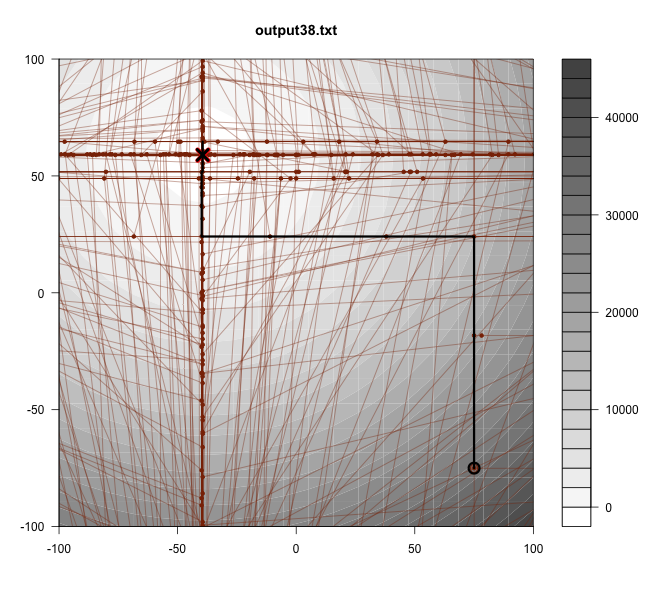}
	\includegraphics[width=0.32\columnwidth, trim={0.7cm 0 3.5cm 1.5cm},clip]{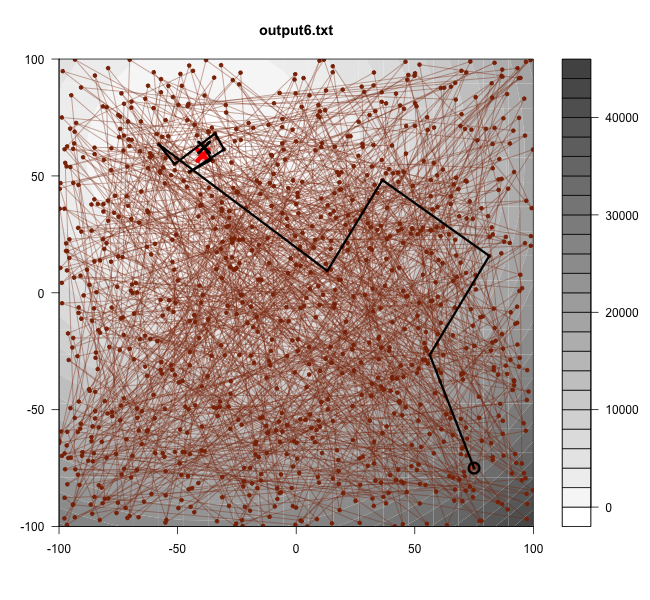}
	
	\caption{Examples of different local optimisers trained on the $F_1$ function, showing some of the diversity of solutions found between runs. All trajectories start at the same point.}
	\label{fig:diversity_f1}
\end{figure}

\paragraph{$F_{13}$ best optimiser} This optimiser, by comparison, has the simplest program. It is a local optimiser and, at each iteration, it adds a random value to one of the dimensions of the best-seen search point, cycling through the dimensions on each subsequent move (hence why it generates a cross-shaped trajectory in 2D). The size of each move is determined by both the sine of the objective value of the current point and the sine of the largest dimension of the search space (accessed using the \code{input.inall} instruction, which is commonly used in the local optimisers). The former causes the move size to vary cyclically as search progresses, and the latter allows it to adapt the move size to the landscape size, at least to an extent. Although this optimiser only (with the exception of the first move) explores moves in one dimension at a time, it still works quite well on the non-separable landscapes.

\paragraph{$F_{14}$ best optimiser} This optimiser is the only one of the five which uses both a larger swarm size (of 25) and the \code{vector.between} instruction. Each iteration, each swarm member uses this instruction to generate a new search point half-way between the swarm best and one of its own previous positions, which are all retained within the vector stack. A small random vector is then added to each half-way point, presumably to inject diversity. Importantly, which previous position is used for a particular swarm member is determined by the swarm member's index: the first swarm member uses its current position, and higher numbered swarm members go back further in time. This allows a backtracking behaviour within the swarm, presumably useful for landscapes, such as $F_{14}$, that are deceptive and have limited gradient information. From Table \ref{tab:usage}, it can be seen that the usage of \code{vector.between} generally increases as the swarm size increases, perhaps reflecting pressure to carry out more coordinated behaviour in larger swarms in order to counteract the smaller number of iterations.

\paragraph{}A general observation from Fig. \ref{fig:generality} is that the optimisers produce visibly different trajectories on landscapes that have very different search bounds to the function they were trained on. For instance, the $F_9$ and $F_{13}$ optimisers both carry out much smaller moves when applied to the $F_1$ and $F_{14}$ landscapes, and the $F_{14}$ optimiser carries out much larger moves on the landscapes with smaller bounds. This suggests that they may have overfit the size of the landscapes to some degree, despite the use of random instance scalings during training, and there may be some benefit to using either using a wider range of scalings, or normalising the search bounds. However, normalising search bounds may be problematic for real-world problems where the region containing the optima is not known.

\afterpage{\FloatBarrier}

\subsection{Diversity of Evolved Optimisers}

\begin{figure}[tbp!]
	%\centering
	
	\subfloat[][Swarm size of 5, trained on $F_{12}$] {
		\begin{tabular}[b]{c}% since line breaks don't work in subfloat
			\includegraphics[width=0.32\columnwidth, trim={0.7cm 0 3.5cm 1.5cm},clip]{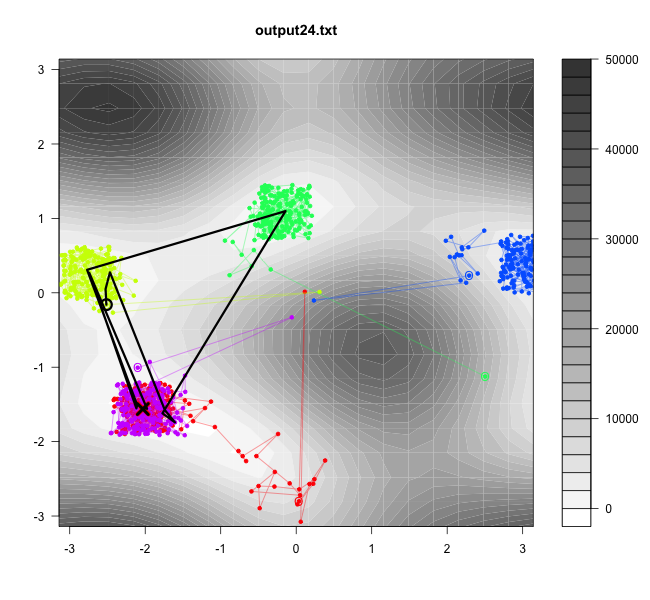}
			\includegraphics[width=0.32\columnwidth, trim={0.7cm 0 3.5cm 1.5cm},clip]{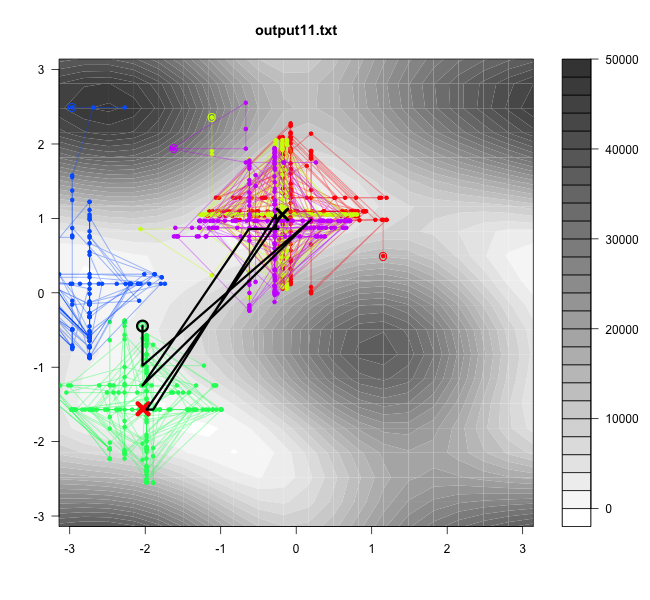}
			\includegraphics[width=0.32\columnwidth, trim={0.7cm 0 3.5cm 1.5cm},clip]{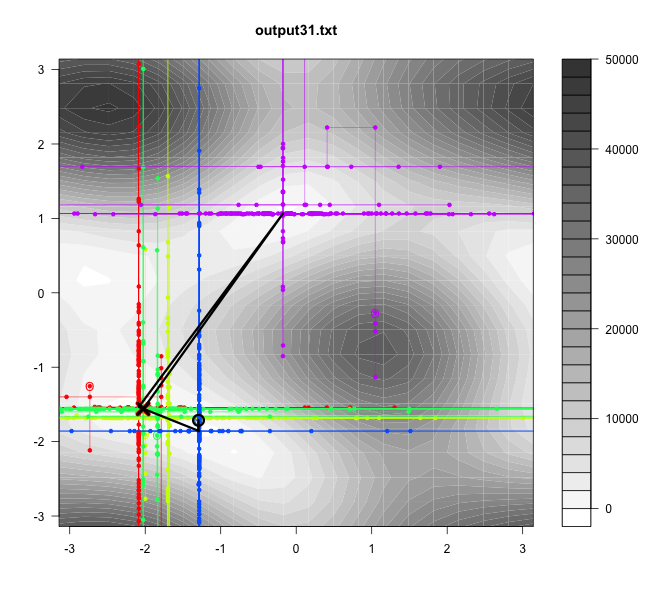}\\
			\includegraphics[width=0.32\columnwidth, trim={0.7cm 0 3.5cm 1.5cm},clip]{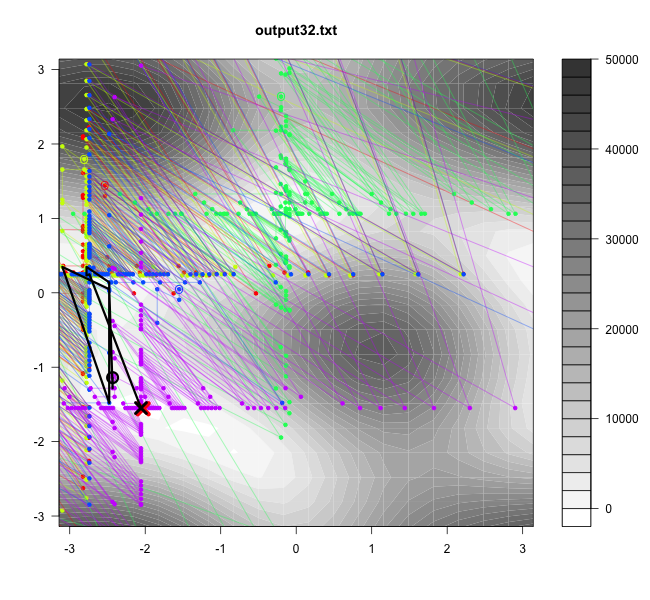}
			\includegraphics[width=0.32\columnwidth, trim={0.7cm 0 3.5cm 1.5cm},clip]{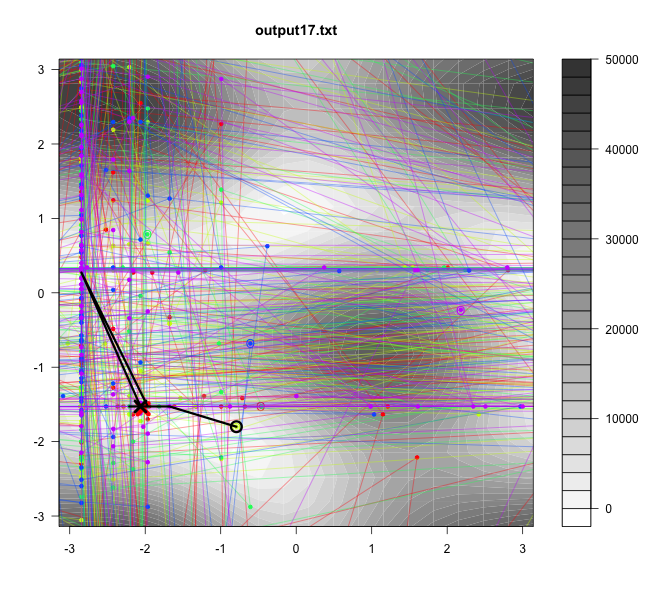}
			\includegraphics[width=0.32\columnwidth, trim={0.7cm 0 3.5cm 1.5cm},clip]{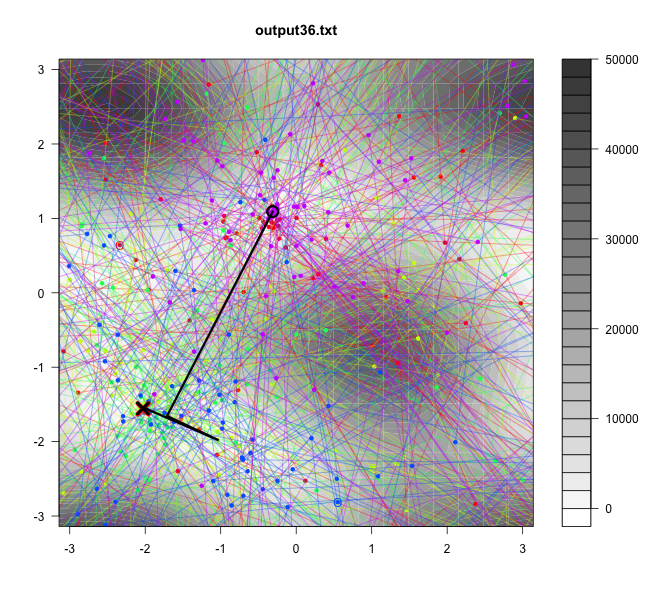}
		\end{tabular}
		\label{fig:diversity::f12}
	}
	
	\subfloat[][Swarm size of 50, trained on $F_{14}$] {
		\begin{tabular}[b]{c}%
			\includegraphics[width=0.32\columnwidth, trim={0.7cm 0 3.5cm 1.5cm},clip]{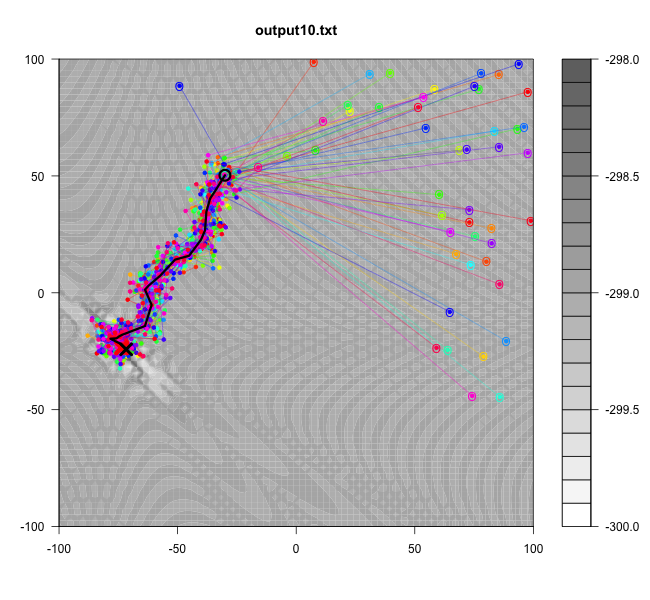}
			\includegraphics[width=0.32\columnwidth, trim={0.7cm 0 3.5cm 1.5cm},clip]{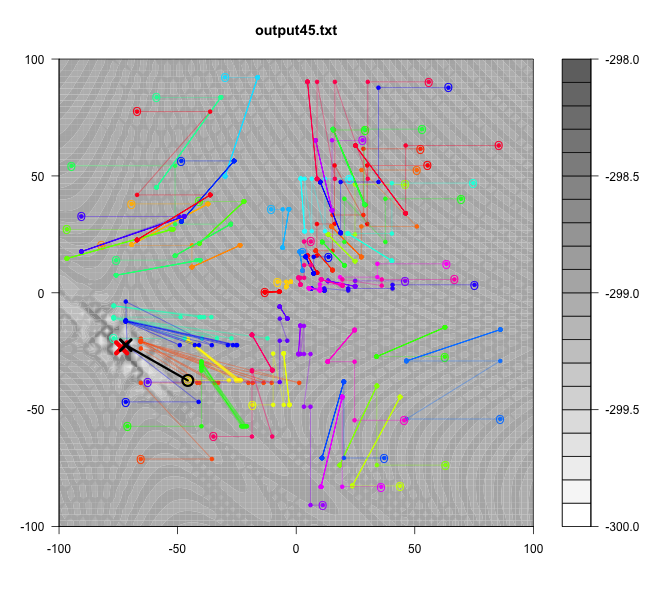}
			\includegraphics[width=0.32\columnwidth, trim={0.7cm 0 3.5cm 1.5cm},clip]{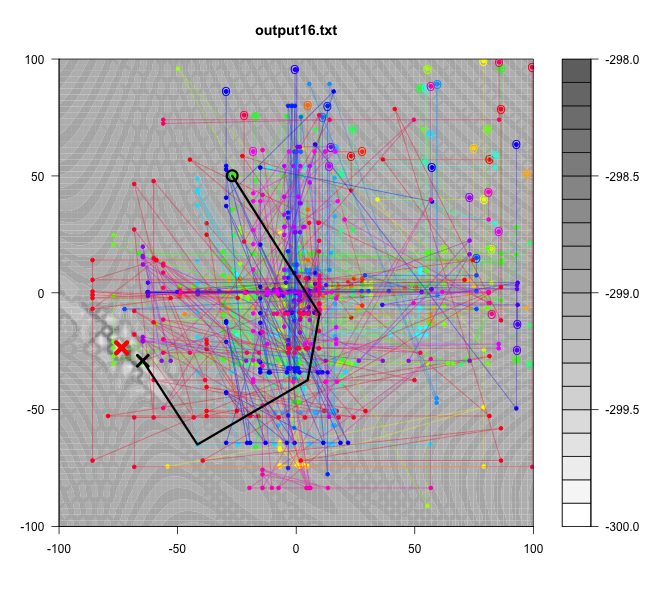}\\
			\includegraphics[width=0.32\columnwidth, trim={0.7cm 0 3.5cm 1.5cm},clip]{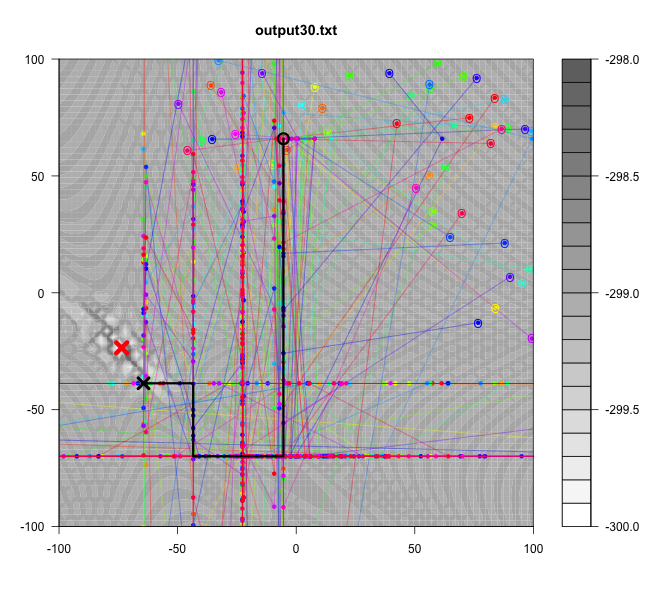}
			\includegraphics[width=0.32\columnwidth, trim={0.7cm 0 3.5cm 1.5cm},clip]{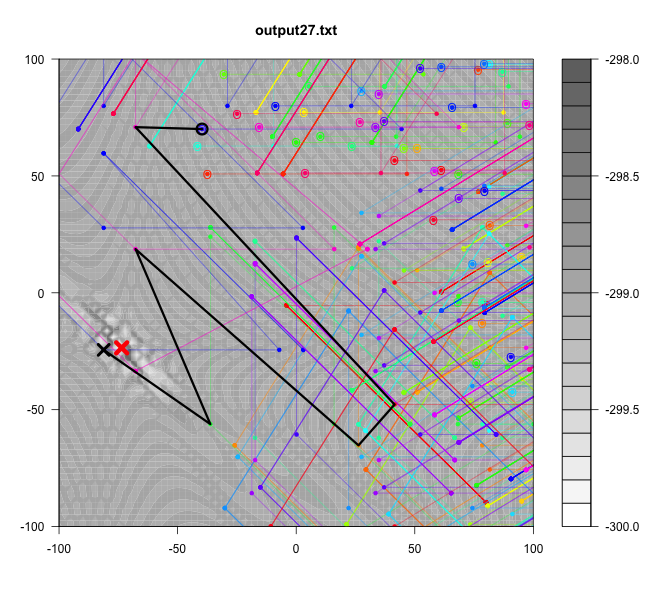}
			\includegraphics[width=0.32\columnwidth, trim={0.7cm 0 3.5cm 1.5cm},clip]{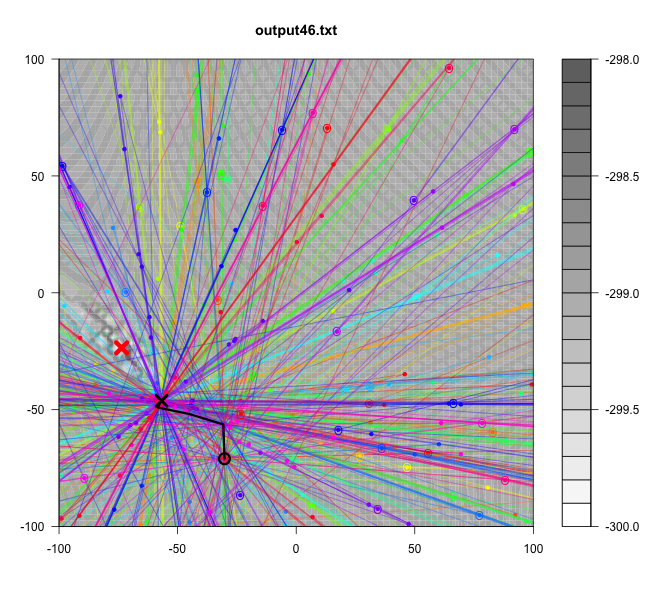}
		\end{tabular}
		\label{fig:diversity::f13}
	}
	
	\caption{Examples of different population-based optimisers.}
	\label{fig:diversity_pop}
\end{figure}

Due to the stochastic nature of EAs, it is common for different solutions to be found in different runs. Figs. \ref{fig:diversity_f1} and \ref{fig:diversity_pop} illustrate some of the diversity found between runs in these experiments. Fig. \ref{fig:diversity_f1} shows trajectories for a selection of best-in-run 1$\times$1000 optimisers trained on the $F_1$ function. These optimisers show significant variation in both the pattern of moves they carry out and the size of moves, highlighting that substantial inter-run diversity is present even for local optimisers trained on a unimodal landscape. Fig. \ref{fig:diversity_pop} shows that considerable diversity is also present amongst the population-based optimisers. These plots suggest that interesting ideas about how to do optimisation could be gained by looking more closely at the diverse solutions found between runs.

\begin{figure}[!p]
	\centering
	\includegraphics[width=\columnwidth, trim={0 0.6cm 0 0},clip]{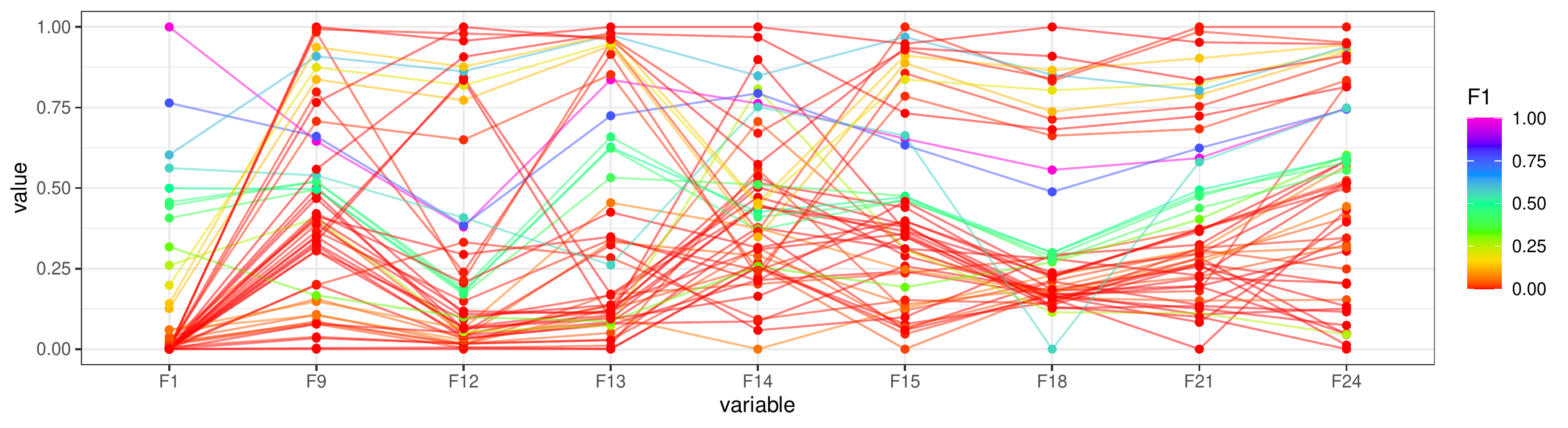}\vspace{2mm}
	\includegraphics[width=\columnwidth, trim={0 0.6cm 0 0},clip]{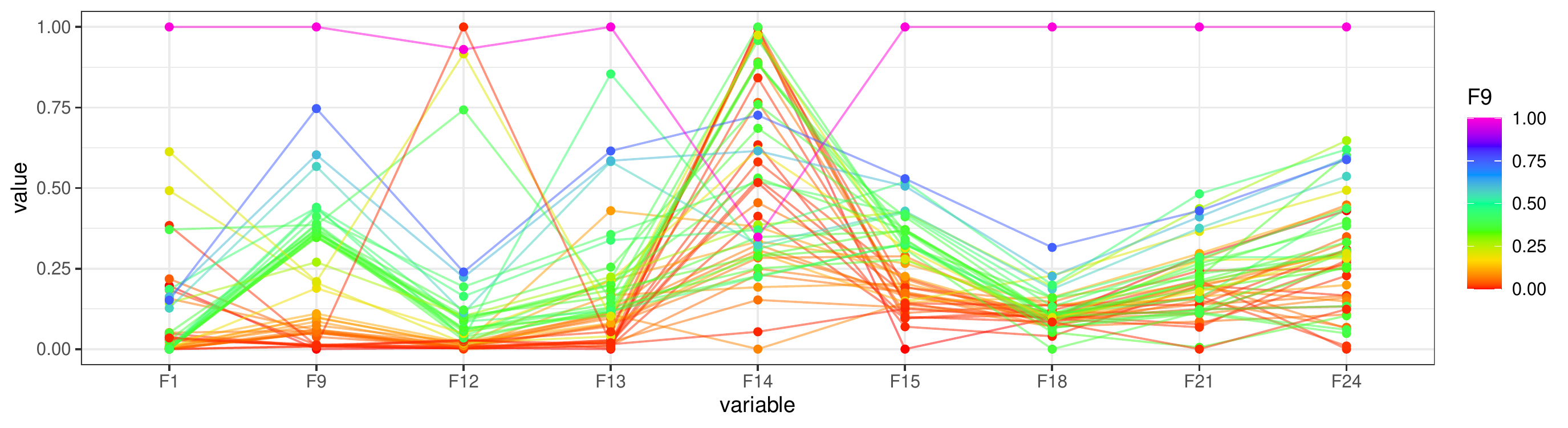}\vspace{2mm}
	\includegraphics[width=\columnwidth, trim={0 0.6cm 0 0},clip]{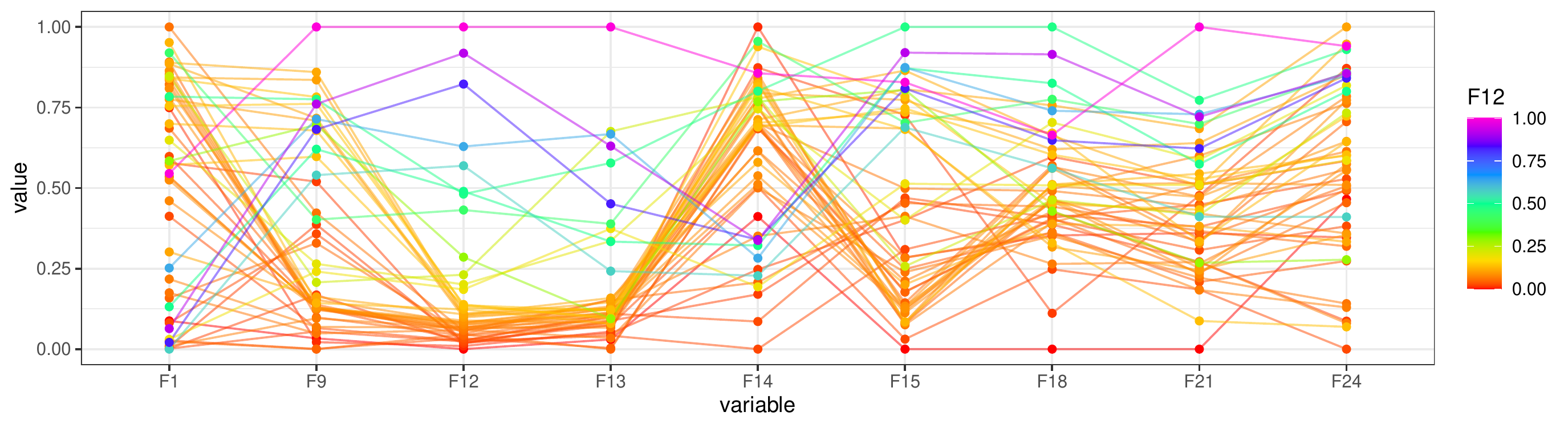}\vspace{2mm}
	\includegraphics[width=\columnwidth, trim={0 0.6cm 0 0},clip]{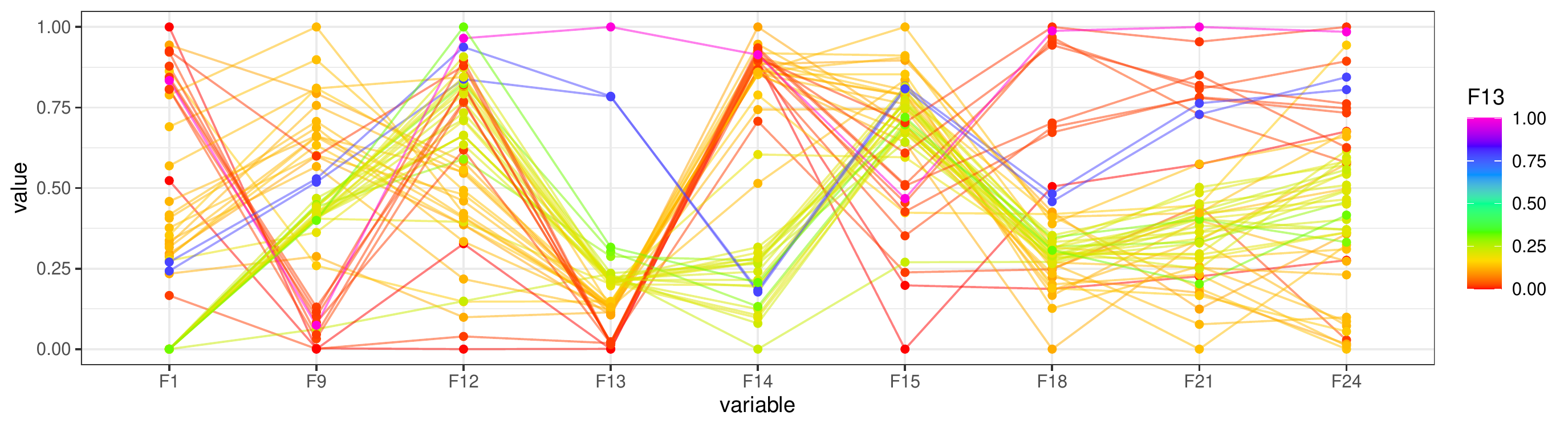}\vspace{2mm}
	\includegraphics[width=\columnwidth, trim={0 0.6cm 0 0},clip]{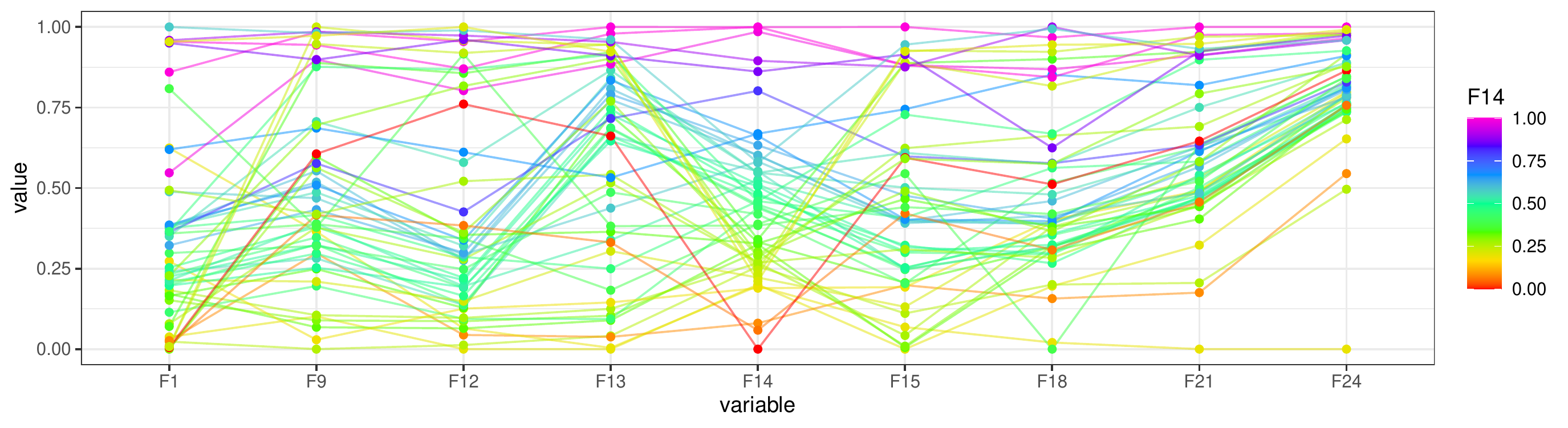}
	\caption{Parallel coordinate plots showing the relative performance of each best-of-run optimiser within a batch of runs. Plots are shown for the five batches of runs that contained the best optimiser for each training function, i.e. from top to bottom: $F_1$ 5$\times$200, $F_9$ 1$\times$1000, $F_{12}$ 5$\times$200, $F_{13}$ 1$\times$1000, $F_{14}$ 25$\times$20. Each line depicts the performance profile of a particular optimiser (low values are better), and the colour of the line shows its relative performance on the function that was used in training, i.e. those towards the red end of the spectrum had a relatively high fitness on their training function.}
	\label{fig:parallelcoords}
\end{figure}

\begin{table}[tb!]
	\centering
	\caption{Mean errors for hybrid optimisers, for 25 runs of 50D functions with a budget of $1E+3$ FEs. For each training function, for the batch of 50 runs containing the best optimiser, results are shown for hybrids formed from a pool composed of the 50 best-in-run optimisers. Results are also shown for function-agnostic pools comprising all best-in-run optimisers for a particular swarm size $\times$ iterations configuration. For each pool, the best result for each function is underlined. The best result overall for each function is highlighted in green. }\label{table:ensemble}
	\resizebox{\textwidth}{!}{
		\begin{tabular}{@{}lrlllllllll@{}}
			\toprule
			Pool & Size &  $F_1$ & $F_9$ & $F_{12}$ & $F_{13}$ & $F_{14}$ & $F_{15}$ & $F_{18}$ & $F_{21}$ & $F_{24}$  \\
			\midrule
			$F_1$ & 1     & 1.89E+5 & 9.40E+2 & 9.25E+6 & 8.33E+5 & 2.43E+1 & 1.48E+3 & 1.45E+3 & 1.57E+3 & 1.73E+3 \\
			5x200 & 5     & 1.23E+5 & 9.18E+2 & 8.68E+6 & 4.13E+6 & 2.43E+1 & 1.34E+3 & 1.50E+3 & 1.65E+3 & 1.77E+3 \\
			& 10    & 4.88E+4 & 6.89E+2 & 5.26E+6 & 3.84E+2 & 2.40E+1 & 1.21E+3 & \cellcolor[rgb]{ .663,  .816,  .557}\underline{9.00E+2} & 1.41E+3 & 1.47E+3 \\
			& 20    & \underline{3.21E+4} & \underline{6.25E+2} & \underline{4.21E+6} & \underline{3.66E+2} & \underline{2.39E+1} & \underline{1.10E+3} & \cellcolor[rgb]{ .663,  .816,  .557}\underline{9.00E+2} & \underline{1.40E+3} & \underline{1.46E+3} \\
			\midrule
			$F_{9}$ & 1     & 1.27E+5 & \cellcolor[rgb]{ .663,  .816,  .557}\underline{1.51E+2} & 7.52E+5 & 9.92E+1 & 2.40E+1 & \underline{4.26E+2} & 1.31E+3 & 1.41E+3 & 1.52E+3 \\
			1x1000 & 5     & \underline{3.05E+4} & 1.65E+2 & \underline{6.59E+5} & \underline{7.38E+1} & \underline{2.39E+1} & 4.43E+2 & 1.29E+3 & \underline{1.38E+3} & \underline{1.46E+3} \\
			& 10    & 5.45E+4 & 2.09E+2 & 1.07E+6 & 2.96E+2 & \underline{2.39E+1} & 4.99E+2 & 1.26E+3 & 1.39E+3 & 1.48E+3 \\
			& 20    & 4.92E+4 & 2.71E+2 & 1.17E+6 & 8.50E+1 & \underline{2.39E+1} & 5.64E+2 & \cellcolor[rgb]{ .663,  .816,  .557}\underline{9.00E+2} & \underline{1.38E+3} & 1.48E+3 \\
			\midrule
			$F_{12}$ & 1     & 1.28E+5 & 4.77E+2 & 3.07E+6 & \underline{2.33E+2} & \underline{2.40E+1} & \underline{8.28E+2} & \underline{9.11E+2} & \underline{1.39E+3} & 1.51E+3 \\
			5x200 & 5     & 1.25E+5 & \underline{4.58E+2} & \underline{1.63E+6} & 4.24E+4 & 2.41E+1 & 8.67E+2 & 9.45E+2 & 1.40E+3 & \underline{1.49E+3} \\
			& 10    & \underline{1.23E+5} & 4.82E+2 & 2.15E+6 & 1.69E+3 & \underline{2.40E+1} & 9.15E+2 & 1.02E+3 & 1.41E+3 & 1.51E+3 \\
			& 20    & 1.37E+5 & 5.41E+2 & 2.95E+6 & 1.05E+5 & 2.41E+1 & 1.02E+3 & 1.05E+3 & 1.45E+3 & 1.55E+3 \\
			\midrule
			$F_{13}$ & 1     & 2.73E+5 & 3.09E+2 & 1.07E+6 & 8.20E+3 & 2.46E+1 & 5.53E+2 & 1.48E+3 & 1.54E+3 & 1.86E+3 \\
			1x1000 & 5     & 1.02E+5 & \underline{1.52E+2} & \cellcolor[rgb]{ .663,  .816,  .557}\underline{5.01E+5} & \cellcolor[rgb]{ .663,  .816,  .557}\underline{2.52E+1} & \underline{2.39E+1} & \cellcolor[rgb]{ .663,  .816,  .557}\underline{4.09E+2} & \underline{9.03E+2} & 1.37E+3 & 1.48E+3 \\
			& 10    & 1.04E+5 & 2.10E+2 & 5.26E+5 & 2.53E+1 & 2.40E+1 & 4.71E+2 & 9.05E+2 & 1.41E+3 & 1.51E+3 \\
			& 20    & \underline{8.00E+4} & 3.49E+2 & 8.92E+5 & 6.28E+1 & 2.42E+1 & 7.05E+2 & 9.19E+2 & \cellcolor[rgb]{ .663,  .816,  .557}\underline{1.35E+3} & \cellcolor[rgb]{ .663,  .816,  .557}\underline{1.43E+3} \\
			\midrule
			$F_{14}$ & 1     & 6.29E+4 & 8.29E+2 & 9.00E+6 & 3.63E+5 & \cellcolor[rgb]{ .663,  .816,  .557}\underline{2.36E+1} & 1.27E+3 & 1.38E+3 & 1.49E+3 & 1.53E+3 \\
			25x40 & 5     & \underline{4.66E+4} & 6.88E+2 & \underline{4.11E+6} & 5.55E+3 & 2.37E+1 & \underline{1.04E+3} & 1.29E+3 & \underline{1.39E+3} & \underline{1.44E+3} \\
			& 10    & 4.84E+4 & 7.09E+2 & 5.19E+6 & 1.07E+4 & 2.38E+1 & 1.14E+3 & 1.22E+3 & 1.42E+3 & 1.47E+3 \\
			& 20    & 5.81E+4 & \underline{6.69E+2} & 4.59E+6 & \underline{6.35E+2} & 2.38E+1 & 1.10E+3 & \underline{9.43E+2} & 1.41E+3 & 1.46E+3 \\
			\midrule
			All & 5     & 5.53E+4 & 4.66E+2 & 2.19E+6 & 3.97E+2 & 2.38E+1 & 8.92E+2 & 1.18E+3 & 1.40E+3 & 1.48E+3 \\
			1x1000 & 25    & \cellcolor[rgb]{ .663,  .816,  .557}\underline{3.01E+4} & \underline{3.48E+2} & \underline{1.30E+6} & 2.02E+2 & \underline{2.37E+1} & 7.03E+2 & 9.87E+2 & 1.39E+3 & 1.47E+3 \\
			& 50    & 3.80E+4 & 3.61E+2 & 1.54E+6 & 1.53E+2 & 2.39E+1 & \underline{6.62E+2} & 9.01E+2 & 1.40E+3 & 1.47E+3 \\
			& 100   & 3.67E+4 & 4.10E+2 & 1.81E+6 & \underline{1.18E+2} & \underline{2.37E+1} & 7.68E+2 & \cellcolor[rgb]{ .663,  .816,  .557}\underline{9.00E+2} & \underline{1.36E+3} & \underline{1.45E+3} \\
			\midrule
			All & 5     & \underline{3.19E+4} & \underline{2.87E+2} & \underline{1.15E+6} & 2.19E+3 & 2.40E+1 & \underline{6.28E+2} & 1.01E+3 & \underline{1.37E+3} & 1.50E+3 \\
			5x200 & 25    & 5.27E+4 & 3.80E+2 & 1.58E+6 & 1.19E+2 & 2.40E+1 & 7.53E+2 & \cellcolor[rgb]{ .663,  .816,  .557}\underline{9.00E+2} & 1.40E+3 & 1.49E+3 \\
			& 50    & 5.41E+4 & 4.08E+2 & 1.82E+6 & \underline{1.12E+2} & 2.39E+1 & 7.87E+2 & \cellcolor[rgb]{ .663,  .816,  .557}\underline{9.00E+2} & 1.39E+3 & 1.49E+3 \\
			& 100   & 5.75E+4 & 4.72E+2 & 2.35E+6 & 1.59E+2 & \underline{2.38E+1} & 8.14E+2 & \cellcolor[rgb]{ .663,  .816,  .557}\underline{9.00E+2} & 1.39E+3 & \underline{1.48E+3} \\
			\midrule
			All & 5     & 8.30E+4 & 4.51E+2 & 2.77E+6 & 4.20E+2 & \underline{2.38E+1} & 8.63E+2 & \underline{9.00E+2} & 1.40E+3 & 1.49E+3 \\
			25x40 & 25    & \underline{4.53E+4} & \underline{4.42E+2} & 2.61E+6 & 3.60E+2 & \underline{2.38E+1} & 8.44E+2 & \underline{9.00E+2} & \underline{1.37E+3} & \underline{1.47E+3} \\
			& 50    & 5.39E+4 & 4.59E+2 & 2.53E+6 & \underline{3.01E+2} & 2.39E+1 & 8.62E+2 & 9.16E+2 & 1.39E+3 & 1.49E+3 \\
			& 100   & 6.02E+4 & 4.85E+2 & \underline{2.40E+6} & 3.29E+2 & \underline{2.38E+1} & \underline{8.40E+2} & 9.01E+2 & \underline{1.37E+3} & \underline{1.47E+3} \\
			\midrule
			All & 5     & 9.64E+4 & 5.74E+2 & \underline{3.20E+6} & 7.70E+3 & \underline{2.38E+1} & 9.37E+2 & 1.30E+3 & \underline{1.39E+3} & \underline{1.47E+3} \\
			50x20 & 25    & \underline{7.27E+4} & 5.85E+2 & 3.73E+6 & 2.82E+4 & 2.39E+1 & 9.95E+2 & 1.32E+3 & \underline{1.39E+3} & 1.48E+3 \\
			& 50    & 7.41E+4 & 5.70E+2 & 3.59E+6 & 1.10E+3 & 2.39E+1 & 9.58E+2 & 1.06E+3 & \underline{1.39E+3} & 1.48E+3 \\
			& 100   & 7.83E+4 & \underline{5.67E+2} & 3.54E+6 & \underline{2.98E+2} & \underline{2.38E+1} & \underline{9.28E+2} & \underline{9.00E+2} & 1.40E+3 & \underline{1.47E+3} \\
			\bottomrule
		\end{tabular}%
	}
\end{table}%

\begin{figure}[tbp!]
	%\centering
	
	\subfloat[][Using a pool size of 5, on $F_{12}$ and $F_{15}$] {
		\makebox[\textwidth]{
			\includegraphics[width=0.35\columnwidth, trim={0.7cm 0 3.5cm 1.5cm},clip]{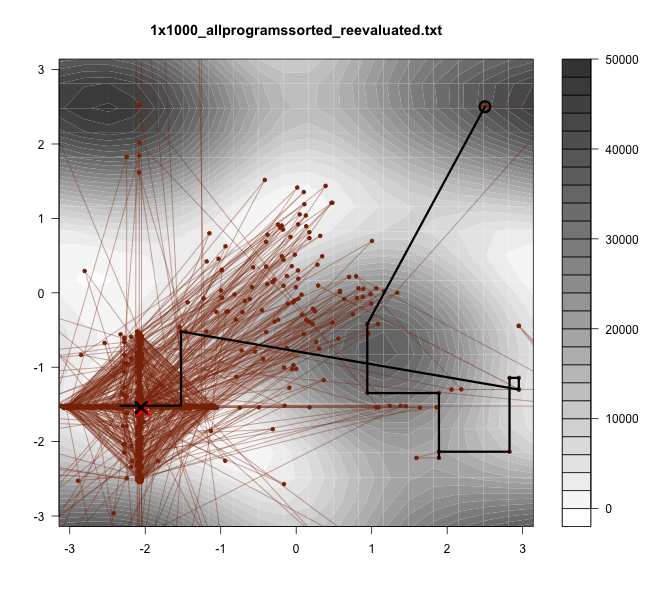}
			\includegraphics[width=0.35\columnwidth, trim={0.7cm 0 3.5cm 1.5cm},clip]{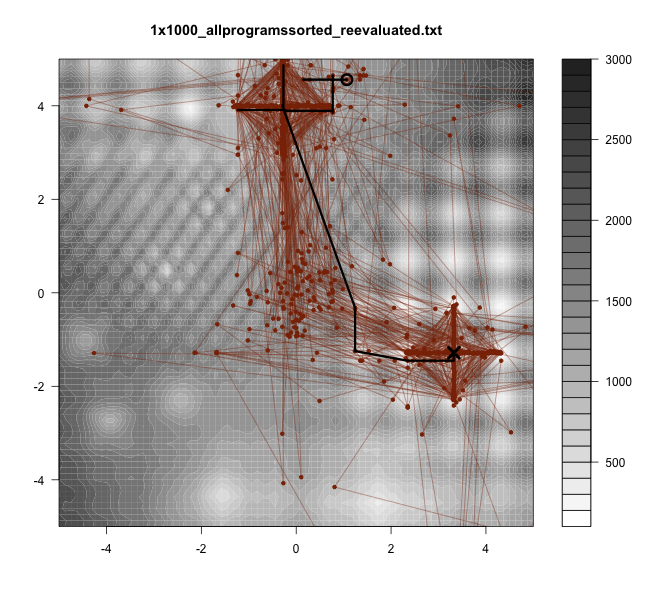}
		} \vspace{-3mm}
		\label{fig:ensemble::5}
	}
	
	\subfloat[][Using a pool size of 20, on $F_{21}$ and $F_{24}$] {
		\makebox[\textwidth]{
			\includegraphics[width=0.35\columnwidth, trim={0.7cm 0 3.5cm 1.5cm},clip]{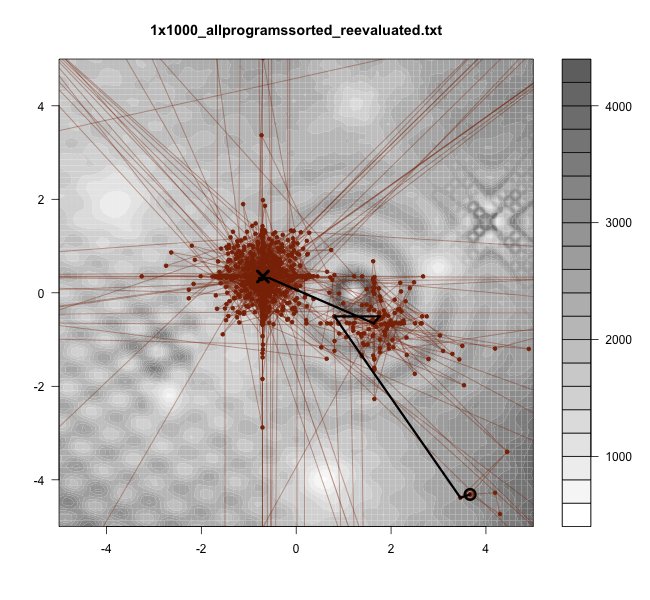}
			\includegraphics[width=0.35\columnwidth, trim={0.7cm 0 3.5cm 1.5cm},clip]{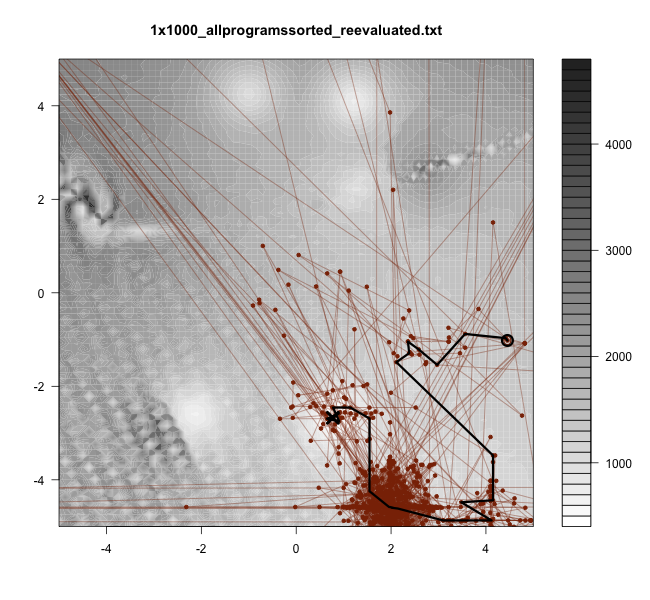}
		} \vspace{-3mm}
		\label{fig:ensemble::20}
	}
	
	\subfloat[][Some of the pool members] {
		\makebox[\textwidth]{
			\begin{tabular}[b]{c}% since line breaks don't work in subfloat
				\includegraphics[width=0.3\columnwidth, trim={0.7cm 0 3.5cm 1.5cm},clip]{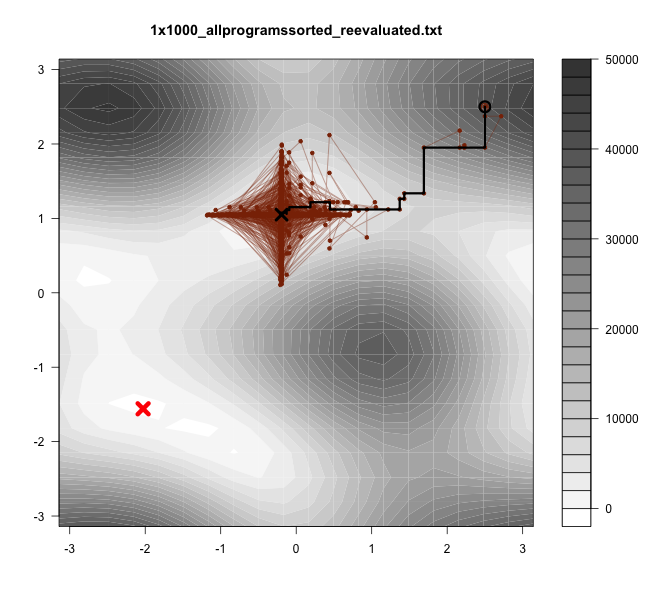}
				\includegraphics[width=0.3\columnwidth, trim={0.7cm 0 3.5cm 1.5cm},clip]{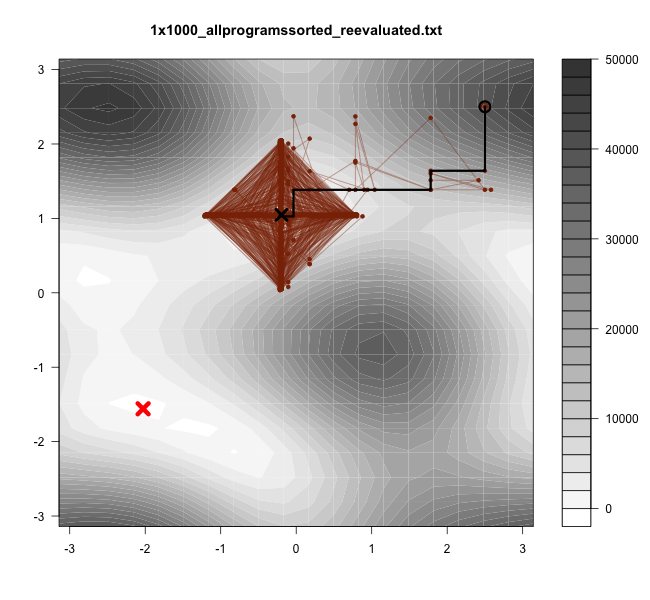}
				\includegraphics[width=0.3\columnwidth, trim={0.7cm 0 3.5cm 1.5cm},clip]{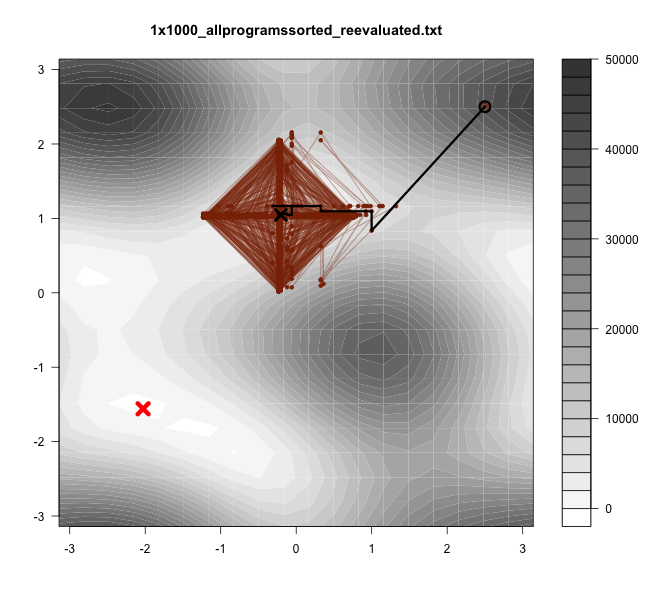}\vspace{-3mm} \\
				\includegraphics[width=0.3\columnwidth, trim={0.7cm 0 3.5cm 1.5cm},clip]{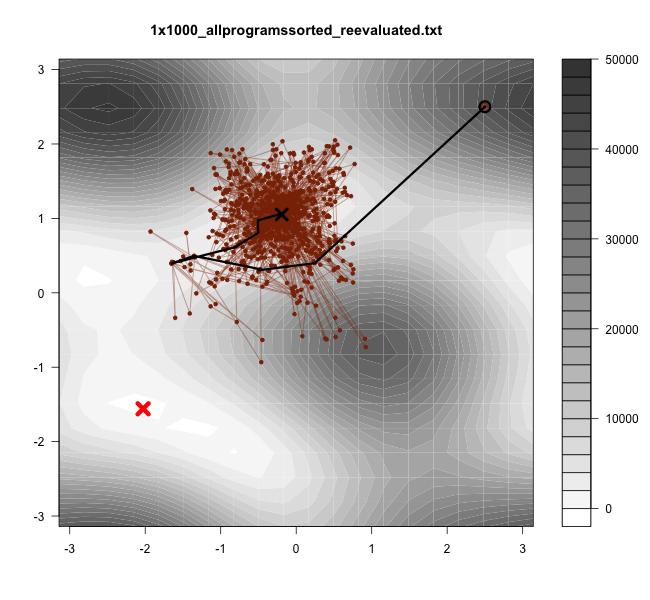}
				\includegraphics[width=0.3\columnwidth, trim={0.7cm 0 3.5cm 1.5cm},clip]{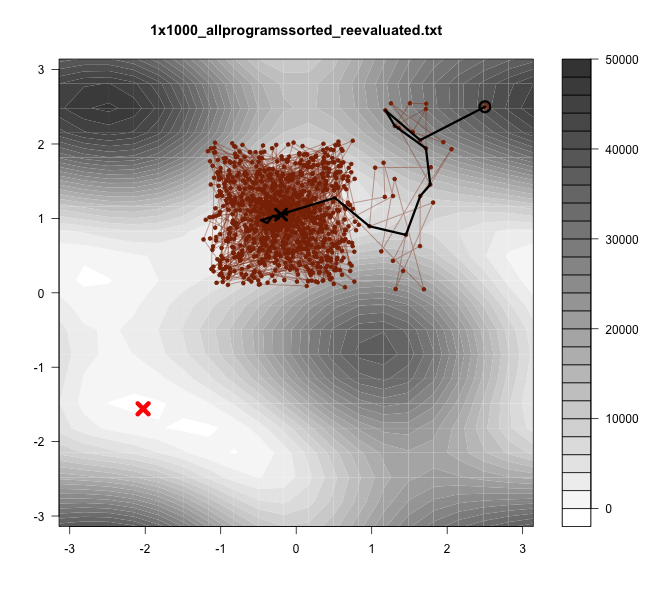}
				\includegraphics[width=0.3\columnwidth, trim={0.7cm 0 3.5cm 1.5cm},clip]{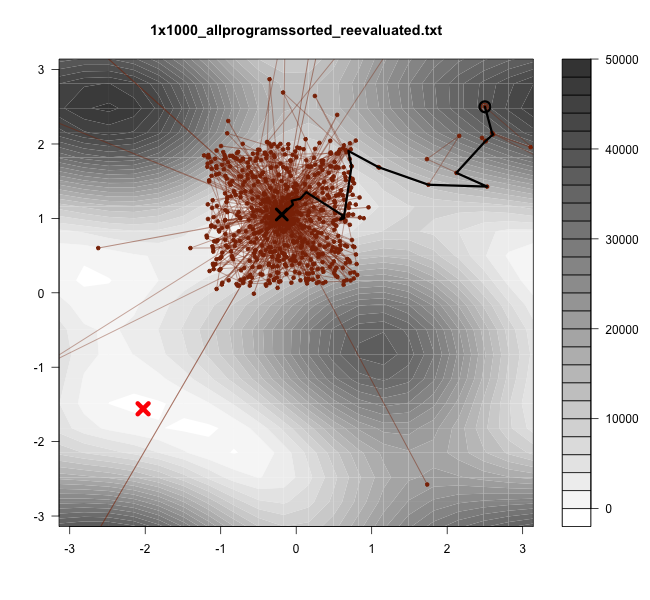}
			\end{tabular}
		} \vspace{-3mm}
		\label{fig:ensemble::pool}
	}
	
	\caption{Example trajectories of $F_{13}$ hybrid optimisers constructed from \protect\subref{fig:ensemble::5} 5 and \protect\subref{fig:ensemble::20} 20 optimisers, and \protect\subref{fig:ensemble::pool} some of their component optimisers.}
	\label{fig:ensemble}
\end{figure}

\begin{figure}[htb!]
	\centering
	\subfloat[][All 1$\times$1000, $n=25$, on $F_{1}$] {
		\includegraphics[width=0.45\columnwidth, trim={0.7cm 0 3.5cm 1.5cm},clip]{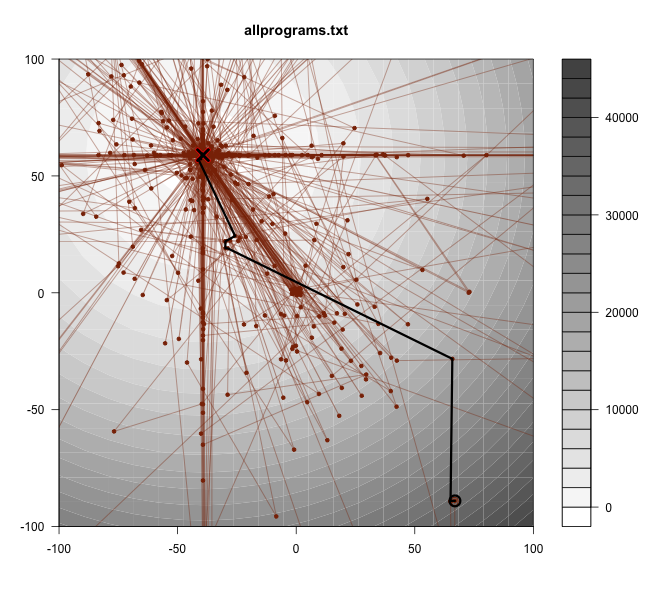}
		\label{fig:ensembles:all}
	}
	\subfloat[][$F_1$ 5$\times$200, $n=10$, on $F_{18}$] {
		\includegraphics[width=0.45\columnwidth, trim={0.7cm 0 3.5cm 1.5cm},clip]{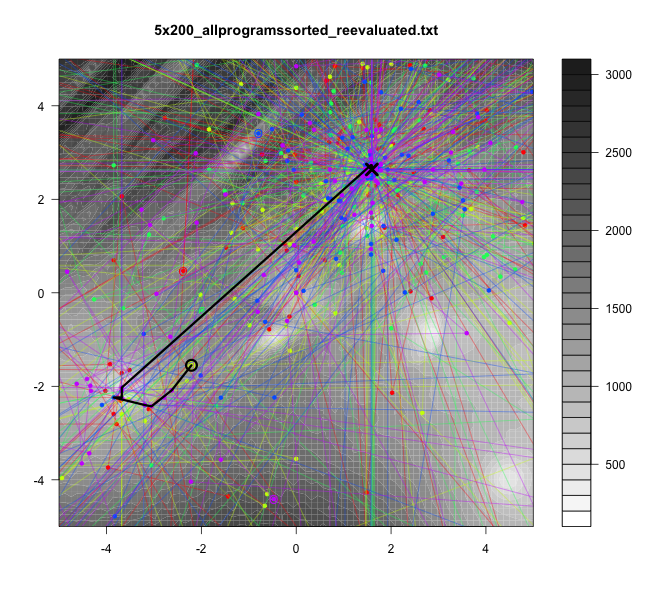}
		\label{fig:ensembles:f1}
	}
	\caption{Some more examples of hybrid optimisers.}
	\label{fig:ensembles}
\end{figure}

Another way to get some insight into the amount of behavioural diversity is to look at whether there are any notable differences in how the optimisers within a batch of runs (corresponding to a particular configuration and training function) generalise to the nine different functions. Fig. \ref{fig:parallelcoords} uses parallel coordinate plots to show this; specifically, for each optimiser within a batch of 50 runs, how they perform relative to one another on each of the CEC 2005 functions. In general, it is evident that not all the optimisers in a batch have the same performance profile across functions. For the $F_{13}$ and $F_{14}$ batches, in particular, it is clear that there are optimisers within the batch that generalise to different functions to the best optimiser within the batch. For example, the $F_{13}$ batch shows three groups: the best $F_{13}$ optimisers (in red), which mostly generalise to $F_{9}$, $F_{12}$ and $F_{15}$, but not to $F_{18}$ and higher; the intermediate $F_{13}$ optimisers (orange), which generalise better to $F_{18}$ and higher, and the weaker $F_{13}$ optimisers (green), which generalise to $F_{14}$  and do better than the first group on $F_{18}$ and higher. However, to varying degrees, this stratification is also present in the other batches, and suggests that repeated PushGP runs do lead to significant behavioural diversity.

As an aside, the $F_{14}$ plot is also notable for showing that the best $F_{14}$ optimisers do not generalise well to most other problems (with the exception of $F_{1}$). Likewise, it is mostly the case that the best optimisers for the other training functions do not generalise well to $F_{14}$, as shown by the upward peak at $F_{14}$ in most of the plots. This suggests that $F_{14}$ may not be a good function for training generalisable optimisers. The same appears true, to a lesser extent, for $F_{13}$; however, the best optimisers here still generalise to several other functions. For optimisers trained on the lower-numbered functions, by comparison, the best optimisers generalise relatively well to the other functions (at least within their batch). This is especially true for $F_{12}$, where the optimisers which do best on $F_{12}$ also tend to do best on the other functions. Taking into account the broader good performance of the best $F_{12}$ optimiser, this may point towards $F_{12}$ being a sweet-spot in terms of complexity --- with the higher-numbered functions being too complex, and therefore encouraging over-learning of their topological characteristics, and the lower-numbered functions being too simple to train behaviourally complex optimisers. However, this is a somewhat speculative conclusion, and requires further investigation.

\subsection{Hybridisation of Evolved Optimisers}

Another potential benefit of having diverse solutions is the capacity to hybridise optimisers. Much like ensembles in machine learning, hybridisation offers the potential to combine multiple optimisers in order to improve their generality. Here we take an initial look at this idea, by combining multiple evolved Push optimisers into one hybrid optimiser. In Section \ref{hybridising}, a simple approach to hybridising an arbitrary number of evolved Push programs into a single optimiser with an arbitrary swarm size was outlined. This involves each swarm member randomly selecting a Push program (from amongst a pool of previously evolved programs) at each iteration. Random allocation on a per-iteration basis avoids having to match the size of the pool to the swarm size. The disadvantage of this approach is that different programs have to share their stack states between invocations. However, initial experiments suggested that this method is more effective than persistently assigning a different program to each swarm member. In future work, it would be interesting to look more closely at the effect of \textit{how} the programs are hybridised, but here the investigation is restricted to evaluating the effect of \textit{which} programs are hybridised. 

Table \ref{table:ensemble} shows the performance of hybrids that are constructed from various pools of evolved Push optimisers. There are two types of pools used. First, function-specific pools. For each of the training functions, these pools are assembled from the best-in-run optimisers from the batch where the best optimiser for that function was evolved; for instance, for $F_1$, the best evolved optimiser had a configuration of $5\times200$, so the pool is constructed by selecting the best optimiser evolved in each of the 50 runs in the $F_1$ $5\times200$ batch. Second, function-agnostic pools, where, for each of the swarm size $\times$ iterations configurations, all of the best-in-run optimisers for that configuration are assembled into a pool. For instance, for the All $1\times1000$ pool, the pool is assembled from the $1\times1000$ batches for each of the five training functions, forming a pool of 250 optimisers. Hybrid optimisers are then constructed from the top $n$ solutions present within each pool, for various values of $n$.

Results are only shown for the hardest (50D) function instances. On these functions, there appears to be a clear benefit to using hybridisation, at least in terms of mean error rate. Notably, for every function except $F_9$, the best error rate (shown in green) in Table \ref{table:ensemble} is produced by a hybrid optimiser, rather than by the stand-alone best optimiser from each pool (these are indicated by a pool size of $1$). Also, for every function-specific pool except $F_{12}$, one or more of the hybrids shows better generalisation across all the functions than the stand-alone best. However, the sweet spot, in terms of $n$, varies across the pools, with a hybrid of 5 optimisers best for the $F_9$, $F_{13}$ and $F_{14}$ pools, and a hybrid of size 20 best for the $F_1$ pool.

Another important observation is that hybrids constructed from function-specific pools tend to perform better than those constructed from function-agnostic pools, with the best error rate being found by function-specific hybrids for all but the easiest function, $F_1$. This is unexpected, since the optimisers trained on one function would be expected to be less diverse than those found within a pool of optimisers trained on different functions. However, the trend is quite clear in Table \ref{table:ensemble}. One possible explanation is that the optimisers within a function-specific pool may be better adapted to one another.

Hybrids assembled from optimisers trained on the $F_{13}$ function appear to generalise particularly well, producing the lowest error rates on five of the nine functions. Apart from $F_{9}$, the $n=5$ $F_{13}$ hybrid performs significantly better on every 50D function than the $F_{9}$ best optimiser, which was the winner amongst the stand-alone optimisers for this dimensionality. This is perhaps surprising given that the $F_{13}$ best optimiser performs relatively poorly on the 50D functions (see Table \ref{table:errors}). On the other hand, Fig. \ref{fig:parallelcoords} shows that the $F_{13}$ pool contains relatively high diversity in terms of the functions that the best-in-run optimisers perform well at. In the case of ensembles, it is beneficial for the models from which they are composed to make mistakes on different problem instances, and it seems plausible that the same would be true when hybridising optimisers.

To give some insight into the behaviour of the $F_{13}$ hybrids, Figs. \ref{fig:ensemble::5} and \ref{fig:ensemble::20} depict the trajectories of the $n=5$ and $n=20$ versions on a number of 2D functions. It is evident that, for at least the 2D versions of these functions, the hybrid optimisers are capable of hopping between different local optima basins on even the most challenging landscapes. Contrast this with the behaviour of some of the optimisers from which they are comprised, whose trajectories are shown in Fig. \ref{fig:ensemble::pool}. On $F_{12}$, the component optimisers struggle to escape from the local optimum basin nearest to their starting point, but the $n=5$ hybrid (Fig. \ref{fig:ensemble::5}, left) is able to hop around the $F_{12}$ landscape and find the global optimum basin.

These results indicate that hybridisation of evolved optimisers could be a productive avenue to explore further. A notable benefit is that it is not necessary to evolve the component optimisers each time. That is, it is only necessary to train a pool of optimisers once, and the pool can then be hybridised in different ways to address particular problems. However, a downside of hybridisation is that it makes optimisers harder to understand, with the behaviour of a hybrid likely to be emergent in unexpected ways from the behaviour of its component optimisers. Consider, for instance, Fig. \ref{fig:ensembles}. This shows the behaviour of two more hybrid optimisers, specifically those with the lowest error rates on $F_1$ and $F_{18}$. In both cases, the optimisers move to the region containing the optimum in a small number of moves. However, due to the large number of programs guiding their moves, the underlying logic behind the optimisation process is unclear, and this arguably takes away one of the advantages of GP over deep learning --- the relative interpretability of its solutions.

\section{Limitations and Future Work} \label{limitations}

This study set out to show two things. First, that PushGP can be used to design useful optimisers, and second, that it can discover optimisers that lie beyond the existing human design space. Both of these have been demonstrated, at least to an extent. PushGP has been shown able to design optimisers, and some of these optimisers have a performance that compares favourably against a well-regarded general-purpose optimiser, in many cases exceeding it. The analysis of evolved optimisers identified a number of novel behaviours, such as the use of trigonometric functions and certain geometric patterns for efficiently exploring a search volume, and the use of distributed backtracking behaviours to deal with deceptive landscapes.

Nevertheless, this is only an initial proof of concept, and there remains a lot more that could be done to explore the optimiser design space in a broader and more systematic way. One limitation of this study is the use of only a small number of functions to evolve optimisers. Whilst these were chosen to represent a range of landscapes, they are not representative of all optimisation landscapes, and this consequently limits the range of behaviours that can be evolved. It also raises the question of which set of functions would be appropriate for this task. Existing benchmark sets like BBOB \cite{hansen2010comparing} are a possibility, but there are known to be significant biases in how they sample the space of landscapes \cite{mersmann2015analyzing,christie2018investigating,lacroix2019limitations}. Perhaps a more productive direction would be to use tunable benchmark functions \cite{gallagher2006general,ronkkonen2011framework}, which generate landscapes with defined features. These features could potentially be used as an orthogonal basis with which to define a particular space of optimisation landscapes, providing a more uniform sampling of optimisation behaviours.

Conversely, by providing a varied selection of optimisers to be evaluated, perhaps the approach outlined in this paper could be used to assess benchmark functions, or even provide a better understanding of the scope for optimiser generality. The latter is important because, although the no free lunch theorem \cite{wolpert1997no} itself does not apply within a restricted subspace of functions such as continuous functions, it is widely believed that there are restrictions on the amount of generality that can be achieved within such sub-spaces. 

One of the objectives of this study was to explore diverse approaches to optimisation. Whilst this has been achieved to an extent both through the use of multiple training functions and the natural diversity that occurs between EA runs, there is also scope for explicitly promoting diversity. This could be done both at the population level, using a niching mechanism \cite{li2016seeking} to encourage the evolution of diverse solutions, or between runs, by penalising solutions which resemble those found in previous runs. The main barrier to doing this, however, lies in finding a suitable distance metric for comparing optimisers. Potentially this could be done by comparing trajectories. One approach would be to derive features from trajectories and compare these, for example search space coverage and move size statistics. This would be relatively easy to do, and would have the added benefit of allowing novelty search techniques such as MAP-elites \cite{mouret2015illuminating} to be applied, but the selection of features could easily introduce bias. An alternative would be to directly compare trajectories. This is much harder to do, but one potential approach would be to map trajectories to local optima networks (as done in \cite{blum2021comparative}), and then compare the trajectories within these networks.

Diversification methods could also be beneficial when generating optimiser pools for forming hybrids, since this could enable explicit selection of component optimisers that make errors on different landscapes. More generally, this study showed the potential benefit of hybridising pools of evolved optimisers. However, both the way in which component optimisers were selected, and the approach used to hybridise them, was somewhat arbitrary. Consequently, there is a lot more scope for research in this area. For instance, a wrapper method could be used to select which component optimisers to use from the pool.

Several of the other design choices made in this study may limit or bias the space of optimisers that is being explored. This includes requiring evolved optimisers to only carry out a single move each time they are invoked, which was done so that PushGP does not have to re-discover an outer loop for each optimiser. However, it is conceivable that this limits the design space, in particular pushing evolution towards optimisers that carry out similar kinds of move every time they are invoked, rather than adapting their behaviour in interesting ways as they explore the optimisation landscape. For the population-based optimisers, the use of persistent search processes and modes of inter-process communication resembling those in PSO are also likely to introduce bias. To address this, in future work, it would be interesting to explore a wider range of models of population-based search.

A number of the design choices made in this study reflected the need to limit computational effort.
Computational effort is undoubtedly a limiting factor for this kind of research, but there are potential approaches for reducing the effort required to evolve optimisers. This includes general EA-based approaches for reducing fitness evaluation time, such as surrogate fitness models \cite{ong2003evolutionary}. Another approach would be to promote the evolution of more efficient optimisers which take less time to evaluate, e.g. by penalising optimisers that use excessive instruction executions. Time and space complexity were not considered in this study, but it would be useful to do so, especially if the aim is to generate practical optimisers.

This study only looked at the design of optimisers for solving continuous functions, and it would be interesting to apply the approach to other domains. Combinatorial optimisation is one possibility, and this could be done by modifying the PushGP function set so that only valid solutions can be generated. An intriguing possibility would be to apply the approach to GP itself, evolving PushGP programs that can optimise programs. The simplest way of doing this would be to optimise grammatical evolution \cite{ryan1998grammatical} programs, since these are normally represented by fixed-length vectors of numbers. Indeed, anything that can be represented as a vector of numbers could potentially be a target of this approach, with relatively little modification.

%\afterpage{\FloatBarrier}

\section{Conclusions} \label{conclusions}

This work uses genetic programming to design optimisers. The optimisers are expressed using the Push language, which allows them to be evolved largely from scratch using low-level primitives and control flow instructions. This discriminates the approach from earlier methods for evolving optimisers, which generally used high-level building blocks derived from existing optimisers. This means that the evolved optimisers are relatively unbiased towards existing human knowledge of how to do optimisation, offering the potential to discover truly novel approaches. This contrasts to recent nature-inspired approaches to optimiser design, many of which have failed to introduce any real novelty.

The results show that PushGP can both discover and express optimisation behaviours that are effective, complex and diverse. Importantly, the evolved optimisers generalise to problems they did not see during training, and often out-perform general-purpose optimisers on these previously unseen problems. Behavioural analysis shows that the evolved optimisers use a diverse range of strategies to explore optimisation landscapes, using behaviours that differ significantly from existing local and population-based optimisers.

This paper also takes an initial look at the hybridisation of evolved optimisers. The approach exploits the diversity found between evolutionary runs to form pools of optimisers, which are then hybridised using a simple mechanism. Many of the resulting hybrid optimisers were found to be significantly better than individual optimisers in terms of performance and generality, although at the expense of understandability. This suggests a potentially productive path for further research into the automatic design of optimisers.

%\begin{acknowledgements}
%\end{acknowledgements}


\bibliographystyle{spmpsci}      % mathematics and physical sciences
\bibliography{gpem2021}

\begin{thebibliography}{10}
\providecommand{\url}[1]{{#1}}
\providecommand{\urlprefix}{URL }
\expandafter\ifx\csname urlstyle\endcsname\relax
  \providecommand{\doi}[1]{DOI~\discretionary{}{}{}#1}\else
  \providecommand{\doi}{DOI~\discretionary{}{}{}\begingroup
  \urlstyle{rm}\Url}\fi

\bibitem{andrychowicz2016learning}
Andrychowicz, M., Denil, M., Gomez, S., Hoffman, M.W., Pfau, D., Schaul, T.,
  Shillingford, B., De~Freitas, N.: Learning to learn by gradient descent by
  gradient descent.
\newblock In: NIPS'16: Proceedings of the 30th International Conference on
  Neural Information Processing Systems, pp. 3988–--3996 (2016)

\bibitem{de2021similarity}
de~Armas, J., Lalla-Ruiz, E., Tilahun, S.L., Vo{\ss}, S.: Similarity in
  metaheuristics: a gentle step towards a comparison methodology.
\newblock Natural Computing  (2021).
\newblock \doi{10.1007/s11047-020-09837-9}

\bibitem{auger2005restart}
Auger, A., Hansen, N.: A restart {CMA} evolution strategy with increasing
  population size.
\newblock In: Proceedings of the IEEE Congress on Evolutionary Computation,
  \emph{IEEE CEC '05}, vol.~2, pp. 1769--1776. IEEE (2005)

\bibitem{blum2021comparative}
Blum, C., Ochoa, G.: A comparative analysis of two matheuristics by means of
  merged local optima networks.
\newblock European Journal of Operational Research \textbf{290}(1), 36--56
  (2021)

\bibitem{blum2016hybrid}
Blum, C., Raidl, G.R.: Hybrid Metaheuristics: Powerful Tools for Optimization.
\newblock Springer (2016)

\bibitem{bogdanova2019franken}
Bogdanova, A., Junior, J.P., Aranha, C.: Franken-swarm: grammatical evolution
  for the automatic generation of swarm-like meta-heuristics.
\newblock In: Proceedings of the Genetic and Evolutionary Computation
  Conference Companion, GECCO '19, pp. 411--412. ACM (2019)

\bibitem{burke2013hyper}
Burke, E.K., Gendreau, M., Hyde, M., Kendall, G., Ochoa, G., {\"O}zcan, E., Qu,
  R.: Hyper-heuristics: A survey of the state of the art.
\newblock Journal of the Operational Research Society \textbf{64}(12),
  1695--1724 (2013)

\bibitem{christie2018investigating}
Christie, L.A., Brownlee, A.E., Woodward, J.R.: Investigating benchmark
  correlations when comparing algorithms with parameter tuning.
\newblock In: Proceedings of the Genetic and Evolutionary Computation
  Conference Companion, pp. 209--210 (2018)

\bibitem{cotta2017memetic}
Cotta, C., Mathieson, L., Moscato, P.: Memetic algorithms.
\newblock In: R.~Mart{\'i}, P.M. Pardalos, M.G.C. Resende (eds.) Handbook of
  Heuristics. Springer (2017)

\bibitem{diocsan2006evolving}
Dio{\c{s}}an, L., Oltean, M.: Evolving crossover operators for function
  optimization.
\newblock In: European Conference on Genetic Programming, pp. 97--108. Springer
  (2006)

\bibitem{edmonds1998metagp}
Edmonds, B.: Meta-genetic programming: Co-evolving the operators of variation.
\newblock Tech. Rep. CPM Report 98-32, Manchester Metropolitan University
  (1998)

\bibitem{gallagher2006general}
Gallagher, M., Yuan, B.: A general-purpose tunable landscape generator.
\newblock IEEE transactions on evolutionary computation \textbf{10}(5),
  590--603 (2006)

\bibitem{goldman2011self}
Goldman, B.W., Tauritz, D.R.: Self-configuring crossover.
\newblock In: Proceedings of the Genetic and Evolutionary Computation
  Conference Companion, GECCO '11, pp. 575--582. ACM (2011)

\bibitem{graham2019thesis}
Graham, K.: An investigation of factors influencing algorithm selection for
  high dimensional continuous optimisation problems.
\newblock Ph.D. thesis, Computing Science and Mathematics, University of
  Stirling (2019)

\bibitem{hansen2010comparing}
Hansen, N., Auger, A., Ros, R., Finck, S., Po{\v{s}}{\'\i}k, P.: Comparing
  results of 31 algorithms from the black-box optimization benchmarking
  {BBOB}-2009.
\newblock In: Proceedings of the 12th annual conference companion on Genetic
  and evolutionary computation, pp. 1689--1696 (2010)

\bibitem{junior2020franken}
{Junior}, J.P., {Aranha}, C., {Sakurai}, T.: A training difficulty schedule for
  effective search of meta-heuristic design.
\newblock In: 2020 IEEE Congress on Evolutionary Computation, CEC 2020, pp.
  1--8 (2020)

\bibitem{kamrath2020automated}
Kamrath, N.R., Pope, A.S., Tauritz, D.R.: The automated design of local
  optimizers for memetic algorithms employing supportive coevolution.
\newblock In: Proceedings of the 2020 Genetic and Evolutionary Computation
  Conference Companion, pp. 1889--1897 (2020)

\bibitem{kantschik1999meta}
Kantschik, W., Dittrich, P., Brameier, M., Banzhaf, W.: Meta-evolution in graph
  {GP}.
\newblock In: European Conference on Genetic Programming, EuroGP 1999, pp.
  15--28. Springer (1999)

\bibitem{lacroix2019limitations}
Lacroix, B., McCall, J.: Limitations of benchmark sets and landscape features
  for algorithm selection and performance prediction.
\newblock In: Proceedings of the Genetic and Evolutionary Computation
  Conference Companion, pp. 261--262 (2019)

\bibitem{langdon2012genetic}
Langdon, W.B.: Genetic programming and data structures: {G}enetic {P}rogramming
  + {D}ata {S}tructures = {A}utomatic {P}rogramming!
\newblock Springer (2012)

\bibitem{li2016seeking}
Li, X., Epitropakis, M.G., Deb, K., Engelbrecht, A.: Seeking multiple
  solutions: an updated survey on niching methods and their applications.
\newblock IEEE Transactions on Evolutionary Computation \textbf{21}(4),
  518--538 (2016)

\bibitem{lones2019instruction}
Lones, M.A.: Instruction-level design of local optimisers using {Push GP}.
\newblock In: Proceedings of the Genetic and Evolutionary Computation
  Conference Companion, GECCO '19, pp. 1487--1494. ACM (2019)

\bibitem{lones2019mitigating}
Lones, M.A.: Mitigating metaphors: A comprehensible guide to recent
  nature-inspired algorithms.
\newblock SN Computer Science \textbf{1}(1), 49 (2020)

\bibitem{lones2020eurogp}
Lones, M.A.: Optimising optimisers with {Push GP}.
\newblock In: Proceedings of the 2020 European Conference on Genetic
  Programming (EuroGP), \emph{LNCS}, vol. 12101. Springer (2020).
\newblock \doi{10.1007/978-3-030-44094-7_7}

\bibitem{lourencco2013learning}
Louren{\c{c}}o, N., Pereira, F., Costa, E.: Learning selection strategies for
  evolutionary algorithms.
\newblock In: International Conference on Artificial Evolution (Evolution
  Artificielle), pp. 197--208. Springer (2013)

\bibitem{handbookofheuristics}
Mart{\'i}, R., Pardalos, P.M., Resende, M.G.C. (eds.): Handbook of Heuristics.
\newblock Springer (2018)

\bibitem{martin2013evolving}
Martin, M.A., Tauritz, D.R.: Evolving black-box search algorithms employing
  genetic programming.
\newblock In: Proceedings of the Genetic and Evolutionary Computation
  Conference Companion, GECCO '13, pp. 1497--1504. ACM (2013)

\bibitem{mersmann2015analyzing}
Mersmann, O., Preuss, M., Trautmann, H., Bischl, B., Weihs, C.: Analyzing the
  {BBOB} results by means of benchmarking concepts.
\newblock Evolutionary computation \textbf{23}(1), 161--185 (2015)

\bibitem{metz2019learned}
Metz, L., Maheswaranathan, N., Nixon, J., Freeman, D., Sohl-dickstein, J.:
  Learned optimizers that outperform {SGD} on wall-clock and test loss.
\newblock In: Proceedings of the 2nd Workshop on Meta-Learning, MetaLearn 2018
  (2018)

\bibitem{mouret2015illuminating}
Mouret, J.B., Clune, J.: Illuminating search spaces by mapping elites.
\newblock arXiv preprint arXiv:1504.04909  (2015)

\bibitem{oltean2005evolving}
Oltean, M.: Evolving evolutionary algorithms using linear genetic programming.
\newblock Evolutionary Computation \textbf{13}(3), 387--410 (2005)

\bibitem{ong2003evolutionary}
Ong, Y.S., Nair, P.B., Keane, A.J.: Evolutionary optimization of
  computationally expensive problems via surrogate modeling.
\newblock AIAA journal \textbf{41}(4), 687--696 (2003)

\bibitem{poli2005exploring}
Poli, R., Di~Chio, C., Langdon, W.B.: Exploring extended particle swarms: a
  genetic programming approach.
\newblock In: Proceedings of the 7th annual conference on Genetic and
  evolutionary computation, pp. 169--176 (2005)

\bibitem{real2020automlzero}
Real, E., Liang, C., So, D., Le, Q.: {AutoML}-zero: evolving machine learning
  algorithms from scratch.
\newblock In: Proc. 37th Int. Conf. on Machine Learning, ICML, pp. 8007--8019.
  PMLR (2020)

\bibitem{richter2018automated}
Richter, S.N., Tauritz, D.R.: The automated design of probabilistic selection
  methods for evolutionary algorithms.
\newblock In: Proceedings of the Genetic and Evolutionary Computation
  Conference Companion, GECCO '18, pp. 1545--1552. ACM (2018)

\bibitem{van2016evolving}
van Rijn, S., Wang, H., van Leeuwen, M., B{\"a}ck, T.: Evolving the structure
  of evolution strategies.
\newblock In: 2016 IEEE Symposium Series on Computational Intelligence (SSCI),
  pp. 1--8. IEEE (2016)

\bibitem{ronkkonen2011framework}
R{\"o}nkk{\"o}nen, J., Li, X., Kyrki, V., Lampinen, J.: A framework for
  generating tunable test functions for multimodal optimization.
\newblock Soft Computing \textbf{15}(9), 1689--1706 (2011)

\bibitem{ross2002searching}
Ross, B.J.: Searching for search algorithms: Experiments in meta-search.
\newblock Tech. rep., Technical Report CS-02-23, Department of Computer
  Science, Brock University (2002)

\bibitem{ryan1998grammatical}
Ryan, C., Collins, J.J., Neill, M.O.: Grammatical evolution: Evolving programs
  for an arbitrary language.
\newblock In: European Conference on Genetic Programming, pp. 83--96. Springer
  (1998)

\bibitem{ryser2016iterative}
Ryser-Welch, P., Miller, J.F., Swan, J., Trefzer, M.A.: Iterative cartesian
  genetic programming: Creating general algorithms for solving travelling
  salesman problems.
\newblock In: European Conference on Genetic Programming, EuroGP '16, pp.
  294--310. Springer (2016)

\bibitem{shirakawa2009evolution}
Shirakawa, S., Nagao, T.: Evolution of search algorithms using graph structured
  program evolution.
\newblock In: European Conference on Genetic Programming, pp. 109--120.
  Springer (2009)

\bibitem{sorensen2015metaheuristics}
S{\"o}rensen, K.: Metaheuristics---the metaphor exposed.
\newblock International Transactions in Operational Research \textbf{22}(1),
  3--18 (2015)

\bibitem{spector2001autoconstructive}
Spector, L.: Autoconstructive evolution: Push, {pushGP}, and pushpop.
\newblock In: Proceedings of the Genetic and Evolutionary Computation
  Conference, \emph{GECCO '19}, vol. 137 (2001)

\bibitem{spector2004push}
Spector, L., Perry, C., Klein, J., Keijzer, M.: Push 3.0 programming language
  description.
\newblock Tech. rep., HC-CSTR-2004-02, School of Cognitive Science, Hampshire
  College (2004)

\bibitem{spector2002genetic}
Spector, L., Robinson, A.: Genetic programming and autoconstructive evolution
  with the push programming language.
\newblock Genetic Programming and Evolvable Machines \textbf{3}(1), 7--40
  (2002)

\bibitem{stork2020new}
Stork, J., Eiben, A.E., Bartz-Beielstein, T.: A new taxonomy of global
  optimization algorithms.
\newblock Natural Computing pp. 1--24 (2020)

\bibitem{suganthan2005problem}
Suganthan, P.N., Hansen, N., Liang, J.J., Deb, K., Chen, Y.P., Auger, A.,
  Tiwari, S.: Problem definitions and evaluation criteria for the {CEC} 2005
  special session on real-parameter optimization.
\newblock KanGAL report \textbf{2005005} (2005)

\bibitem{teller1996aigp2}
Teller, A.: Evolving programmers: The co-evolution of intelligent recombination
  operators.
\newblock In: P.J. Angeline, K.E. {Kinnear, Jr.} (eds.) Advances in Genetic
  Programming 2, chap.~3, pp. 45--68. MIT Press (1996)

\bibitem{wichrowska2017learned}
Wichrowska, O., Maheswaranathan, N., Hoffman, M.W., Colmenarejo, S.G., Denil,
  M., de~Freitas, N., Sohl-Dickstein, J.: Learned optimizers that scale and
  generalize.
\newblock In: Proceedings of the 34th International Conference on Machine
  Learning-Volume 70, ICML '17, pp. 3751--3760 (2017)

\bibitem{wolpert1997no}
Wolpert, D.H., Macready, W.G.: No free lunch theorems for optimization.
\newblock IEEE Transactions on Evolutionary Computation \textbf{1}(1), 67--82
  (1997)

\bibitem{woodward2012automatic}
Woodward, J.R., Swan, J.: The automatic generation of mutation operators for
  genetic algorithms.
\newblock In: Proceedings of the Genetic and Evolutionary Computation
  Conference, GECCO '12, pp. 67--74. ACM (2012)

\end{thebibliography}

\end{document}